\pdfoutput=1
\documentclass[11pt]{article}

\usepackage{colortbl}
\usepackage{xcolor}
\usepackage{times}
\usepackage{soul}
\usepackage{url}
\usepackage{hyperref}
\usepackage[utf8]{inputenc}
\usepackage{amsthm}
\usepackage{multirow}
\usepackage{makecell}
\usepackage{bm}
\usepackage{amsmath}

\usepackage{array}
\newcolumntype{C}[1]{>{\centering\let\newline\\\arraybackslash\hspace{0pt}}m{#1}}
\newcolumntype{L}[1]{>{\raggedright\let\newline\\\arraybackslash\hspace{0pt}}m{#1}}

\usepackage{multicol}
\usepackage{wrapfig}

\usepackage{algorithm}
\usepackage{algorithmic}
\definecolor{LightCyan}{rgb}{0.88,1,1}
\definecolor{WildStrawberry}{rgb}{1.0, 0.26, 0.64}
\definecolor{YellowGreen}{rgb}{0.6, 0.8, 0.2}
\definecolor{GoldenRod}{rgb}{0.85, 0.65, 0.13}
\usepackage[final]{acl}
\usepackage{subfigure}
\usepackage{subcaption}

\usepackage{times}
\usepackage{latexsym}

\usepackage[T1]{fontenc}
\usepackage[utf8]{inputenc}

\usepackage{microtype}

\usepackage{inconsolata}

\usepackage{graphicx}

\title{MIDAS: Multi-level Intent, Domain, And Slot\\Knowledge Distillation for Multi-turn NLU}

\author{Yan Li$^{1,3}$, So-Eon Kim$^2$, Seong-Bae Park$^2$, Soyeon Caren Han$^{1,3,}$\thanks{Corresponding Author}, \\
$^1$The University of Sydney, $^2$Kyung Hee University,\\ $^3$The University of Melbourne \\
$^1$\texttt{yali3816@uni.sydney.edu.au}, $^2$\texttt{\{sekim0211, sbpark71\}@khu.ac.kr}, \\ $^3$\texttt{caren.han@unimelb.edu.au}\\
}

\begin{document}
\maketitle

\begin{abstract}
Although Large Language Models (LLMs) can generate coherent text, they often struggle to recognise user intent behind queries. In contrast, Natural Language Understanding (NLU) models interpret the purpose and key information of user input for responsive interactions. Existing NLU models typically map utterances to a dual-level semantic frame, involving sentence-level intent (SI) and word-level slot (WS) labels. However, real-life conversations primarily consist of multi-turn dialogues, requiring the interpretation of complex and extended exchanges. Researchers encounter challenges in addressing all facets of multi-turn dialogue using a unified NLU model. This paper introduces MIDAS, a novel approach leveraging multi-level intent, domain, and slot knowledge distillation for multi-turn NLU. We construct distinct teachers for SI detection, WS filling, and conversation-level domain (CD) classification, each fine-tuned for specific knowledge. A multi-teacher loss is proposed to facilitate the integration of these teachers, guiding a student model in multi-turn dialogue tasks. Results demonstrate the efficacy of our model in improving multi-turn conversation understanding, showcasing the potential for advancements in NLU through multi-level dialogue knowledge distillation. Our implementation is open-sourced on GitHub\footnote{\url{https://github.com/adlnlp/Midas}}.

% Although Large Language Models(LLMs) can generate coherent and contextually relevant text, they often struggle to recognise the intent behind the human user's query. Natural Language Understanding (NLU) models, however, interpret the purpose and key information of the user's input to enable responsive interactions.
%  Existing NLU models generally map individual utterances to a dual-level semantic frame, involving SI and WS labels. However, real-life conversations primarily consist of multi-turn conversations, involving the interpretation of complex and extended dialogues. Researchers encounter challenges addressing all facets of multi-turn dialogue conversations using a unified single NLU model.
% This paper introduces a novel approach, MIDAS, leveraging a multi-level intent, domain, and slot knowledge distillation for multi-turn NLU. To achieve this, we construct distinct teachers for varying levels of conversation knowledge, namely, SI detection, WS filling, and CD classification. These teachers are then fine-tuned to acquire specific knowledge of their designated levels. A multi-teacher loss is proposed to facilitate the combination of these multi-level teachers, guiding a student model in multi-turn dialogue tasks. 
% The experimental results demonstrate the efficacy of our model in improving the overall multi-turn conversation understanding, showcasing the potential for advancements in NLU models through the incorporation of multi-level dialogue knowledge distillation techniques.
\end{abstract}

% \begin{wrapfigure}{r}{0.65\textwidth}
%   \vspace*{-4mm}
%   \begin{center}
%     \includegraphics[width=\linewidth]{images/conv_sample.png}
%   \end{center}
%   \vspace*{-2mm}
%   \label{fig:m2m}
%   \caption{An example of conversations with WSs, sentence-level sentences, and conversation-level domain annotation from M2M. B-NP (B-Number of People), B-RN (B-Restaurant Name), O (Others)}
% \end{wrapfigure}

\section{Introduction}
\begin{figure}[t]
    \centering
    \includegraphics[width=1.0\linewidth]{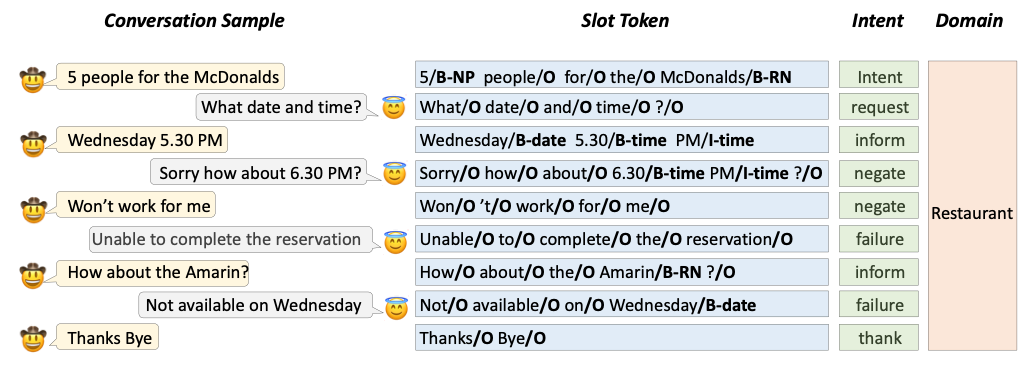}
    \caption{An example of conversations with WSs, SIs, and CD annotation from M2M. B-NP (B-Number of People), B-RN (B-Restaurant Name), O (Others).}
    \label{fig:m2m-sample}
    \vspace{-3mm}
\end{figure}

Natural Language Understanding (NLU) within the realm of Natural Language Processing (NLP) explores the mechanisms through which computers comprehend human language. Developing a hierarchical semantic framework encompassing domain, intent, and slot has become pivotal in representing the meaning embedded in natural language \cite{weld2022survey}.
We present a conversation example that shows the way of annotation for WSs, SI, and CD from the M2M dataset in Figure \ref{fig:m2m-sample}. 
The dialogue consists of a total of 9 turns, and each turn includes WS tokens and SI information, and the dialogue corresponds to one domain, `restaurant'.

\begin{table*}[t]
\centering
\setlength\tabcolsep{1 pt}
\scriptsize
    \begin{tabular}{l|c|c|c|c|c|l}
        \hline
        \multicolumn{1}{c|}{\multirow{2}{*}{\textbf{Model}}} & \multirow{2}{*}{~~\textbf{Year}~~} & {~~\textbf{Word}~~} & {~~\textbf{Sentence}~~} & {~~\textbf{Document}~~} & \multirow{2}{*}{\textbf{Dialogue Type}} & \multirow{2}{*}{~~\textbf{Joint Integration}~~} \\
        % \cline{3-5}
        & & \textbf{(Slot)} & \textbf{(Intent)} & \textbf{(Domain)} & & \\
        \hline
        SeqSeq \citet{Liu2016} & 2016 & $\bigcirc$ & $\bigcirc$ & $\times$ & Single-Turn & BiRNN + Attention \\
        SDEN \citet{Bapna2017} & 2017 & $\bigcirc$ & $\bigcirc$ & $\bigcirc$ & Multi-Turn & BiRNN + Memory Network \\
        Slot-Gated \citet{Goo2018} & 2018 & $\bigcirc$ & $\bigcirc$ & $\times$ & Single-Turn & BiLSTM + Slot Gate\\
        BLSTM+attention \citet{Tingting2019} & 2019 & $\bigcirc$ & $\bigcirc$ & $\times$ & Single-Turn & BiLSTM + Attention \\
        STD \citet{Jiang2021} & 2021 & $\bigcirc$ & $\bigcirc$ & $\times$ & Single-Turn & Transformer + One-teacher KD \\
        SDJN \citet{Chen2022:ICASSP} & 2022 & $\bigcirc$ & $\bigcirc$ & $\times$ & Single-Turn & BiLSTM + self KD \\
        XAI Attention \citet{Gunaratna2022} & 2022 & $\bigcirc$ & $\bigcirc$ & $\times$ & Multi-Turn & eXplainable AI \\
        Tri-level JNLU \citet{Weld2023} & 2023 & $\bigcirc$ & $\bigcirc$ & $\bigcirc$ & Multi-Turn & Cross Transformer \\
        PACL \citet{chen2024two} & 2024 & $\bigcirc$ & $\bigcirc$ & $\times$ & Multi-Turn & Contrastive Learning + Attention \\
        BiJM \citet{luo2024bi} & 2024 & $\bigcirc$ & $\bigcirc$ & $\times$ & Single-Turn & Transformer + Enhance Layer \\
        \hline
        \textbf{Ours} & 2024 & $\bigcirc$ & $\bigcirc$ & $\bigcirc$ & Multi-Turn & Multi-teacher KD \\
        \hline
    \end{tabular}
    \vspace*{-1mm}
    \caption{Summary of existing joint NLU models and ours. Word, Sentence, and Document columns indicate whether the relevant information is used for joint integration. KD refers to knowledge distillation. The complete set of summary tables is detailed in Appendix \ref{ap:rel}.}
    \label{tab:related}
    \vspace*{-4mm}
\end{table*}

Large Language Models (LLM) have received much attention in generating human-like text based on user prompts. However, they are still limited when it comes to deeper communication and diverse key information\footnote{We tested NLU benchmarks with several LLMs, including LLaMa2, LLama3.1, Gemma, QWen2, GPT3.5, and GPT4o visualised in Appendix \ref{ap:prompt-quant} and \ref{ap:prompt-qual}.}. Hence, we investigate how to improve the state-of-the-art existing NLU techniques. While existing NLU literature predominantly concentrates on single-turn utterances within a single domain, recent advancements in multi-turn datasets have paved the way for annotations at the dialogue level, spanning across diverse domains.
Interpreting more extended and intricate conversations with multiple turns necessitates understanding the ongoing context and retaining previously gathered information. Traditional NLU involves mapping single utterances to a dual-level semantic structure, encompassing SI and WS labels. With real-life conversations extending across multiple turns, there is an evident demand for research incorporating dialogue history, as demonstrated by improved performance through dialogue context. The challenge extends beyond dual-level understanding to encompass a three-level comprehension: SI, WS, and CD classification. However, researchers encounter challenges in handling all aspects of multi-turn dialogue conversations through a single unified NLU model, due to computational complexity and a lack of distillability of multi-level knowledge.

This paper introduces a novel multi-level multi-teacher knowledge distillation model to enhance NLU understanding in multi-turn dialogues, leveraging diverse levels of knowledge embedded in these datasets. Notably, our model is the pioneering approach in multi-teacher knowledge distillation, catering to distinct facets of knowledge within a dialogue. To achieve this, our approach involves the construction of teachers at different levels, specifically focusing on SI detection, WS filling, and CD classification. We fine-tune these multi-level teachers to acquire the relevant knowledge and combine these to educate the student model in dialogue tasks facilitated by novel multi-level teacher loss functions.
There are two major contributions: 

\noindent 1) We introduce a novel multi-level, multi-teacher knowledge distillation model to enhance multi-turn NLU. It outperforms widely-used multi-NLU datasets, producing superior performance in all intent detection, slot filling, and domain classification, even compared with the LLMs. 

\noindent 2) We introduce multi-level teacher loss functions, shedding light on their impact within the multi-teacher knowledge distillation and guiding a student model.

\section{Related works}
There is a large body of NLU modelling literature, and we briefly introduce the joint NLU models and knowledge distillation models. A summary of these models and our model are in Table \ref{tab:related}.

\textbf{Natural Language Understanding} 
Early works addressed slot filling and intent detection separately. Current research commonly employs joint models with transfer learning \cite{Rongali2021,Mhamdi2021}, where fine-tuning language models \cite{Dao2021,Abro2022,Mei2023} enhances generalisation by leveraging high-quality representations. Typically, intent is classified through the \texttt{[CLS]} token and slots through individual token embeddings \cite{Chen2019,han2021bi,Heo2022,luo2024bi}. Another transfer learning strategy is knowledge distillation, where a smaller student model learns from a larger teacher model, often using self-distillation \cite{Chen2022:ICASSP,Cheng2023}. However, these methods primarily address single-turn dialogues or use only one teacher model. Multi-turn dialogues benefit from encoding dialogue history, leading to performance gains \cite{Bapna2017,Weld2023,Wu2023,Tu2023,chen2024two}. 
Our model is the first to employ multi-teacher knowledge distillation for multi-turn NLU. Distinct teachers specialise in intent classification, slot filling, and domain classification, thus effectively distilling multi-level knowledge. To the best of our knowledge, no previous work has explored multi-teacher distillation in multi-turn dialogue NLU.

\textbf{Knowledge Distillation}
% Knowledge Distillation (KD) defines a learning approach involving using a well-trained network of teachers to guide the training of a student network for various tasks. Early KD transfers knowledge from one teacher to one student model \cite{Hinton2015}. 
% Multi-teacher KD, inspired by ensemble learning, aims to enhance performance by incorporating knowledge from multiple teacher models into a student model \cite{Wu2021,Yuan2021,Jung2023,Wang2021:WWW,Huang2023:ICASSP,Amirkhani2021,Mirzadeh2020,Son2021}.
% It was common for KD to use multiple teachers to learn the same domain, regardless of whether the teacher had the same or a different architecture. 
% Recent methods have been proposed in which each teacher learns a different domain and imparts knowledge to the student \cite{Pan2021,Ji2023}. 
% Additionally, methods for learning different modalities have also been proposed, which are classified into two types: the teacher and student learn different modalities, a concept known as cross-modality \cite{Kong2019,Ni2022} and the teacher learns different modalities, and the student receives all modalities \cite{Jin2021}.
%Reduction
Knowledge Distillation (KD) defines a framework where a well-trained teacher network guides the training of a student network for various tasks. Traditional KD employs a single teacher to train one student model \cite{Hinton2015}. Multi-teacher KD, inspired by ensemble learning, integrates knowledge from multiple teachers to enhance student model performance \cite{Wu2021,Yuan2021,Jung2023,Wang2021:WWW,Huang2023,Amirkhani2021,Mirzadeh2020,Son2021}. Typically, multiple teachers focus on the same domain, regardless of architecture. Recent approaches involve each teacher specialising in different domains and imparting domain-specific knowledge to the student \cite{Pan2021,Ji2023}. Cross-modality methods have been explored: either the teacher and student learn different modalities \cite{Kong2019,Ni2022}, or the teacher learns multiple modalities, and the student receives all modalities \cite{Jin2021}.

\begin{figure}[t]
    \centering
    \includegraphics[width=1.0\linewidth]{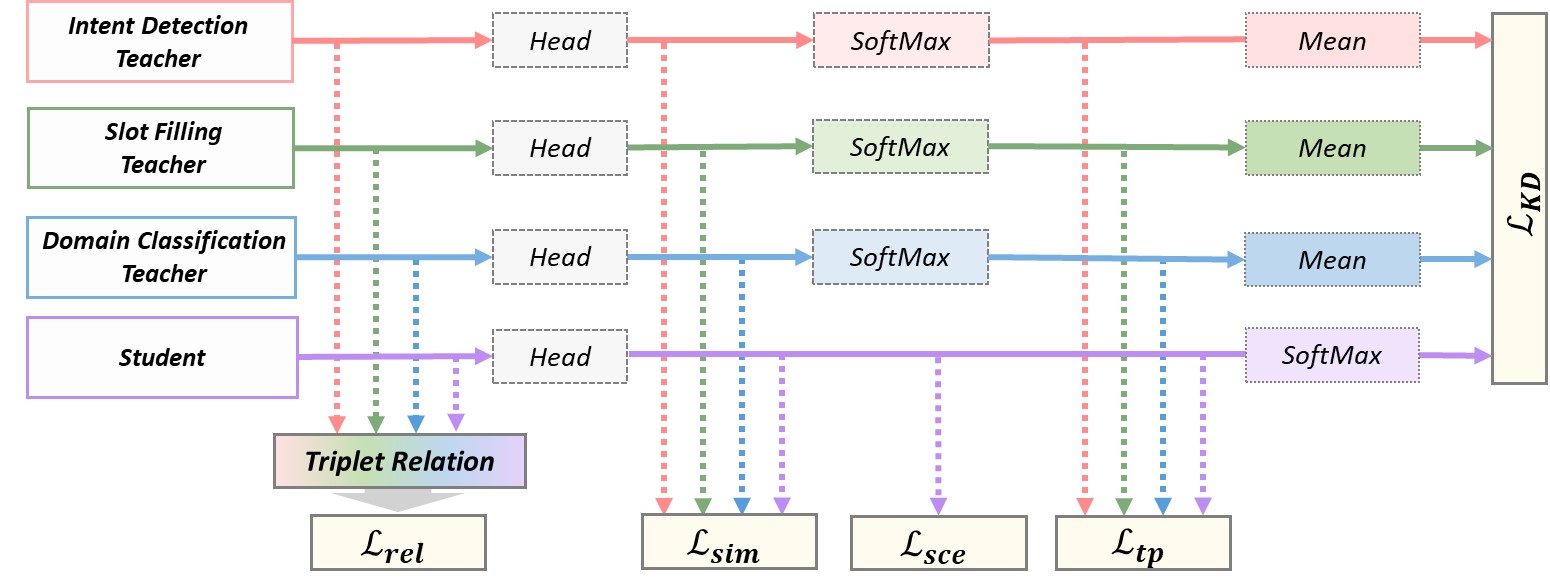}
    \caption{The proposed multi-level teacher knowledge distillation framework for the multi-turn NLU task. Note that we applied three multi-level teachers: Intent Detection, Slot Filling, and Domain Classification. In this framework, we conduct diverse Loss objectives, including $\mathcal L_{rel}$, $\mathcal L_{sim}$, $\mathcal L_{sce}$, $\mathcal L_{tp}$ and $\mathcal L_{KD}$, which represent relation loss, similarity loss, student cross-entropy loss, teacher prediction supervise loss, and Kullback-Leibler Divergence loss, respectively.}
    \label{fig:framework}
\end{figure}

\section{MIDAS}
  We propose a new multi-level dialogue teacher knowledge distillation framework, MIDAS, that trains the student model $S$ with multi-level teachers to enhance the NLU capabilities. We have three multi-level dialogue knowledge teachers, including intent detection, slot filling and domain topic classification. To achieve this, we initially construct teachers with distinct levels of dialogue knowledge, denoted as $T=\{T_{ID},T_{SF},T_{DC}\}$, where $T$ is the set of teacher models, and $ID$, $SF$, and $DC$ correspond to Intent Detection, Slot Filling, Domain Classification. Then, we fine-tune the teacher models $T$ to acquire knowledge from each task. 
  % Finally, a combination of all three multi-level teachers $T$ is employed to instruct the student model $S$ in dialogue tasks using our newly proposed combination of multi-teacher knowledge loss objectives. The comprehensive architecture is depicted in Figure \ref{fig:framework}.
On top of simple KD losses, we also introduce two novel loss functions, relation loss $\mathcal L_{rel}$ and teacher prediction supervised loss $\mathcal L_{tp}$, specifically designed for this domain. These facilitate knowledge transfer from multi-level teachers to the student model. The overall architecture is illustrated in Figure \ref{fig:framework}.

% In this paper, We propose to train the student model $S$ with multi-level teachers to enhance the NLU ability of the student model with the help of different levels of knowledge in the dialogue datasets. First, we construct the teachers of different levels of dialogue knowledge, represented by $T=\{T_{ID},T_{DC},T_{SF}\}$, where $T$ represents the teacher model set,  $ID$, $DC$ and $SF$ represents Intent Detection, Domain Classification, and Slot Filling task respectively. After that, we fine-tune the teacher models $T$ to make them learn the knowledge  from each task. Finally, we use a combination of multi-level teachers $T$ to teach the student model $S$ in the dialogue tasks with the help of our proposed new multi-teacher knowledge loss. The overall architecture is shown in Figure \ref{fig:quantum_diagram}

\subsection{Multi-Level teacher construction}
\label{subsec:mtc}
We first construct the teachers of different dialogue document component understanding levels, including WS, SI, and CD knowledge. The inputs for all teachers consist of utterances from each turn in dialogue datasets, denoted by $X^i={x^i_1,x^i_2,...,x^i_l}$, where $X^i$ represents the $i_{th}$ utterance in the entire dataset, $l$ is the length of the utterance, and $x^i_l$ signifies a word in the utterance. 

\noindent\textbf{1) Word-level teacher} $T_{SF}$ predicts the slot type for each word, providing knowledge to the student model about key slots in the dialogue. The output of $T_{SF}$ is $\hat{Y}^i_{SF}={\hat{Y}^i_{SF,1},\hat{Y}^i_{SF,2},...,\hat{Y}^i_{SF,l}}$, representing the predicted slot types for each word, where $\hat{Y}^i_{SF,l} \in {0,1,...,k_{SF}-1}$, and $k_{SF}$ is the number of slot types. 

\noindent\textbf{2) Sentence-level teacher} $T_{ID}$ predicts the intent of the utterance, aiding the student model in comprehending the overall intent of each turn. The prediction of $T_{ID}$ is symbolised as $\hat{Y}^i_{ID}$, where $\hat{Y}^i_{ID} \in {0,1,...,k_{ID}-1}$, and $k_{ID}$ represents the number of intents in the dataset.

\noindent\textbf{3) Conversation Document-level teacher} $T_{DC}$ forecasts the dialogue's domain, providing knowledge to classify it and understand its background knowledge. The prediction of $T_{DC}$ is indicated as $\hat{Y}^i_{DC}$, where $\hat{Y}^i_{DC} \in {0,1,...,k_{DC}-1}$, and $k_{DC}$ denotes the number of domains in the dataset.

Using these three levels of teachers, our objective is to instruct the student model to comprehend dialogues from multiple perspectives, incorporating WS, SI, and CD background knowledge. By doing so, we enhance the student model's grasp of dialogues across various levels. There are two primary reasons for utilising multi-level dialogue knowledge teachers to train a student. 
First, individually deploying a pre-trained model for each task consumes more computational resources, and some machines may not support running multiple pre-trained models. Instead, the knowledge distillation process leads to more robust models and is resistant to adversarial attacks. Incorporating soft targets from the teacher model can help the student model learn smoother decision boundaries.
Secondly, we posit that diverse levels of knowledge derived from multi-turn conversation understanding datasets can enhance the comprehension of each specific natural language understanding task, surpassing the benefits of learning from single-level dialogue knowledge. 
Note that we use pre-trained models as the foundational structure for our teachers. After experimenting with various backbones, we determined that BERT yields one of the best results overall, as detailed in Section \ref{subsec:teacher}. These pre-trained models undergo fine-tuning using specific data for each level, resulting in distinct teachers with expertise in intent detection, slot filling, and document classification. Pre-trained models, having been trained on extensive text data, exhibit the capacity to transfer knowledge effectively. Ultimately, we leverage the collective knowledge of these refined teachers to train the student model comprehensively.

\subsection{Multi-level Teacher Fine-tuning}

We perform separate fine-tuning of pre-trained models on $ID$, $SF$, and $DC$ tasks. This yields multi-level teachers, $T_{ID}$, $T_{SF}$, and $T_{DC}$ respectively, corresponding to sentence-level, word-level and sentence-level knowledge, respectively. Each pre-trained model specialises in learning knowledge at one specific level from the dialogue datasets, resulting in teachers possessing different levels of dialogue document component understanding. It's important to note that each teacher focuses on one level of dialogue knowledge. 
This approach is motivated by two factors.
First, learning knowledge from a single task is less complex than incorporating knowledge from all tasks, simplifying the fine-tuning of pre-trained models. Secondly, instead of burdening a single model with the challenge of mastering knowledge from all aspects of dialogues, each teacher focuses on a specific level of understanding, such as WS filling, SI detection, or CD classification. For each task, we consolidate data from two datasets (MultiWOZ and M2M) by merging split and corresponding label sets. For example, the training set for fine-tuning includes data from both datasets. We apply cross-entropy loss and fine-tune the pre-trained models for a fixed number of epochs, utilising the checkpoint from the last epoch as the teacher model. The process is described as follows:
\begin{eqnarray}
\begin{split}
\label{eq:1}
X_{j,tr} =&X^1_{j,tr},\dots,X^{N_{M2M}}_{j,tr},X^1_{j,tr},\dots,\nonumber\\
&X^{N_{MWOZ}}_{j,tr},\nonumber\\
\mathcal L_{tce}=&cross\_entropy(T_{j}(X_{j,tr}),Y_j), \nonumber\\
j \in&\{DC,ID,SF\} \nonumber
\end{split}
\end{eqnarray}
where $N_{M2M}$ and $N_{MultiWoz}$ are the number of training samples, and $Y_j$ is the ground truth. 

\subsection{Multi teachers knowledge distillation}
\label{subsec:mtkd}
Following the acquisition of multi-level teachers $T$, we employ a blend of these teachers to instruct the student model $S$ through multi-teacher knowledge distillation. The combination of teachers comprises different levels, such as \{BERT-Base ID, BERT-Base SF, BERT-Base DC\}, \{BERT-Base ID, RoBERTa-Base DC, and LLaMa2-7b SF\}. The student model undergoes separate training for each task, enabling it to grasp the intricacies of individual tasks with the assistance of diverse levels of dialogue knowledge.

% After obtaining multi-level teachers $T$, we use the combination of multi-level teachers to train the student model $S$ using multi-teacher knowledge distillation. The teachers in the combination come from different levels, e.g., BERT-ID, RoBERTa-DC and LLaMA-SF. We train the student model on each task separately to make it learn each task with the help of different levels of dialogue knowledge. 
We delve into the exploration and introduction of five distinct loss functions to assess their efficacy within the MIDAS. We propose relation loss and teacher prediction supervised loss, specifically designed for multi-level knowledge distillation. Furthermore, with MIDAS, we explore three previously established losses tailored for multi-level teacher integration. These encompass Kullback-Leibler Divergence loss, Similarity loss, and Student Cross Entropy loss, each designed to enhance the learning dynamics in the context of multi-level knowledge distillation.

The two newly proposed losses, relation loss and teacher prediction supervised loss, are designed to better handle the issue of knowledge conflicts between different teachers in multi-level teacher knowledge distillation. Relation loss uses a voting mechanism, guiding the student to learn inter-sample relationships from the majority of teachers. Teacher prediction supervised loss leverages the class probability distributions from each teacher as pseudo-labels, aligning the students’ predictions with the dominant teacher perspectives. The details of the two losses are shown in the following:

\textbf{Relation loss $\mathcal L_{rel}$}: During training, for each batch of data, triplets are randomly generated, and the internal relations of the triplets are determined by aggregating the votes from the combination of teacher models. Employing TripletMarginLoss \cite{trapleletmargin}, the student model learns internal relations among the batch data, aligning its understanding with that of the teacher models and ensuring a consistent perspective on the dataset.
\begin{eqnarray}
\mathcal L_{rel}=\frac{1}{N}\sum_i^N{TripletMarginLoss(\mathcal{T}_i)}\nonumber
\end{eqnarray}
where $N$ is the batch size, and triplet $\mathcal{T}$ is generated by and articulated in Algorithm \ref{alg:algorithm}, see the details in Section \ref{sec:role of loss}.

\textbf{Teacher prediction supervised loss $\mathcal L_{tp}$}: In addition to utilising the ground truth for each task, we incorporate the predictions made by the teacher models as pseudo-labels to facilitate the training of the student model. We employ the probabilities assigned by the teacher models for each class, ensuring that the student comprehensively acquires the knowledge embedded in the teacher models.

\vspace{-5mm}
\begin{eqnarray}
\mathcal L_{tp}=\sum_j^{n_T}{cross\_entropy(v_{s},P_{j})}\nonumber
\end{eqnarray}
\vspace{-5mm}

% We incorporate Kullback–Leibler Divergence loss and Student Cross Entropy loss, widely utilised knowledge distillation tools. Furthermore, with MIDAS, we explore three specific types of losses tailored for multi-level teacher integration. These encompass relation loss, similarity loss, and teacher-prediction supervised loss, each designed to enhance the learning dynamics in the context of multi-level knowledge distillation.

% modify here
% \textbf{Kullback–Leibler Divergence(KD) Loss $\mathcal L_{KD}$}: We compute the KD loss \citet{hinton2015distilling} by comparing the mean probabilities generated by the combination of teacher models with the probabilities derived from the student model. This process aims to align their probability distributions, facilitating the learning of the student model from multiple teachers.

In addition to the two newly proposed losses, we also introduce the previously established losses, briefly described as follows:

\textbf{Kullback–Leibler divergence(KD) loss $\mathcal L_{KD}$}: We compute the KD loss \cite{hinton2015distilling} by comparing the mean probabilities generated by the combination of teacher models with the probabilities derived from the student model, i.e., $\mathcal L_{KD}=KLDivLoss(\frac{1}{n_T}{\sum}_j^{n_T}{P_{j}},P_{s})$, where $P_j=softmax(T_j(X)),P_s=softmax(S(X))$, and $n_T$ is the number of teachers. It helps us to align the prediction probability distributions between the student model and the teacher models.

% We use KD loss to align the prediction probability distributions between the student model and the teacher models.
% facilitating the learning of the student model from multiple teachers.
% \begin{eqnarray}
% \begin{split}
% \mathcal L_{KD}=&KLDivLoss(\frac{1}{n_T}{\sum}_j^{n_T}{P_{j}},P_{s}),\nonumber\\
% P_j=&softmax(T_j(X)),P_s=softmax(S(X))\nonumber
% \end{split}
% \end{eqnarray}
% where $n_T$ is the number of teachers.

% modify here
% \textbf{Student Cross Entropy(SCE) Loss $\mathcal L_{sce}$}: This loss function is computed by comparing the student model's predictions with the ground truth of each task. By employing the cross-entropy loss, the student model receives direct supervisory signals, aiding in its learning process.

\textbf{Similarity loss $\mathcal L_{sim}$}: The similarity loss is computed by maximising the logit similarities between the student model and teacher models. With this, the student model can learn the knowledge from the teacher models in the feature space, not only the prediction probabilities. The loss equation is:

\vspace{-5mm}
\begin{eqnarray}
% \vspace{-10mm}
\begin{split}
\label{lsim}
\mathcal L_{sim}=&-\sum_j^{n_T}{\mathcal{F}_{sim}(v_j,v_{s})}, \\ v_j=&T_j(X), \, v_s=S(X)
\end{split}
% \vspace{-3mm}
\end{eqnarray}
\vspace{-3mm}

Here, $\mathcal{F}_{sim}$ denotes the similarity function, and $v_j$ represents the teacher logit.

\textbf{Student cross entropy(SCE) loss $\mathcal L_{sce}$}: This loss function is computed by comparing the student model's predictions with the ground truth of each task, i.e., $\mathcal L_{sce}=cross\_entropy(v_{s},Y_{true}), \, v_s=S(X)$, where $v_s$ represents the student logits. It's the basis loss in our supervised learning task.

We experiment with diverse combinations of the aforementioned loss functions to assess their impact on the student's performance across various datasets and NLU tasks. The summary of the loss function is described in Section \ref{sec:role of loss} and the detailed result analysis can be found in Section \ref{subsec:lossanalysis}.

\section{Experimental setup}
\subsection{Datasets and Baselines}
We focus on multi-turn dialogue analysis in the dialogue state tracking (DST) domain, which consists of all three NLU tasks, including intent detection $ID$, slot filling $SF$, and domain(topic) classification $DC$. Following by \cite{Weld2023}, we utilise two widely-used benchmark datasets in multi-turn dialogue NLU: Multi-Domain Wizard-of-Oz 2.2 (MWOZ) and Machines Talking To Machines (M2M) in the DST field. 
Details for datasets are shown below:

\begin{itemize}
    \item \textbf{MWOZ} \cite{Zang2020} is specifically designed for Dialogue State Tracking (DST) and adopts the conventional human-vs-human Wizard of Oz approach across diverse domains, including attraction, bus, hospital, hotel, police, restaurant, taxi, and train. It incorporates 30 slot types and 11 intent types. The dataset comprises 8,437 dialogues, with an average of 5.68 turns per dialogue and 14.07 tokens per turn. Following by \cite{Liu2016,Goo2018,Weld2023}, we do not consider any multi-label samples but utilise the data with a single domain and intent.

    \item \textbf{M2M} \cite{Liu2018} is introduced with virtual agents and user-generated interactions to emulate goal-directed conversations through paraphrasing with templated utterances. M2M has movies and restaurant domains. The slots and intents are categorical, with 21 slot types and 15 intent types. The dataset comprises 1,500 dialogues, with an average of 9.86 turns per dialogue and 8.25 tokens per turn.
\end{itemize}

We adopted the three published results as baselines,  \textbf{SeqSeq}\cite{Liu2016}, \textbf{Slot-Gated} \cite{Goo2018} and \textbf{Tri-level JNLU} \cite{Weld2023}. Additionally, we fine-tuned the pre-trained language models commonly used in the NLU, \textbf{BERT-Base}\footnote{\url{https://huggingface.co/bert-base-uncased}},  \textbf{RoBERTa-Base}\footnote{\url{https://huggingface.co/roberta-base}}, \textbf{ALBERT-Base}\footnote{\url{https://huggingface.co/albert-base-v2}}. The details are shown in Appendix \ref{ap:baselines}.

\subsection{The role of each loss function}
\label{sec:role of loss}
The 5 types of losses are utilized for training the student model, each playing a distinct role: 

1) $\bm{L_{kd}}$: This loss facilitates the transfer of knowledge from the teacher models to the student model, enabling the student to mimic the general behaviour of multiple teachers. 2) $\bm{L_{rel}}$: This loss is designed to capture the relationships between different samples in the input data. It helps to align the student's understanding with that of the teacher models and ensures a consistent perspective on the dataset.
3) $\bm{L_{sim}}$: This loss encourages the student model to generate outputs similar to those of the teacher models in terms of their overall structure and distribution. It helps to maintain consistency between the student and teacher predictions.
4) $\bm{L_{sce}}$: This loss function is the fundamental mechanism for training the student model. It entails the student learning to predict the correct labels associated with the input data.
5) $\bm{L_{tp}}$: This loss leverages the predictions of the teacher models to provide additional supervision signals to the student model. It helps to guide the student towards making predictions that align with those of the teachers.

% The proposed model was compared with several baselines as follow. 
% Due to the limited number of baselines available for multi-turn NLU, 
% four pre-trained models and three related works are adopted as baselines.
% \begin{itemize}
%     \item \textbf{BERT-Base}\footnote{\url{https://huggingface.co/bert-base-uncased}} and \textbf{BERT-Large}\footnote{\url{https://huggingface.co/bert-large-uncased}} are representative transformer encoder-based language models. 
%     BERT-Base refers to the basic model, while BERT-Large represents a larger model with more parameters than the base model.
%     \item \textbf{ALBERT-Base}\footnote{\url{https://huggingface.co/albert-base-v2}} and \textbf{ALBERT-Large}\footnote{\url{https://huggingface.co/albert-large-v2}} are BERT-based models that demonstrated superior performance with reducing the model size through factorized embedding parameterization and cross-layer parameter sharing.
%     \item \textbf{SeqSeq} \citet{Liu2016} is a neural network model based on RNN and attention mechanisms, designed for the joint tasks of intent detection and slot filling.
%     \item \textbf{Slot-Gated} \citet{Goo2018} utilized a slot gate to capture the relationship between intent and slot, aiming for improving semantic understanding through global optimization.
%     \item \textbf{Tri-level JNLU} \citet{Weld2023} is the pioneering model to incorporate domain information in the joint modeling of slot filling and intent detection.
% \end{itemize}

\subsection{Metrics and implementations}
\label{subsec:metrics}
This paper evaluates the performance of baseline models and MIDAS in all three multi-turn dialogue tasks, including $ID$, $SF$, and $DC$ for each dataset. Following by \cite{Liu2016,Goo2018,Weld2023}, the metrics for each task are shown as follows: \textit{Accuracy} for $ID$ and $DC$ and \textit{F1 score} for $SF$. Accuracy is the most commonly used metric for $ID$ as determining the intent of an utterance is typically framed as a classification task. Accuracy is calculated as the ratio of correct predictions to the total number of tests. $DC$ also employs accuracy as it is a classification task. On the other hand, $SF$ employs F1 score. F1 score is directed towards assessing the prediction effectiveness for slot tokens. It computes an F1 score for each class and determines the token-based micro-averaged F1 score across all classes.
% \subsection{Implementation}
% \label{sec:implementation}

We introduce some implementation details in this section and the complete details in Appendix \ref{ap:imp-detail}. For \textbf{Multi-teacher fine-tuning}, we use BERT-Base, RoBERTa-Base and LLaMa2-7b\footnote{\url{https://huggingface.co/meta-LLaMA/LLaMA-2-7b}} as the teacher backbones and fine-tune them on each task. For fine-tuning LLaMa2-7b, we adopt an unmask strategy used in \cite{li2023label}. We use AdamW \cite{loshchilov2018decoupled} and CrossEntropy loss to fine-tune the pre-trained models for 3 epochs. The learning rate is 5e-5 and is warm-uped linearly from 0 to 5e-5 during the first 10\% training steps. The batch size is 32. For \textbf{Multi-level Distillation}, we use AdamW and the aforementioned losses to train the student with multi-level teachers. We use Squared Euclidean distance in algorithm \ref{alg:algorithm} and cosine similarity in equation \ref{lsim}. For the combination of these losses, we sum them without any weight. We use the same optimiser, learning rate, warm-up strategy, and batch size as the one used in teacher fine-tuning, and use a vanilla Transformer encoder as a student.

% as shown in Table \ref{student model structure}

% \begin{table}
%     \centering
%     \begin{tabular}{ll}
%         \hline
%         Layer   & 6 \\
%         Head    & 8 \\
%         Embedding Dim. & 768  \\
%         Hidden Dim. & 2048 \\
%         Dropout  & 0.3 \\
%         \hline
%     \end{tabular}1
%     \caption{Vanilla Transformer Details}
%     \label{student model structure}
% \end{table}

\section{Results}
\subsection{Overall performance}
\label{sec:result}

We compare MIDAS with fine-tuned PLM baselines and published pioneering model results for two mainstream multi-turn natural language understanding tasks, Intent Detection and Slot Filling, with the same evaluation setup. Table \ref{tab:exp} shows that MIDAS remarkably outperforms other baselines. 
To demonstrate the improvement achieved through MIDAS, we conduct experiments utilising two widely recognised multi-turn dialogue understanding datasets, MWOZ and M2M. Note that all baselines and MIDAS are individually fine-tuned for each task. 
As detailed in Section \ref{subsec:mtc}, our approach involves the utilisation of pre-trained models, BERT or RoBERTa, for the fine-tuning of our three multi-level teacher models. These teachers encompass $ID$, $SF$, and $DC$. It is important to highlight that MIDAS undergoes knowledge distillation from three distinct multi-level teachers, each specialising in SI, word token-level slot, and CD topic. Thus, Table \ref{tab:exp} shows the results \textit{MIDAS (BERT)} and \textit{MIDAS (RoBERTa)} that all teachers are constructed using either the BERT or RoBERTa architecture.

\begin{table}[t]
\scriptsize
\setlength\extrarowheight{1pt}
\setlength\tabcolsep{2.0 pt}
\centering
    % \vspace{-5mm}
    \begin{tabular}[b]{l|c|c|c|c|c|c}
        \hline
        \multirow{3}{*}{} & \multicolumn{2}{c|}{\cellcolor{YellowGreen!25}{\small\textbf{ID}}} & \multicolumn{2}{c|}{\cellcolor{YellowGreen!25}{\small\textbf{SF}}} & \multicolumn{2}{c}{\small\textbf{DC}} \\

        % \multirow{3}{*}{} & \multicolumn{2}{c|}{\small\textbf{ID}} & \multicolumn{2}{c|}{\small\textbf{SF}} & \multicolumn{2}{c}{\small\textbf{DC}} \\
        \cline{2-7}
        & \textbf{MWOZ} & \textbf{M2M} & \textbf{MWOZ} & \textbf{M2M} & \textbf{MWOZ} & \textbf{M2M}  \\
        & \textbf{(ACC)} & \textbf{(ACC)} & \textbf{(F1)} & \textbf{(F1)} & \textbf{(ACC)} & \textbf{(ACC)} \\
        \hline 
        \hline 
        BERT-Base       &   0.6534  &   0.8675  &   0.9218  &   0.8543  &   0.8667 &    0.8923 \\
        RoBERTa-Base    &   0.8424  &   0.9252  &   0.9748  &   0.9132  &   0.8675 &  0.8909 \\
        ALBERT-Base     &   0.6531  &   0.8654   &   0.9187   &   0.8542   &   0.8694 &   0.8919 \\
        SeqSeq          &   0.6641   &   0.9250   &   0.8543   &   0.9172   &   -      &   - \\
        Slot-Gated      &   0.6883   &   0.9327   &   0.8776   &   0.9279   &   -      &   - \\
        Tri-level JNLU  &   0.7849   &   0.9419   &   0.9798   &   0.9302   &   0.2572 &   0.8938 \\
        \hline
        \textbf{MIDAS (BERT)}&   0.8464   &   \textbf{0.9427}   &   \textbf{0.9928}   &   \textbf{0.9856} &   0.8793   &   \textbf{0.8952} \\
        \textbf{MIDAS (RoBERTa)} &   \textbf{0.8502}   &   0.9377   &   \textbf{0.9928}   &   0.9813 &   \textbf{0.8816}   &   0.8945 \\
        \hline
    \end{tabular}
    % \vspace{-3mm}
    \caption{The comparison of the MIDAS with baselines. $ID$, $SF$ and $DC$ indicate intent detection, slot filling and domain classification, respectively, as mentioned in Section \ref{subsec:metrics}. ACC and F1 stand for accuracy and micro F1, respectively, and scores in bold indicate leadership among the metrics.}
    \label{tab:exp}
    \vspace{-2mm}
\end{table}

Two versions of MIDAS exhibit superior performance across both datasets, presenting outstanding outcomes with a slot-filling error rate below 2\%. While the RoBERTa-Base model demonstrates superiority in MWOZ, the BERT-Base model excels in M2M. What should be noted is the performance difference between these models is not substantial, with both consistently outperforming other baseline models. In $ID$ and $SF$ tasks, MIDAS showcases notably higher performance compared to baselines. 
We also conduct experiments on the $DC$ task with the same datasets to better compare the differences between MIDAS and other PLM baselines. 
However, while surpassing BERT-Base and ALBERT-Base, the performance difference is marginal. We assume that this discrepancy is attributed to the small number of the domain class. In contrast to other baseline models, Seq2Seq and Slot-Gated lack a structure incorporating domain information, making them unable to assess domain classification performance.
%modify end

Overall, the observation highlights that bolstering multi-level conversation knowledge substantially improves the comprehension of each Natural Language Understanding (NLU) task. Specifically, enhancing results in $ID$ is achievable by refining a student model through the distillation of multi-level knowledge, encompassing SI, WSs, and CD knowledge. The following two sections (Sections \ref{subsec:teacher} and \ref{subsec:comb}) delve into a more comprehensive exploration of multi-level teacher models and the combination of multi-level teachers.

\subsection{Effect of pretrained model for teachers}
\label{subsec:teacher}
We then evaluate the efficacy of different pre-trained models for our multi-level teachers. As detailed in Section \ref{sec:result} and illustrated in Table \ref{tab:exp}, we employed all three multi-level teachers ($ID$, $SF$, and $DC$) based on BERT and/or RoBERTa, resulting in a superb performance. In this section, we investigate how various pre-trained language models can impact the knowledge distillation ability of our multi-level teachers in instructing the student model. In addition to using BERT or RoBERTa, we also incorporate LLaMa2-7b, a decoder-only based pre-trained model, into our analysis. 

Table \ref{tab:ablation} shows the results of the effectiveness of using various pre-trained models as base models for all three multi-level teachers\footnote{Note that the \textit{MIDAS (BERT)} and \textit{MIDAS (RoBERTa)} models are identical to those presented in the Table \ref{tab:exp}.}. 
Compared to the high-achieving two encoder-based models, BERT and RoBERTa, the MIDAS (LLaMa) multi-level teachers produce lower performance\footnote{Any decoder-only LM produces a similar low performance.}. We assume a decoder-only model like LLaMa is primarily used for generating coherent and contextually relevant text. In contrast, BERT and RoBERTa are encoder-based models that have a deep understanding of context and relationships between words and excel in classification tasks.

In addition to having multi-level teachers using a single pre-trained model, we adopt a mixed type of pre-trained model for preparing multi-level teachers. For instance, we can apply BERT as a pre-trained model for teaching SI knowledge, utilise RoBERTa as a teacher model for WSs, and adopt LLaMa as a CD topic teacher model. Table \ref{tab:ablation} shows that using mixed types of pre-trained teacher models is less effective than employing a consistent single pre-trained model as the teacher. This implies that knowledge distillation from teachers with inconsistencies in their feature spaces may impede the learning process for a single student model.

% In addition to having multi-level teachers using a single pre-trained model, we adopt a mixed type of pre-trained model for preparing multi-level teachers. For instance, we can apply BERT as a pre-trained model for teaching SI knowledge, utilise RoBERTa as a teacher model for WSs, and adopt LLaMA as a CD topic teacher model. 
% We conducted various mixed-type teacher combinations, illustrated in Appendix \ref{ap:teacher-type}. The last row of Table \ref{tab:ablation} presents the `Mixed' type, indicating the best mixed type combinations: BERT for Intent Detection Teacher, RoBERTa for Slot Filling Teacher, and Domain Classification Teacher. 
% Overall, the result shows that using mixed types of pre-trained teacher models is less effective than employing a consistent single pre-trained model as the teacher. This implies that knowledge distillation from teachers with inconsistencies in their feature spaces may impede the learning process for a single student.

\begin{table}[t]
\vspace{-3mm}
\scriptsize
\setlength\extrarowheight{1pt}
\setlength\tabcolsep{2.2pt}
\centering
    \begin{tabular}[b]{l|c|c|c|c|c|c}
        \hline
        \multirow{3}{*}{} & \multicolumn{2}{c|}{\cellcolor{YellowGreen!25}{\small\textbf{ID}}} & \multicolumn{2}{c|}{\cellcolor{YellowGreen!25}{\small\textbf{SF}}} & \multicolumn{2}{c}{\small\textbf{DC}} \\
        \cline{2-7}
        & \textbf{MWOZ} & \textbf{M2M} & \textbf{MWOZ} & \textbf{M2M} & \textbf{MWOZ} & \textbf{M2M}  \\
        & \textbf{(ACC)} & \textbf{(ACC)} & \textbf{(F1)} & \textbf{(F1)} & \textbf{(ACC)} & \textbf{(ACC)} \\
        \hline 
        \hline 
        MIDAS (BERT)&   0.8464   &   \textbf{0.9427}   &   \textbf{0.9928}   &   \textbf{0.9856} &   0.8793   &   \textbf{0.8952} \\[3.5pt]
        MIDAS (RoBERTa) &   \textbf{0.8502}   &   0.9377   &   \textbf{0.9928}   &   0.9813 &   \textbf{0.8816}   &   0.8945 \\[3.5pt]
        MIDAS (LLaMa)     &   0.8403  &   0.9392   &   0.9912   &   0.9833   &   0.8702 &   0.8804 \\[3.5pt]
        % MIDAS (Mixed 1)    &   0.8433   &   0.9401   &   \textbf{0.9931}   &   0.9758   &   0.8753 &   0.8919 \\
        % MIDAS (Mixed 2)    &   \textbf{0.8507}   &   0.9418   &   0.9930   &   0.9801   &   0.8768 &   0.8662 \\
        % MIDAS (Mixed 3)    &   0.8472   &   0.9411   &   0.9839   &   0.9745   &   0.8808 &   0.8929 \\
        % MIDAS (Mixed 4)    &   0.8473   &   0.9401   &   0.9928   &   0.9764   &   0.8769 &   0.8925 \\
        MIDAS (Mixed 1)    &   0.8472   &   0.9411   &   0.9839   &   0.9745   &   0.8808 &   0.8929 \\[3.5pt]
        MIDAS (Mixed 2)    &   0.8473   &   0.9401   &   \textbf{0.9928}   &   0.9764   &   0.8769 &   0.8925 \\[3.5pt]
        \hline
% mixed 4  o-ID-BERT-Base,  o-DC-RoBERTa-Base,  o-SF-RoBERTa-Base
% mixed 3  o-ID-BERT-Base,  o-DC-RoBERTa-Base,  o-SF-LLaMA-Base
% mixed 2  o-ID-BERT-Base	o-SF-BERT-Base	 o-SF-RoBERTa-Base
% mixed 1  o-DC-BERT-Base,  o-SF-BERT-Base,  o-SF-RoBERTa-Base
    \end{tabular}
    \caption{The performance based on the type of teacher models. The \textit{MIDAS (BERT)} and \textit{(RoBERTa)} are identical to those presented in the table \ref{tab:exp} whose all teachers are either BERT or RoBERTa. \textit{MIDAS (LLaMa)} refers to the outcome of utilising the LLaMa2-7b as teacher models of all tasks. The \textit{MIDAS (Mixed 1 and 2)} represents the mixed type teacher combination; Mixed 1: BERT ($ID$), LLaMa ($SF$) and RoBERTa ($DC$); Mixed 2: BERT ($ID$), RoBERTa ($SF$) and RoBERTa ($DC$).}
    \label{tab:ablation}
     \vspace{-4mm}
\end{table}

\begin{table*}[ht]
\setlength\extrarowheight{1pt}
\setlength\tabcolsep{4pt}
\scriptsize
\centering
    \begin{tabular}{c|c|c|c|c|c|c|c|c|c|c}
        \hline

        % \multirow{3}{*}{} & \multicolumn{2}{c|}{\cellcolor{WildStrawberry!25}{\small\textbf{ID}}} & \multicolumn{2}{c|}{\cellcolor{WildStrawberry!25}{\small\textbf{SF}}} & \multicolumn{2}{c}{\cellcolor{YellowGreen!25}{\small\textbf{DC}}} \\
        % \cline{2-7}

        \multirow{3}{*}{{\small $\bm{\mathcal L_{\scriptstyle\textbf{\textit{KD}}}}$}} &  
        
        \multirow{3}{*}{{\small$\bm{\mathcal L_{\scriptstyle\textbf{\textit{sce}}}}$}} &

        \multirow{3}{*}{{\small$\bm{\mathcal L_{\scriptstyle\textbf{\textit{sim}}}}$}} & 
        
        \multirow{3}{*}{{\small$\bm{\mathcal 
        L_{\scriptstyle\textbf{\textit{rel}}}}$}} & \multirow{3}{*}{{\small$\bm{\mathcal L_{\scriptstyle\textbf{\textit{tp}}}}$}} &

        \multicolumn{2}{c|}{\cellcolor{YellowGreen!25}{\small\textbf{ID}}} & \multicolumn{2}{c|}{\cellcolor{YellowGreen!25}{\small \textbf{SF}}} & \multicolumn{2}{c}{{\small\textbf{DC}}} \\
        \cline{6-11}
        &&&&& ~~~\textbf{MWOZ}~~~ & ~~~\textbf{M2M}~~~ & ~~~\textbf{MWOZ}~~~ & ~~~\textbf{M2M}~~~ & ~~~\textbf{MWOZ}~~~ & ~~~\textbf{M2M}~~~  \\
        &&&&& \textbf{(ACC)} & \textbf{(ACC)} & \textbf{(F1)} & \textbf{(F1)} & \textbf{(ACC)} &\textbf{(ACC)} \\[0.2ex]
        \hline
        \hline 
        $\bigcirc$ & $\bigcirc$  & $\bigcirc$ & $\times$ & $\times$ & 0.8429 & \underline{0.9411} & \textbf{0.9928} & \textbf{0.9856} & 0.8750 & 0.8928 \\
        
        $\bigcirc$ & $\bigcirc$ & $\times$ & $\bigcirc$ & $\times$ & 0.8427 & \underline{0.9411} & \textbf{0.9928} & 0.9791 & 0.8750 & 0.8927 \\
        % $\bigcirc$ & $\bigcirc$ & $\ovßoid$ & $\bigcirc$ & $\times$ & 0.8395 & 0.9342 & 0.9924 & 0.9783 & 0.8777 & \underline{0.8937} \\
        \rowcolor{LightCyan}$\bigcirc$ & $\bigcirc$ & $\bigcirc$ & $\bigcirc$ & $\times$ & \textbf{0.8464} & \textbf{0.9427} & \textbf{0.9928} & \underline{0.9850} & \underline{0.8780} & \textbf{0.8952} \\
        
        $\bigcirc$ & $\bigcirc$ & $\bigcirc$ & $\bigcirc$ & $\bigcirc$ & \underline{0.8462} & 0.9373 & \underline{0.9927} & 0.9761 & \textbf{0.8793} & 0.8903 \\
        \hline
    \end{tabular}
    \vspace*{-0mm}
    \caption{The comparison of the diverse loss function combinations. Only BERT is utilised as the teacher model, and the results of RoBERTa are presented in Table \ref{tab:ap-lossab}. The full names of each loss can be found in Section \ref{subsec:mtkd}. We adopt two $\mathcal L_{KD}$ and $\mathcal L_{sce}$ as compulsory knowledge distillation loss and also explore three $\mathcal L_{rel}$, $\mathcal L_{sim}$, and $\mathcal L_{tp}$ for MIDAS. Scores in bold indicate leadership among the metrics, and underlined scores indicate the second-best.}
    \label{tab:lossab}
    \vspace*{-2mm}
\end{table*}

% Table \ref{tab:teacher-ablation} shows the results of using various models as teacher models.
% The BERT-only model and RoBERTa-only model are identical to those presented in the table \ref{tab:exp}. 
% The LLaMA-only model exhibits lower performance than the previous two models.
% LLaMA refers to the outcome of utilizing the LLaMA2-7b model as teacher models.
% LLaMA emphasizes decoding whereas BERT and RoBERTa are encoder-based models which excel in classification tasks.
% The Mixed model represents the outcome of employing BERT as an intent detection teacher and RoBERTa as both a slot filling and domain classification teacher. 
% It shows that using a variety of models is less efficient than employing the same model as the teacher. 
% This finding implies that knowledge distillation from teachers with diverse structures may actually hinder the learning process for a single student.

\subsection{Effect of multi-level combinations}
\label{subsec:comb}
\begin{table}[htp]
\scriptsize
\setlength\extrarowheight{1pt}
\setlength\tabcolsep{3.9pt}
\centering
    \begin{tabular}[b]{l|c|c|c|c|c|c}
        \hline
        \multirow{3}{*}{} & \multicolumn{2}{c|}{\cellcolor{YellowGreen!25}{\small\textbf{ID}}} & \multicolumn{2}{c|}{\cellcolor{YellowGreen!25}{\small\textbf{SF}}} & \multicolumn{2}{c}{\small\textbf{DC}} \\
        \cline{2-7}
        & \textbf{MWOZ} & \textbf{M2M} & \textbf{MWOZ} & \textbf{M2M} & \textbf{MWOZ} & \textbf{M2M}  \\
        & \textbf{(ACC)} & \textbf{(ACC)} & \textbf{(F1)} & \textbf{(F1)} & \textbf{(ACC)} & \textbf{(ACC)} \\
        \hline 
        \hline 
        ID-only     &   0.8406  &   0.9366   &   0.8590   &   0.9684   &   0.7977 &   0.7159 \\
        SF-only    &   0.8310   &   0.9377   &   0.9619   &   0.9718   &   0.2425 &   0.8930 \\
        DC-only    &   0.8408   &   0.9321   &   0.8888   &   0.9534   &   0.6330 &   0.8915 \\
        ID+SF    &   0.8422   &   0.9399   &   0.9924   &   0.9835   &   0.8760 &   0.8939 \\
        ID+DC    &   0.8400   &   0.9292   &   0.9923   &   0.9848   &   0.8756 &   0.8929 \\
        SF+DC    &   0.8376   &   0.9416   &   0.9923   &   0.9825   &   0.8760 &   0.8940 \\
        \hline
        \textbf{ID+SF+DC}    &   \textbf{0.8464}   &   \textbf{0.9427}   &   \textbf{0.9928}   &   \textbf{0.9850} &   \textbf{0.8780}   &   \textbf{0.8952} \\
        \hline
    \end{tabular}
    \caption{The performance based on the type of teacher models. The first column indicates the type of teacher used. For example, \textit{ID+SF+DC} uses all intent classification, slot filling, and domain classification teachers, while \textit{ID-only} uses only the intent classification teacher. Only BERT is utilised as the teacher model.}
    % Results using RoBERTa are presented in Appendix \ref{ap:teacher-comb}
    \label{tab:ablation-tecomb}
     \vspace{0mm}
\end{table}

We explore the impact of incorporating each multi-level teacher ($ID$, $SF$, $DC$) in all three multi-turn dialogue understanding tasks. MIDAS is evaluated with individual teachers ($ID$, $SF$, $DC$), all possible pairs from \{$ID$, $SF$, $DC$\}\footnote{Note that we do not adopt $\mathcal L_{rel}$ since it is not possible to adopt when there are two teachers.}, and then with all three teachers. 
Table \ref{tab:ablation} presents the results for each combination of teacher models for three different dialogue understanding tasks. Note that the table demonstrates the outcome of \textit{MIDAS (BERT)} teachers, and we produce that of \textit{MIDAS (RoBERTa)} in Table \ref{tab:ap-teacher-comb}. The experimental findings highlight that the \textbf{$ID+SF+DC$} combination attains the highest performance, underscoring the advantage of the student model integrating knowledge from all teachers for each task.

% Table \ref{tab:teacher-comb} presents the results for each combination of teacher models.
% The first three results represent using only one teacher for each task, respectively.
% The next three results represents using only two out of the three teachers for the tasks. 
% The last result shows using all teachers for the three tasks.
% The experimental results demonstrate that the student model benefits from incorporating all the knowledge from the teachers for each task.
% Furthermore, when there are two teachers, it is not possible to use $\mathcal L_{rel}$. 
% The last one shows the best performance, which can utilize $\mathcal L_{rel}$.

\subsection{Effect of knowledge distillation loss}
\label{subsec:lossanalysis}
As mentioned in Section \ref{subsec:mtkd}, we conducted the loss function ablation study for MIDAS. This comprehensive evaluation aims to identify the most effective combinations that enhance the student model's proficiency in handling different aspects of dialogue understanding across multiple NLU tasks. Note that we use $\mathcal L_{KD}$ and $\mathcal L_{sce}$ as compulsory knowledge distillation losses, and conduct an ablation study of three other multi-level teacher losses: $\mathcal L_{sim}$, $\mathcal L_{rel}$, and $\mathcal L_{tp}$. As shown in Table \ref{tab:lossab}, the results indicate that incorporating $\mathcal L_{rel}$ with $\mathcal L_{sim}$ achieves the best or the second best performance across all tasks and datasets. Although $\mathcal L_{rel}$ and $\mathcal L_{sim}$ share a similar trend, their impact on model learning may be somewhat superior when employed independently, particularly when utilising $\mathcal L_{sim}$. 
While incorporating $\mathcal L_{tp}$ with the others led to a slight performance increase, it did not match the effectiveness observed with the sole application of the earlier losses.\footnote{We conducted testing with $\mathcal L_{tp}$ only, it produces much lower performance than any others. See the details in section 5.6}
 % See the details in Appendix \ref{ap:teacher-loss}.

% Key loss functions employed for training the proposed model include $\mathcal L_{KD}$, $\mathcal L_{sim}$, and $\mathcal L_{rel}$.
% Table \ref{tab:lossab} presents the results of the ablation study conducted on the loss functions.
% $\mathcal L_{KD}$ plays a crucial role, serving as the loss function for the student to distill knowledge from the teacher model.
% Results indicate that incorporating $\mathcal L_{sim}$ with $\mathcal L_{KD}$ leads to improved overall performance.
% When $\mathcal L_{rel}$ and $\mathcal L_{sim}$ are used alongside $\mathcal L_{KD}$, a slight performance degradation is observed compared to using $\mathcal L_{KD}$ and $\mathcal L_{sim}$ alone.
% This indicates that although $\mathcal L_{rel}$ and $\mathcal L_{sim}$ share a similar role, their impact on model learning may be somewhat superior when employed independently, particularly when utilizing $\mathcal L_{sim}$
% Additional common loss functions involve $\mathcal L_{sce}$ and $\mathcal L_{tp}$. 
% The combined use of $\mathcal L_{sce}$ with the previous three losses resulted in a significant decline in performance. 
% While incorporating $\mathcal L_{tp}$ with the others led to a slight performance increase, it did not match the effectiveness observed with the sole application of the earlier losses.

\begin{table}[ht]
\scriptsize
\setlength\extrarowheight{1pt}
\setlength\tabcolsep{3.9pt}
\centering
    \begin{tabular}[b]{l|c|c|c|c|c|c}
        \hline
        \multirow{3}{*}{} & \multicolumn{2}{c|}{\cellcolor{YellowGreen!25}{\small\textbf{ID}}} & \multicolumn{2}{c|}{\cellcolor{YellowGreen!25}{\small\textbf{SF}}} & \multicolumn{2}{c}{\small\textbf{DC}} \\
        \cline{2-7}
        & \textbf{MWOZ} & \textbf{M2M} & \textbf{MWOZ} & \textbf{M2M} & \textbf{MWOZ} & \textbf{M2M}  \\
        & \textbf{(ACC)} & \textbf{(ACC)} & \textbf{(F1)} & \textbf{(F1)} & \textbf{(ACC)} & \textbf{(ACC)} \\
        \hline 
        \hline 
        ID-only     &   0.8339  &   0.8097   &   0.9079   &   0.9326   &   0.6183 &   0.7147 \\
        SF-only    &   0.8403   &   0.8945   &   0.9620   &   0.9434   &   0.2471 &   0.8917 \\
        DC-only    &   0.8469   &   0.8929   &   0.9547   &   0.9251   &   0.7521 &   0.8913 \\
        ID+SF    &   0.8451   &   0.9083   &   0.9928   &   0.9802   &   0.8734 &   0.8923 \\
        ID+DC    &   0.8373   &   0.9114   &   0.9921   &   0.9797   &   0.8763 &   0.8888 \\
        SF+DC    &   0.8453   &   0.9147   &   0.9927   &   0.9805   &   0.8707 &   0.8912 \\
        \hline
        \textbf{ID+SF+DC}    &   \textbf{0.8502}   &   \textbf{0.9377}   &   \textbf{0.9928}   &   \textbf{0.9813} &   \textbf{0.8816}   &   \textbf{0.8945} \\
        \hline
    \end{tabular}
    % \vspace*{-1mm}
    \caption{The performance based on the type of teacher models. The first column indicates the type of teacher used. For example, \textit{ID+SF+DC} uses all intent classification, slot filling, and domain classification teachers, while \textit{ID-only} uses only the intent classification teacher. Only RoBERTa is utilised as the teacher model.}
    \vspace*{-5mm}
    \label{tab:ap-teacher-comb}
\end{table}

\subsection{Combination-based ablation study}
We explore the impact of incorporating each multi-level teacher ($ID$, $SF$, $DC$) in all three multi-turn dialogue understanding tasks.
Table \ref{tab:ap-teacher-comb} presents the results for each combination of teacher models for three different dialogue understanding tasks. 
The experimental results are when only RoBERTa is adopted as the teacher model.
MIDAS is evaluated with individual teachers ($ID$, $SF$, $DC$), all possible pairs from \{$ID$, $SF$, $DC$\}, and then with all three teachers. 
For example, $ID$+$SF$+$DC$ uses all intent classification, slot filling, and domain classification teachers, while ID-only uses only the intent classification teacher.
Note that we do not adopt $\mathcal L_{rel}$ while two models are used since it is not possible to adopt when there are two teachers. 
The experimental findings highlight that the \textbf{$ID$+$SF$+$DC$} combination attains the highest performance, underscoring the advantage of the student model integrating knowledge from all teachers for each natural language understanding task.

\subsection{Loss function ablation study}

\begin{table*}[t]
\setlength\extrarowheight{1pt}
\setlength\tabcolsep{4pt}
\scriptsize
\centering
    \begin{tabular}{c|c|c|c|c|c|c|c|c|c|c}
        \hline
        
        \multirow{3}{*}{{\small $\bm{\mathcal L_{\scriptstyle\textbf{\textit{KD}}}}$}} &  
        
        \multirow{3}{*}{{\small$\bm{\mathcal L_{\scriptstyle\textbf{\textit{sce}}}}$}} &

        \multirow{3}{*}{{\small$\bm{\mathcal L_{\scriptstyle\textbf{\textit{sim}}}}$}} & 
        
        \multirow{3}{*}{{\small$\bm{\mathcal 
        L_{\scriptstyle\textbf{\textit{rel}}}}$}} & \multirow{3}{*}{{\small$\bm{\mathcal L_{\scriptstyle\textbf{\textit{tp}}}}$}} &

        \multicolumn{2}{c|}{\cellcolor{YellowGreen!25}{\small\textbf{ID}}} & \multicolumn{2}{c|}{\cellcolor{YellowGreen!25}{\small \textbf{SF}}} & \multicolumn{2}{c}{{\small\textbf{DC}}} \\
        \cline{6-11}
        &&&&& ~~~\textbf{MWOZ}~~~ & ~~~\textbf{M2M}~~~ & ~~~\textbf{MWOZ}~~~ & ~~~\textbf{M2M}~~~ & ~~~\textbf{MWOZ}~~~ & ~~~\textbf{M2M}~~~  \\
        &&&&& \textbf{(ACC)} & \textbf{(ACC)} & \textbf{(F1)} & \textbf{(F1)} & \textbf{(ACC)} &\textbf{(ACC)} \\[0.2ex]
        \hline
        \hline 
        
        $\bigcirc$ & $\bigcirc$ & $\bigcirc$ & $\times$ & $\times$ & 0.8441 & 0.9362 & \textbf{0.9928} & \textbf{0.9842} & 0.8744 & \textbf{0.8945} \\
        
        $\bigcirc$ & $\bigcirc$ & $\times$ & $\bigcirc$ & $\times$ & 0.8459 & \textbf{0.9377} & {0.9610} & {0.8415} & \textbf{0.8816} & 0.8914 \\
        
        \rowcolor{LightCyan} $\bigcirc$ & $\bigcirc$ & $\bigcirc$ & $\bigcirc$ & $\times$ &  \textbf{0.8502} & \underline{0.9376} & \textbf{0.9928} & \underline{0.9813} & {0.8803} & \textbf{0.8945} \\
        
        $\bigcirc$ & $\bigcirc$ & $\bigcirc$ & $\bigcirc$ & $\bigcirc$ & \underline{0.8488} & 0.9264 & \underline{0.9912} & 0.9704 & \underline{0.8811} & \underline{0.8922} \\
        \hline
    \end{tabular}
    \caption{The comparison of the diverse loss function combinations. Only RoBERTa is utilised as the teacher model. We adopt two $\mathcal L_{KD}$ and $\mathcal L_{sce}$ as compulsory knowledge distillation loss and also explore three $\mathcal L_{rel}$, $\mathcal L_{sim}$, and $\mathcal L_{tp}$ for MIDAS. Scores in bold indicate leadership among the metrics, and underlined scores indicate the second-best.}
    \label{tab:ap-lossab}
    \vspace*{-1mm}
\end{table*}

We conducted the loss function ablation study for MIDAS with RoBERTa Teacher. This comprehensive evaluation aims to identify the most effective combinations that enhance the student model's proficiency in handling different aspects of dialogue understanding across multiple NLU tasks. Note that we use $\mathcal L_{KD}$ and $\mathcal L_{sce}$ as compulsory knowledge distillation losses, and conduct an ablation study of three multi-level teacher losses: $\mathcal L_{sim}$, $\mathcal L_{rel}$, and $\mathcal L_{tp}$. Among them, $\mathcal L_{rel}$, and $\mathcal L_{tp}$ are newly proposed losses in our work.  As shown in Table \ref{tab:ap-lossab}, the results indicate that incorporating $\mathcal L_{sim}$ with $\mathcal L_{rel}$ achieves the best or the second best performance across all tasks and datasets. Although $\mathcal L_{rel}$ and $\mathcal L_{sim}$ share a similar trend, their impact on model learning may be somewhat superior when employed independently, particularly when utilising $\mathcal L_{sim}$. 
While incorporating $\mathcal L_{tp}$ with the others led to a slight performance increase, it did not match the effectiveness observed with the sole application of the earlier losses. We assume the reason is that it does not mainly focus on the teacher prediction supervised loss, not really consider the relations with student models

\subsection{Qualitative analysis: Case study}
\label{main:qual-analysis}

We further evaluate MIDAS using a qualitative assessment of three NLU tasks with the M2M dataset. As shown in Table \ref{tab:case}, we assume a three-utterance dialogue: “\textit{how many tickets would you like to buy?},” “\textit{1},” and “\textit{what date do you want to go?}.” Based on this conversation, we test $ID$, $SF$, and $DC$. We compare MIDAS with BERT, LLaMa3.1, and GPT4o. \textbf{MIDAS} is trained with three teacher models, $BERT_{ID}$, $BERT_{SF}$, and $BERT_{DC}$. BERT is a fine-tuned model (\textit{BERT-Base}) focused on one task per prediction, while LLaMa3.1 and GPT4o are evaluated using few-shot learning, with three examples provided for each slot type, intent, and domain. The results reveal the limitations of LLMs in handling multi-level knowledge. For instance, LLaMa3.1 correctly identifies the slot \textbf{B-num\_tickets} but incorrectly classifies the domain as \textbf{restaurant}. 

Furthermore, both GPT4o and LLaMa3.1 struggle to follow the prompt instructions, failing to predict enough slot types even when words are explicitly separated. While fine-tuned BERT also fails to capture the conversation domain and slot types, \textbf{MIDAS}, aided by multi-level teacher models, consistently predicts slot types, intents, and domains correctly, matching the ground truth.

\begin{table}[tp]
% \begin{minipage}{0.5\textwidth}
\scriptsize
\setlength\extrarowheight{1pt}
\setlength\tabcolsep{0.3 pt}
\centering
 \vspace*{-1mm}
    \begin{tabular}[b]{c|c|L{4.7cm}|C{0.8cm}|C{0.8cm}}
        \hline
        \textbf{Model} &\textbf{T.}& \multicolumn{1}{c|}{~~~~~~~~~~~~~~~~\textbf{SF}~~~~~~~~~~~~~~~~} & ~ \textbf{ID}~ & ~\textbf{DC}~ \\
% MIDAS (BERT)
        \hline 
        \hline 

        \multirow{3}{*}{\textbf{Utterance}} & \textbf{1} & \hspace{1mm}how, many, tickets, would, you, like, to, buy, ? & - & -
\\ \cline{2-5} & \textbf{2} &\hspace{1mm}1 & - & -
\\ \cline{2-5} & \textbf{3} &\hspace{1mm}what, date, do, you, want, to, go, ? & - & -
\\ \cline{1-5}\\[-3.05mm]\cline{1-5} \\ [-3mm]\multirow{3}{*}{\textbf{GT}} & \textbf{1} &\hspace{1mm}O, O, O, O, O, O, O, O, O & request & movie
\\ \cline{2-5} & {\textbf{2}} &\hspace{1mm}B-num\_tickets & inform & movie
\\ \cline{2-5} & {\textbf{3}} &\hspace{1mm}O, O, O, O, O, O, O, O & request & movie
%GPT4o
\\ \cline{1-5}\\[-3.05mm]\cline{1-5} \\ [-3mm]\multirow{3}{*}{\textbf{GPT4o}} & \textbf{1}& \cellcolor{GoldenRod!25}\hspace{1mm}O, O, O, O, O, O, O, NaN, NaN & \cellcolor{YellowGreen!25}request & \cellcolor{YellowGreen!25}movie
\\ \cline{2-5}\\[-3.1mm]\cline{2-5} \\ [-3mm] & \textbf{2}& \cellcolor{GoldenRod!25}\hspace{1mm}(O) & \cellcolor{WildStrawberry!25}(O) & \cellcolor{WildStrawberry!25}restau
\\ \cline{2-5}\\[-3.1mm]\cline{2-5} \\ [-3mm] & \textbf{3}& \cellcolor{GoldenRod!25}\hspace{1mm}O, B-date, O, O, O, O, NaN, NaN & \cellcolor{YellowGreen!25}request & \cellcolor{WildStrawberry!25}restau
%LLaMa3.1
\\ \cline{1-5}\\[-3.05mm]\cline{1-5} \\ [-3mm]\multirow{3}{*}{\textbf{LLaMa3.1}} & \textbf{1}& \cellcolor{GoldenRod!25}\hspace{1mm}O, O, O, O, O, O, B-num\_tickets, O, NaN & \cellcolor{YellowGreen!25}request & \cellcolor{WildStrawberry!25}restau
\\ \cline{2-5}\\[-3.1mm]\cline{2-5} \\ [-3mm] & \textbf{2}& \cellcolor{GoldenRod!25}\hspace{1mm}B-category & \cellcolor{WildStrawberry!25}request & \cellcolor{YellowGreen!25}movie
\\ \cline{2-5}\\[-3.1mm]\cline{2-5} \\ [-3mm] & \textbf{3}& \cellcolor{YellowGreen!25}\hspace{1mm}O, O, O, O, O, O, O, O & \cellcolor{YellowGreen!25}request & \cellcolor{YellowGreen!25}movie
%BERT-Only
\\ \cline{1-5}\\[-3.05mm]\cline{1-5} \\ [-3mm]\multirow{3}{*}{\textbf{BERT}} & \textbf{1}& \cellcolor{YellowGreen!25}\hspace{1mm}O, O, O, O, O, O, O, O, O & \cellcolor{YellowGreen!25}request & \cellcolor{WildStrawberry!25}restau
\\ \cline{2-5}\\[-3.1mm]\cline{2-5} \\ [-3mm] & \textbf{2}& \cellcolor{GoldenRod!25}\hspace{1mm}O & \cellcolor{YellowGreen!25}inform & \cellcolor{WildStrawberry!25}restau
\\ \cline{2-5}\\[-3.1mm]\cline{2-5} \\ [-3mm] & \textbf{3}& \cellcolor{YellowGreen!25}\hspace{1mm}O, O, O, O, O, O, O, O & \cellcolor{YellowGreen!25}request & \cellcolor{WildStrawberry!25}restau
%MIDAS (BERT)
\\ \cline{1-5}\\[-3.05mm]\cline{1-5} \\ [-3mm]\multirow{3}{*}{\textbf{MIDAS}} & \textbf{1}& \cellcolor{YellowGreen!25}\hspace{1mm}O, O, O, O, O, O, O, O, O & \cellcolor{YellowGreen!25}request & \cellcolor{YellowGreen!25}movie
\\ \cline{2-5}\\[-3.1mm]\cline{2-5} \\ [-3mm] & \textbf{2}& \cellcolor{YellowGreen!25}\hspace{1mm}B-num\_tickets & \cellcolor{YellowGreen!25}inform & \cellcolor{YellowGreen!25}movie
\\ \cline{2-5}\\[-3.1mm]\cline{2-5} \\ [-3mm] & \textbf{3}& \cellcolor{YellowGreen!25}\hspace{1mm}O, O, O, O, O, O, O, O & \cellcolor{YellowGreen!25}request & \cellcolor{YellowGreen!25}movie
 \\

        \hline
    \end{tabular}
    \vspace{-3mm}
    \caption{A Prediction example with a three-turn conversation on slot filling, intent detection, and domain classification. Green: the result that perfectly matches the Ground Truth (GT), Red: Entirely Incorrect, and Yellow: Partially Correct Results. `NaN' means the value at this position is empty, and `(O)' means the outputs of LLMs are out of the domain defined in the prompt. Additional prediction case study examples are articulated in Appendix \ref{ap:case-study}.}
    \label{tab:case}
    \vspace{-3mm}
    % \end{minipage}
\end{table}

% These error cases come from one conversation of M2M. Both our model and RoBERTa are trained/fine-tuned on each task separately. The key point is that our model is trained with three teachers $RoBERTa_{ID}$, $RoBERTa_{SF}$, and $RoBERTa_{ID}$. The RoBERTa models fine-tuned on each single task don't have the knowledge from other tasks. Although the fine-tuned RoBERTa can sometimes predict the intent correctly, it can't make use of the knowledge from word level and domain level simultaneously to recognise the restaurant name and the domain. So we can see in the first two cases, RoBERTa can't recognise the restaurant name at all, and in the third case, even if it recognises the restaurant name, it still predicts the domain as movie. However, our student model can recognise the restaurant name without domain labels when trained on the SF task separately, as well as in the ID and DC tasks. Cases like these prove our assumption that diverse levels of knowledge derived from multi-turn conversation understanding datasets can enhance the comprehension of each specific natural language understanding task, surpassing the benefits of learning from single-level dialogue knowledge.

\section{Conclusion}
\label{sec_con}
This paper introduces a novel multi-level teacher knowledge distillation framework to enhance multi-turn natural language understanding (NLU). By fine-tuning pre-trained models at word, sentence, and document levels, we construct multi-level teachers, imparting their knowledge to a student model. Various loss functions are introduced and explored, and the experiment results demonstrate the framework's effectiveness in improving the student model's understanding across diverse NLU tasks. It shows better than the LLM result. 

\section{Limitation}
\label{sec_lim}
There are some spaces for future work, including a more fine-grained analysis of the impact of each loss, covering multilingual multi-turn dialogue. The quality of the multilingual pre-trained model would be the potential risk to achieve enough multi-turn NLU performance. We believe this work will provide valuable insights into various aspects of dialogue knowledge for NLU and multi-level knowledge distillation.

\section{Acknowledgement}
This work was supported by the Institute of Information \& communications Technology Planning \& Evaluation (IITP) grant funded by the Korea government(MSIT) (No.RS-2022-00155911, Artificial Intelligence Convergence Innovation Human Resources Development (Kyung Hee University))
\bibliography{custom}

\newpage

\appendix

\begin{table*}[t]
\centering
\setlength\tabcolsep{0.5 pt}
\scriptsize
    \begin{tabular}{l|c|C{1cm}|C{1.3cm}|C{1.3cm}|c|l}
        \hline
        \multicolumn{1}{c|}{\textbf{Model}} & ~~\textbf{Year}~~ & \textbf{Word (Slot)} & \textbf{Sentence (Intent)} & {\textbf{Document (Domain)}} & ~\textbf{Dialogue Type}~ &  \multicolumn{1}{c}{\textbf{Joint Integration}} \\
        \hline
        SeqSeq \citet{Liu2016} & 2016 & $\bigcirc$ & $\bigcirc$ & $\times$ & Single-Turn & BiRNN + Attention \\
        SDEN \citet{Bapna2017} & 2017 & $\bigcirc$ & $\bigcirc$ & $\bigcirc$ & Multi-Turn & BiRNN + Memory Network \\
        Slot-Gated \citet{Goo2018} & 2018 & $\bigcirc$ & $\bigcirc$ & $\times$ & Single-Turn & BiLSTM + Slot Gate\\
        BLSTM+attention \citet{Tingting2019} & 2019 & $\bigcirc$ & $\bigcirc$ & $\times$ & Single-Turn & BiLSTM + Attention \\
        Co-Interactive Transformer \citet{Qin2021} & 2021 & $\bigcirc$ & $\bigcirc$ & $\times$ & Single-Turn & BiLSTM + Attention \\
        GL-GIN \citet{Qin2021:ICASSP} & 2021 & $\bigcirc$ & $\bigcirc$ & $\times$ & Single-Turn & BiLSTM + GAT \\
        SyntacticTF \citet{Wang2021} & 2021 & $\bigcirc$ & $\bigcirc$ & $\times$ & Single-Turn & Transformer \\
        STD \citet{Jiang2021} & 2021 & $\bigcirc$ & $\bigcirc$ & $\times$ & Single-Turn & Transformer + One-teacher KD \\
        JointIDSF \citet{Dao2021} & 2021 & $\bigcirc$ & $\bigcirc$ & $\times$ & Single-Turn & CRF + Attention \\
        CaBERT-SLU \citet{Wu2021:INTERSPEECH} & 2021 & $\bigcirc$ & $\bigcirc$ & $\bigcirc$ & Multi-Turn &  Attention \\
        SDJN \citet{Chen2022:ICASSP} & 2021 & $\bigcirc$ & $\bigcirc$ & $\times$ & Single-Turn & BiLSTM + self KD \\
        HAN \citet{Chen2022} & 2022 & $\bigcirc$ & $\bigcirc$ & $\times$ & Single-Turn & BiLSTM + Attention \\
        ReLA-NET \citet{Xing2022} & 2022 & $\bigcirc$ & $\bigcirc$ & $\times$ & Single-Turn & BiLSTM + GAT \\
        XAI Attention \citet{Gunaratna2022} & 2022 & $\bigcirc$ & $\bigcirc$ & $\times$ & Multi-Turn & XAI \\
        WFST-BERT \citet{Abro2022} & 2022 & $\bigcirc$ & $\bigcirc$ & $\times$ & Single-Turn & WFST \\
        Contextual SLU \citet{Tran2022} & 2022 & $\bigcirc$ & $\bigcirc$ & $\bigcirc$ & Multi-Turn & BiLSTM + Attention \\
        TKDF \citet{Cheng2023} & 2023 & $\bigcirc$ & $\bigcirc$ & $\times$ & Single-Turn & SSRAN + One-teacher KD \\
        MISCA \citet{Pham2023} & 2023 & $\bigcirc$ & $\bigcirc$ & $\times$ & Single-Turn & BiLSTM + Attention \\
        PAGM \citet{Mei2023} & 2023 & $\bigcirc$ & $\bigcirc$ & $\times$ & Single-Turn & Gate \\
        FAN \citet{Huang2023} & 2023 & $\bigcirc$ & $\bigcirc$ & $\times$ & Single-Turn & Attention \\
        Tri-level JNLU \citet{Weld2023} & 2023 & $\bigcirc$ & $\bigcirc$ & $\bigcirc$ & Multi-Turn & Transformer \\
        CKA-NLU \citet{Wu2023} & 2023 & $\bigcirc$ & $\bigcirc$ & $\bigcirc$ & Multi-Turn & Attention \\
        BiSLU \citet{Tu2023} & 2023 & $\bigcirc$ & $\bigcirc$ & $\times$ & Single-Turn & self KD \\
        PACL \citet{chen2024two} & 2024 & $\bigcirc$ & $\bigcirc$ & $\times$ & Multi-Turn & Contrastive Learning + Attention \\
        BiJM \citet{luo2024bi} & 2024 & $\bigcirc$ & $\bigcirc$ & $\times$ & Single-Turn & Transformer + Enhance Layer \\
        \hline
        \textbf{MIDAS (Ours)} & \textbf{2024} & \textbf{$\bigcirc$} & \textbf{$\bigcirc$} & \textbf{$\bigcirc$} & \textbf{Multi-Turn} & \textbf{Multi-teacher KD} \\
        \hline
    \end{tabular}
    \caption{Summary of previous joint NLU models and MIDAS. Word, Sentence, and Document columns indicate whether the relevant information is used for joint integration. GAT in the Joint Integration column refers to the graph attention network, KD refers to knowledge distillation, and WFST refers to Weighted Finite-State Transducers.}
    \vspace{-0mm}
    \label{tab:ap_related}
\end{table*}

\section{Related works}
\label{ap:rel}

Table \ref{tab:ap_related} presents a comparison of MIDAS with 23 previous joint NLU models.
Recently, most NLU studies have embraced a joint learning model capable of handling all NLU tasks to mitigate error propagation inherent in pipelined approaches \cite{Wang2021,han2021bidir,Gunaratna2022,Huang2023}. The initial joint models employed traditional neural networks like RNN \cite{Liu2016} and LSTM \cite{Tingting2019,Qin2021:ICASSP,Chen2022,Xing2022,Tran2022,Pham2023} with attention mechanisms. 

All models leverage slot-level knowledge and intent-level knowledge, but only five previous works incorporate domain-level knowledge. 
This implies that only five prior studies utilised a multi-turn dialogue dataset. 

Only one previous study \cite{Weld2023} conducted tests on domain classification.
Hence, we chose \cite{Weld2023} as a representative baseline. 
What sets the proposed model apart is its utilisation of multi-teacher knowledge distillation. While two previous works employed self-knowledge distillation and another two adopted one-teacher knowledge distillation, MIDAS represents the first attempt at employing multi-teacher knowledge distillation for joint learning in natural language understanding.

% \section{The role of each loss function}
% \label{ap:each_loss}
% The following losses are utilized to train the student model, each playing a distinct role:
% \begin{itemize}
%     \item $L_{kd}$: This loss facilitates the transfer of knowledge from the teacher models to the student model, enabling the student to mimic the general behavior of multiple teachers.
%     \item $L_{rel}$: This loss is designed to capture the relationships between different samples in the input data. It helps to align the student's understanding with that of the teacher models and ensures a consistent perspective on the dataset.

%     \item $L_{sim}$: This loss encourages the student model to generate outputs that are similar to those of the teacher models in terms of their overall structure and distribution. It helps to maintain consistency between the student and teacher predictions.
%     \item $L_{sce}$: This loss function serves as the fundamental mechanism for training the student model. It entails the student learning to predict the correct labels associated with the input data.
%     \item $L_{tp}$: This loss leverages the predictions of the teacher models to provide additional supervision signals to the student model. It helps to guide the student towards making predictions that align with those of the teachers.
% \end{itemize}

\begin{algorithm}[tb]
    \scriptsize
    \caption{Triplet Relations}
    \label{alg:algorithm}
    \begin{multicols}{2}
    \vspace*{-3mm}
    \textbf{Input}: The hidden states of the batch data from the teachers $H_t=\{h_1^1,h_1^2,...h_1^n,...,h_j^n\}$,  the hidden states of the batch data from the student $H_s=\{h_s^1,h_s^2,...h_s^n\}$, the teacher model set $T=\{T_1,T_2,...,T_j\}$\\
    \textbf{Parameter}: Distance function $\mathcal{F}_D$\\
    \textbf{Output}: The batch size of triplet relations $\mathcal{T}$\\
        \begin{algorithmic}[1] %[1] enables line numbers
            \STATE Let $i=0$, $\mathcal{T}=\emptyset$.
            \FOR{ $i < n $} 
            \STATE Randomly select three samples from the batch and label their indexes in the batch as $r1, r2, r3$.
            \STATE Treat the sample indexed $r1$ as the anchor, $r2$ as the positive sample, $r3$ as the negative sample.
            \STATE Let $l=0,flag=0$
            \FOR{ $l<j$ }
            \STATE $d_{1,2}=\mathcal{F}_D(h_l^{r1}, h_l^{r2})$,\\$d_{1,3}=\mathcal{F}_D(h_l^{r1}, h_l^{r3})$
            \IF {$d_{1,2}>d_{1,3}$}
            \STATE $flag\mathrel{+}=1$
            \ELSE
            \STATE $flag\mathrel{-}=1$
            \ENDIF
            \STATE $l\mathrel{+}=1$
            \ENDFOR
            \IF {$flag > 0$}
            \STATE Swap the labels of $h_l^{r2}$ and $h_l^{r3}$.
            \ENDIF
            \STATE $i\mathrel{+}=1$
            \STATE $\mathcal{T} \mathrel{+}= [h_s^{r1}, h_s^{r2}, h_s^{r3}]$
            \ENDFOR
            \STATE \textbf{return} $\mathcal{T}$
        \end{algorithmic}
    \end{multicols}
    \vspace*{-4mm}
\end{algorithm}

% \section{Details of datasets}
% \label{ap:datasets}

% \textbf{MWOZ} \cite{Zang2020} is specifically designed for Dialogue State Tracking (DST) and adopts the conventional human-vs-human Wizard of Oz approach across diverse domains, including attraction, bus, hospital, hotel, police, restaurant, taxi, and train. It incorporates 30 slot types and 11 intent types. The dataset comprises 8,437 dialogues, with an average of 5.68 turns per dialogue and 14.07 tokens per turn. Following by \cite{Liu2016,Goo2018,Weld2023}, we do not consider any multi-label samples but utilise the data with a single domain and intent.

% \textbf{M2M} \cite{Liu2018} is introduced with virtual agents and user-generated interactions to emulate goal-directed conversations through paraphrasing with templated utterances. M2M has movies and restaurant domains. The slots and intents are categorical, with 21 slot types and 15 intent types. The dataset comprises 1,500 dialogues, with an average of 9.86 turns per dialogue and 8.25 tokens per turn.

\section{Details of baselines}
\label{ap:baselines}
% Due to the limited number of baselines available for Multi-turn Dialogue Understanding, we adopted the following models as baselines. \textbf{BERT-Base}\footnote{\url{https://huggingface.co/bert-base-uncased}} is representative transformer encoder-based language model.
% \textbf{RoBERTa-Base}\footnote{\url{https://huggingface.co/roberta-base}} builds on BERT and adjusts hyperparameters by eliminating the next-sentence prediction objective. It also trains with larger mini-batches and learning rates. 
% \textbf{ALBERT-Base}\footnote{\url{https://huggingface.co/albert-base-v2}} is a BERT-based model that demonstrated superior performance by reducing the model size with factorised embedding parameterisation and cross-layer parameter sharing.
% \textbf{SeqSeq} \cite{Liu2016} is an RNN with attention mechanisms, designed for the joint tasks of $ID$ and $SF$. 
% \textbf{Slot-Gated} \cite{Goo2018} introduces a slot gate to capture the relationship between intent and slot, aiming to improve semantic understanding through global optimisation.
% \textbf{Tri-level JNLU} \cite{Weld2023} is a pioneering model that incorporates domain information in the joint modelling of $ID$ and $SF$.

Given the limited number of baselines available for Multi-turn Dialogue Understanding, we selected the following models as baselines. \textbf{BERT-Base}\footnote{\url{https://huggingface.co/bert-base-uncased}} is a transformer-based language model that serves as a standard benchmark. \textbf{RoBERTa-Base}\footnote{\url{https://huggingface.co/roberta-base}} improves upon BERT by removing the next-sentence prediction task and optimizing hyperparameters, such as using larger mini-batches and higher learning rates. \textbf{ALBERT-Base}\footnote{\url{https://huggingface.co/albert-base-v2}} further enhances efficiency through factorized embedding parameterization and cross-layer parameter sharing, achieving better performance with fewer parameters. \textbf{SeqSeq} \cite{Liu2016} is an RNN model with attention mechanisms, developed for joint intent detection ($ID$) and slot filling ($SF$) tasks. \textbf{Slot-Gated} \cite{Goo2018} introduces a slot-gating mechanism to capture the relationship between intent and slot labels, improving semantic understanding through global optimization. Lastly, \textbf{Tri-level JNLU} \cite{Weld2023} incorporates domain information for enhanced joint modeling of $ID$ and $SF$.

\section{Implementation details}
\label{ap:imp-detail}
% We introduce some implementation details in this section 396
% and the complete implementation details in Appendix G. For 397
% Multi-teacher fine-tuning, we use BERT-Base, RoBERTa- 398
% Base and LLaMA2-7b5 as the teacher backbones and fine-tune 399
% them on each task. For fine-tuning LLaMA2-7b, we adopt an 400
% unmask strategy used in [Li et al., 2023]. We use AdamW 401
% [Loshchilov and Hutter, 2018] and CrossEntropy loss to fine- 402
% tune the pre-trained models for 3 epochs. The learning rate 403
% is 5e-5 and is warm-uped linearly from 0 to 5e-5 during the 404
% first 10% training steps. The batch size is 32. For Multi-level 405
% Distillation, we use AdamW and the aforementioned losses 406
% to train the student model with multi-level teachers. We use 407
% Squared Euclidean distance in algorithm 1 and cosine simi- 408
% larity in equation 1. When we use the combination of these 409
% losses, we sum them without any weight. We use the same 410
% optimiser, learning rate, warm-up strategy, and batch size as 411
% the one used in teacher fine-tuning. We use a vanilla Trans- 412
% former encoder as the student model.

% Hyperparameter tables
% Teacher Model parameters and size table
% Student Model detail table: , with 6 layers and 8 heads. The embedding, hidden dimension and dropout rate are 768, 2048, and 0.3 respectively.
% Hardware and platform information table
% \textbf{Hardware information}. Our experiments are run on the Linux platform with A6000 Nvidia graphic card and AMD Ryzen Threadripper PRO 5955WX 16-Cores CPU, and the rem is 128G.

\subsection{Experiment hyperparameters}
Table \ref{tab:hp} presents the hyperparameters, used in our proposed Multi-level Teacher Fine-tuning, as well as Multi-Teacher Knowledge Distillation. The Implementation details can be found in Section \ref{subsec:metrics}. of the main submission.  

\begin{table}[H]
\centering
\setlength\tabcolsep{2.5 pt}
\scriptsize
    \begin{tabular}{l|c|c}
        \hline
        \multicolumn{1}{c|}{\textbf{Hyper-parameter}} & ~~\textbf{Fine-tuning}~~ & ~~\textbf{Knowledge Distillation}~~ \\
        \hline
        Learning Rate & 5e-5 & 5e-5 \\
        \hline
        Batch Size & 32 & 32 \\
        \hline
        Warm-up Steps & 10\% of Max epoch & 10\% of Max epoch \\
        \hline
        Mex epoch & 3 & 100 \\
        \hline
        Stop Strategy & Max Epoch & Early Stopping on loss \\
        \hline
        Stop Patience & - & 10 \\
        \hline
        Optimizer  & AdamW & AdamW \\
        \hline
        Optimizer Weight Decay & 1e-2 & 1e-2 \\
        \hline
        Optimizer Betas & 0.9, 0.999 & 0.9, 0.999 \\
        \hline
        Margin in $\mathcal L_{rel}$ & - & 0.2 \\
        \hline
        Norm in $\mathcal L_{rel}$ & - & 2 \\
        \hline
        $\mathcal{F}_D$ in $\mathcal L_{rel}$ & - & L2-Norm \\
        \hline
        Similarity in $\mathcal L_{sim}$ & - & Cosine Similarity \\
        \hline
        Max Token Length & 512 & 512 \\
        \hline
    \end{tabular}
    \caption{The hyper-parameters used in our experiments.}
    \label{tab:hp}
\end{table}

We further present the results of various experiments conducted to select hyperparameters, particularly the learning rate, in Table \ref{tab:ap-learning-rate}. 
In all tests, the temperature is fixed at 20, and only the learning rate is changed to 0.0005, 0.00005, and 0.000005. 
In the experiments on the M2M dataset, the performance of Gemma-7b alongside LLaMa2-7b is also measured to compare performance with the generative model. 
The highest accuracy is shown when the learning rate was 0.00005, and Gemma-7b shows similar performance to LLaMa2-7b, but LLaMa2-7b is slightly superior.
The best performance is observed when the learning rate is 0.00005, which is also the case in experiments on the MWOZ dataset.

\subsection{Model details}
We display the visualisation of teacher models and our student model Vanilla Transformer Encoder together. Those two summarises can be found in Table \ref{tab:model_detail}. Note that we use LoRA to fine-tune LLaMa 2-7b.

\subsection{Hardware information}
Our experiments are run on the Linux platform with an A6000 Nvidia graphic card and an AMD Ryzen Threadripper PRO 5955WX 16-core CPU, and the RAM is 128G.

\begin{table}[H]
\centering
\setlength\tabcolsep{4pt}
\scriptsize
    \begin{tabular}{c|c|c|c}
        \hline
        ~~~~~\textbf{Model}~~~~~ & ~~~\textbf{Task}~~~ & ~~\textbf{Learning Rate}~~ & ~~\textbf{Accuracy}~~ \\
        \hline
        \multicolumn{4}{c}{\textbf{M2M}}\\
        \hline 
        \multirow{9}{*}{LLaMa2-7b} & \multirow{3}{*}{ID} & 0.0005 & 0.9121 \\
        & & 0.00005 & 0.9392 \\
        & & 0.000005 & 0.9093 \\
        \cline{2-4}
        & \multirow{3}{*}{SF} & 0.0005 & 0.9696 \\
        & & 0.00005 & 0.9833 \\
        & & 0.000005 & 0.9349 \\
        \cline{2-4}
        & \multirow{3}{*}{DC} & 0.0005 & 0.8895 \\
        & & 0.00005 & 0.8804 \\
        & & 0.000005 & 0.8375 \\
        \hline
        \multirow{9}{*}{Gemma-7b} & \multirow{3}{*}{ID} & 0.0005 & 0.9204 \\
        & & 0.00005 & 0.9357 \\
        & & 0.000005 & 0.9102 \\
        \cline{2-4}
        & \multirow{3}{*}{SF} & 0.0005 & 0.9693 \\
        & & 0.00005 & 0.9816 \\
        & & 0.000005 & 0.9429 \\
        \cline{2-4}
        & \multirow{3}{*}{DC} & 0.0005 & 0.8799 \\
        & & 0.00005 & 0.8840 \\
        & & 0.000005 & 0.7890 \\
        \hline
        \multicolumn{4}{c}{\textbf{MWOZ}}\\
        \hline 
        \multirow{9}{*}{LLaMa2-7b} & \multirow{3}{*}{ID} & 0.0005 & 0.8021 \\
        & & 0.00005 & 0.8403 \\
        & & 0.000005 & 0.7952 \\
        \cline{2-4}
        & \multirow{3}{*}{SF} & 0.0005 & 0.9776 \\
        & & 0.00005 & 0.9912 \\
        & & 0.000005 & 0.9740 \\
        \cline{2-4}
        & \multirow{3}{*}{DC} & 0.0005 & 0.8411 \\
        & & 0.00005 & 0.8702 \\
        & & 0.000005 & 0.7026 \\
        \hline
    \end{tabular}
    \caption{Summary of performance changes according to learning rate changes.}
    \label{tab:ap-learning-rate}
\end{table}

\begin{table}[htp]
\centering
\setlength\tabcolsep{4pt}
\scriptsize
    \begin{tabular}{l|c|c|c|c}
        \hline
        \multicolumn{1}{c|}{} & ~~\textbf{BERT}~~ & ~~\textbf{RoBERTa}~~ & ~~\textbf{LLaMa}~~ & ~~\textbf{Student}~~\\
        \hline
        Architecture & Encoder & Encoder & Decoder & Encoder\\
        \hline
        Parameters & 110M & 125M & 7B & 58M\\
        \hline
        Layers & 12 & 12 & 32 & 6 \\
        \hline
        Heads & 12 & 12 & 32 & 8\\
        \hline
        Hidden Dim. & 768 & 768 & 4096 & 768\\
        \hline
        Feed Forward Dim. & 3072 & 3072 & 11008 & 2048\\
        \hline
        Dropout Rate & 0.1 & 0.1 & 0.0 & 0.3\\
        \hline
        Rank of LoRA & - & - & 64 \\
        \hline
        Alpha of LoRA & - & - & 16 \\
        \hline
        Dropout of LoRA & - & - & 0.1 \\
        \hline
    \end{tabular}    \caption{The details of the models used in our work.}
    \label{tab:model_detail}
\end{table}

\begin{table*}[tp]
\tiny
\setlength\extrarowheight{1pt}
\setlength\tabcolsep{1pt}
\centering
    \begin{tabular}[b]{c|c|L{9.5cm}|c|c}
        \hline
        \textbf{No.} & ~~~\textbf{Model}~~~ & \multicolumn{1}{c|}{~~~~~~~~~~~~~~~~\textbf{Tokens (Slot)}~~~~~~~~~~~~~~~~} & ~~~ \textbf{Intent}~~~ & ~~~\textbf{Domain}~~~ \\
        \hline 
        \hline 
        \multirow{4}{*}{\textbf{1}} & \textbf{Utterance} & near, kirkland, and, i, don, ', t, care, about, the ratings & - & -\\ \cline{2-5} & \textbf{Ground Truth} & O, B-location, O, O, O, O, O, O, O, O, O & inform & restaurant\\
                           \cline{2-5}\\[-2.75mm]\cline{2-5}\\[-2.65mm]
                           & MIDAS (BERT) &  \cellcolor{YellowGreen!25}O, B-location, O, O, O, O, O, O, O, O, O &  \cellcolor{YellowGreen!25}inform & \cellcolor{YellowGreen!25}restaurant\\  \cline{2-5}\\[-2.65mm]
                           & BERT-Only &  \cellcolor{GoldenRod!25}O, B-restaurant\_name, O, O, O, O, O, O, O, O, O &  \cellcolor{YellowGreen!25}infrom & \cellcolor{WildStrawberry!25}movie\\        \cline{1-5}\\[-2.7mm]\cline{1-5}\\[-2.65mm]
        \multirow{4}{*}{\textbf{2}} & \textbf{Utterance} & let, ', s, go, with, the, view& - & -\\ \cline{2-5}
                           & \textbf{Ground Truth} & O, O, O, O, O, B-restaurant\_name, I-restaurant\_name & affirm & restaurant\\  
                           \cline{2-5}\\[-2.75mm]\cline{2-5}\\[-2.65mm]
                           & MIDAS (BERT) &   \cellcolor{YellowGreen!25}O, O, O, O, O, B-restaurant\_name, I-restaurant\_name &  \cellcolor{YellowGreen!25}affirm & \cellcolor{YellowGreen!25}restaurant\\  \cline{2-5}\\[-2.65mm]
                           & BERT-Only & \cellcolor{GoldenRod!25}O, O, O, O, O, O, O &  \cellcolor{YellowGreen!25}affirm & \cellcolor{WildStrawberry!25}movie\\ 
                           \cline{1-5}\\[-2.7mm]\cline{1-5}\\[-2.65mm]
        \multirow{4}{*}{\textbf{3}} & \textbf{Utterance} & then, find, me, one, in, the, expensive, price, range. & - & -\\ \cline{2-5}
                           & \textbf{Ground Truth} & O, O, O, O, O, O, O, O, O & find\_hotel & hotel\\
                           \cline{2-5}\\[-2.75mm]\cline{2-5}\\[-2.65mm]
                           & MIDAS (BERT) &  \cellcolor{YellowGreen!25}O, O, O, O, O, O, O, O, O &  \cellcolor{YellowGreen!25}find\_hotel & \cellcolor{YellowGreen!25}hotel\\  \cline{2-5}\\[-2.65mm]
                           & BERT-Only &  \cellcolor{YellowGreen!25}O, O, O, O, O, O, O, O, O &  \cellcolor{WildStrawberry!25}find\_restaurant & \cellcolor{WildStrawberry!25}restaurant\\
                           \cline{1-5}\\[-2.7mm]\cline{1-5}\\[-2.65mm]
        \multirow{4}{*}{\textbf{4}} & \textbf{Utterance} & which, ever, is, nice., i, will, need, some, info, on, it, too.& - & -\\ \cline{2-5}
                           & \textbf{Ground Truth} & O, O, O, O, O, O, O, O, O, O, O, O & find\_attraction & attraction\\  
                           \cline{2-5}\\[-2.75mm]\cline{2-5}\\[-2.65mm]
                           & MIDAS (BERT) &   \cellcolor{YellowGreen!25}O, O, O, O, O, O, O, O, O, O, O, O &  \cellcolor{YellowGreen!25}find\_attraction & \cellcolor{YellowGreen!25}attraction\\  \cline{2-5}\\[-2.65mm]
                           & BERT-Only & \cellcolor{YellowGreen!25}O, O, O, O, O, O, O, O, O, O, O, O & \cellcolor{WildStrawberry!25}find\_hotel & \cellcolor{WildStrawberry!25}restaurant\\ 
                           \cline{1-5}\\[-2.7mm]\cline{1-5}\\[-2.65mm]
        \multirow{4}{*}{\textbf{5}} & \textbf{Utterance} & great, we, are, meeting, friends, at, wandlebury, country, park, before, we, eat,, can, you, tell, me, about, that, place, and, where, it, is? & - & -\\ \cline{2-5}
                           & \textbf{Ground Truth} & O, O, O, O, O, O, B-attraction-name, I-attraction-name, I-attraction-name, O, O, O, O, O, O, O, O, O, O, O, O, O, O & find\_attraction & attraction\\  \cline{2-5}\\[-2.75mm]\cline{2-5}\\[-2.65mm]
                           & MIDAS (BERT) &  \cellcolor{YellowGreen!25}O, O, O, O, O, O, B-attraction\_name, I-attraction\_name, I-attraction\_name, O, O, O, O, O, O, O, O, O, O, O, O, O, O & \cellcolor{YellowGreen!25}find\_attraction &  \cellcolor{YellowGreen!25}attraction\\  \cline{2-5}\\[-2.65mm]
                           & BERT-Only &  \cellcolor{GoldenRod!25}O, O, O, O, O, O, B-attraction-name, I-hotel-name, O, O, O, O, O, O, O, O, O, O, O, O, O, O, O  &  \cellcolor{WildStrawberry!25}find\_restaurant &  \cellcolor{WildStrawberry!25}restaurant\\
                           \cline{1-5}\\[-2.7mm]\cline{1-5}\\[-2.65mm]
        \multirow{4}{*}{\textbf{6}} & \textbf{Utterance} & yes, may, i, have, the, address, post,code, and, phone, number, for, golden, house?, i'll, book, it, myself. & - & -\\ \cline{2-5}
                           & \textbf{Ground Truth} & O, O, O, O, O, O, O, O, O, O, O, B-restaurant\_name, I-restaurant\_name, O, O, O, O
                           & find\_restaurant & restaurant\\  \cline{2-5}\\[-2.75mm]\cline{2-5}\\[-2.65mm]
                           & MIDAS (BERT) &  \cellcolor{YellowGreen!25}O, O, O, O, O, O, O, O, O, O, O, B-restaurant\_name, I-restaurant\_name, O, O, O, O
                           & \cellcolor{YellowGreen!25}find\_restaurant &  \cellcolor{YellowGreen!25}restaurant\\  \cline{2-5}\\[-2.65mm]
                           & BERT-Only &  \cellcolor{GoldenRod!25}O, O, O, O, O, O, O, O, O, O, O, O, O, O, O, O, O
                           &  \cellcolor{WildStrawberry!25}find\_hotel &  \cellcolor{WildStrawberry!25}hotel\\
                           \cline{1-5}\\[-2.7mm]\cline{1-5}\\[-2.65mm]
        \multirow{4}{*}{\textbf{7}} & \textbf{Utterance} & can, you, book, for, arrival, closer, to, 17:30, for, one, person, and, give, me, the, reference, number., also, i, would, like, to, see, a, college, in, centre. & - & -\\ \cline{2-5}
                           & \textbf{Ground Truth} & O, O, O, O, O, O, O, O, O, O, O, O, O, O, O, O, O, O, O, O, O, O, O, O, O, O, O & find\_attraction & attraction\\  \cline{2-5}\\[-2.75mm]\cline{2-5}\\[-2.65mm]
                           & MIDAS (BERT) &  \cellcolor{YellowGreen!25}O, O, O, O, O, O, O, O, O, O, O, O, O, O, O, O, O, O, O, O, O, O, O, O, O, O, O & \cellcolor{YellowGreen!25}find\_attraction &  \cellcolor{YellowGreen!25}attraction\\  \cline{2-5}\\[-2.65mm]
                           & BERT-Only &  \cellcolor{GoldenRod!25}O, O, O, O, O, O, O, B-train\_leaveat, O, O, O, O, O, O, O, O, O, O, O, O, O, O, O, O, O, O, O &  \cellcolor{WildStrawberry!25}book\_train &  \cellcolor{WildStrawberry!25}train\\
                           \cline{1-5}\\[-2.7mm]\cline{1-5}\\[-2.65mm]
        \multirow{4}{*}{\textbf{8}} & \textbf{Utterance} & yes, please., i, need, an, address, and, phone, number, too. & - & -\\ \cline{2-5}
                           & \textbf{Ground Truth} & O, O, O, O, O, O, O, O, O, O & find\_attraction & attraction\\  \cline{2-5}\\[-2.75mm]\cline{2-5}\\[-2.65mm]
                           & MIDAS (BERT) &  \cellcolor{YellowGreen!25}O, O, O, O, O, O, O, O, O, O & \cellcolor{YellowGreen!25}find\_attraction &  \cellcolor{YellowGreen!25}attraction\\  \cline{2-5}\\[-2.65mm]
                           & BERT-Only &  \cellcolor{YellowGreen!25}O, O, O, O, O, O, O, O, O, O &  \cellcolor{WildStrawberry!25}find\_restaurant &  \cellcolor{WildStrawberry!25}restaurant\\
                           \cline{1-5}\\[-2.7mm]\cline{1-5}\\[-2.65mm]
        \multirow{4}{*}{\textbf{9}} & \textbf{Utterance} & just, need, to, know, what, area, "its", in. & - & -\\ \cline{2-5}
                           & \textbf{Ground Truth} & O, O, O, O, O, O, O, O & find\_attraction & attraction\\  \cline{2-5}\\[-2.75mm]\cline{2-5}\\[-2.65mm]
                           & MIDAS (RoBERTa) &  \cellcolor{YellowGreen!25}O, O, O, O, O, O, O, O & \cellcolor{YellowGreen!25}find\_attraction &  \cellcolor{YellowGreen!25}attraction\\  \cline{2-5}\\[-2.65mm]
                           & RoBERTa-Only &  \cellcolor{YellowGreen!25}O, O, O, O, O, O, O, O &  \cellcolor{WildStrawberry!25}find\_hotel &  \cellcolor{WildStrawberry!25}hotel\\
                           \cline{1-5}\\[-2.7mm]\cline{1-5}\\[-2.65mm]
        \multirow{4}{*}{\textbf{10}} & \textbf{Utterance} & i, would, actually, like, to, book, 5, people, and, would, like, to, know, the, reference, number, for, the, tickets, and, the, address, of, la, tasca, restaurant. & - & -\\ \cline{2-5}
                           & \textbf{Ground Truth} & O, O, O, O, O, O, O, O, O, O, O, O, O, O, O, O, O, O, O, O, O, O, O, B-restaurant-name, I-restaurant-name, O & find\_restaurant & restaurant\\  \cline{2-5}\\[-2.75mm]\cline{2-5}\\[-2.65mm]
                           & MIDAS (RoBERTa) &  \cellcolor{YellowGreen!25}O, O, O, O, O, O, O, O, O, O, O, O, O, O, O, O, O, O, O, O, O, O, O, B-restaurant-name, I-restaurant-name, O & \cellcolor{YellowGreen!25}find\_restaurant &  \cellcolor{YellowGreen!25}restaurant\\  \cline{2-5}\\[-2.65mm]
                           & RoBERTa-Only &  \cellcolor{GoldenRod!25}O, O, O, O, O, O, O, O, O, O, O, O, O, O, O, O, O, O, O, O, O, O, O, B-restaurant-name, B-hotel-name, O &  \cellcolor{WildStrawberry!25}book\_train &  \cellcolor{YellowGreen!25}restaurant\\
                           \cline{1-5}\\[-2.7mm]\cline{1-5}\\[-2.65mm]
        \multirow{4}{*}{\textbf{11}} & \textbf{Utterance} & "its", not, a, restaurant,, "its", an, attraction., nusha. & - & -\\ \cline{2-5}
                           & \textbf{Ground Truth} & O, O, O, O, O, O, O, B-attraction-name & find\_attraction & attraction\\  \cline{2-5}\\[-2.75mm]\cline{2-5}\\[-2.65mm]
                           & MIDAS (RoBERTa) &  \cellcolor{YellowGreen!25}O, O, O, O, O, O, O, B-attraction-name & \cellcolor{YellowGreen!25}find\_attraction &  \cellcolor{YellowGreen!25}attraction\\  \cline{2-5}\\[-2.65mm]
                           & RoBERTa-Only &  \cellcolor{GoldenRod!25}O, O, O, O, O, O, O, O &  \cellcolor{YellowGreen!25}find\_attraction &  \cellcolor{WildStrawberry!25}restaurant\\
                           \cline{1-5}\\[-2.7mm]\cline{1-5}\\[-2.65mm]
        \multirow{4}{*}{\textbf{12}} & \textbf{Utterance} & will, you, give, me, the, phone, number,, address,, and, postcode, for, graffiti,, please? & - & -\\ \cline{2-5}
                           & \textbf{Ground Truth} & O, O, O, O, O, O, O, O, O, O, O, B-restaurant-name, O & find\_restaurant & restaurant\\  \cline{2-5}\\[-2.75mm]\cline{2-5}\\[-2.65mm]
                           & MIDAS (RoBERTa) &  \cellcolor{YellowGreen!25}O, O, O, O, O, O, O, O, O, O, O, B-restaurant-name, O & \cellcolor{YellowGreen!25}find\_restaurant &  \cellcolor{YellowGreen!25}restaurant\\  \cline{2-5}\\[-2.65mm]
                           & RoBERTa-Only &  \cellcolor{GoldenRod!25}O, O, O, O, O, O, O, O, O, O, O, O, O &  \cellcolor{WildStrawberry!25}find\_attraction &  \cellcolor{YellowGreen!25}restaurant\\
                           \cline{1-5}\\[-2.7mm]\cline{1-5}\\[-2.65mm]
        \multirow{4}{*}{\textbf{13}} & \textbf{Utterance} & i, am, looking, to, get, to, the, rajmahal, restaurant, please,, how, do, i, get, there? & - & -\\ \cline{2-5}
                           & \textbf{Ground Truth} & O, O, O, O, O, O, O, B-restaurant\_name, O, O, O, O, O, O, O & find\_restaurant & restaurant\\  \cline{2-5}\\[-2.75mm]\cline{2-5}\\[-2.65mm]
                           & MIDAS (RoBERTa) &  \cellcolor{YellowGreen!25}O, O, O, O, O, O, O, B-restaurant\_name, O, O, O, O, O, O, O & \cellcolor{YellowGreen!25}find\_restaurant &  \cellcolor{YellowGreen!25}restaurant\\  \cline{2-5}\\[-2.65mm]
                           & RoBERTa-Only &  \cellcolor{YellowGreen!25}O, O, O, O, O, O, O, B-restaurant\_name, O, O, O, O, O, O, O &  \cellcolor{WildStrawberry!25}find\_taxi &  \cellcolor{WildStrawberry!25}taxi\\
                
        \hline
    \end{tabular}
    \caption{13 Prediction examples with both datasets on slot filling, intent detection, and domain classification results of each model. The first two utterances are from M2M, while the rest are from MultiWOZ 2.2 (MWOZ). The first eight results come from \textit{MIDAS (BERT)} and \text{BERT-Only}, whereas the remaining five results pertain to \textit{MIDAS (RoBERTa)} and \text{RoBERTa-Only}. The green cell represents a result that matches the ground truth, the red cell indicates incorrect results, and the yellow cell indicates partially correct results.}
    \label{tab:ap-case}
\end{table*}

\begin{figure*}[tp]   
\centering
\includegraphics[width=0.95\linewidth]{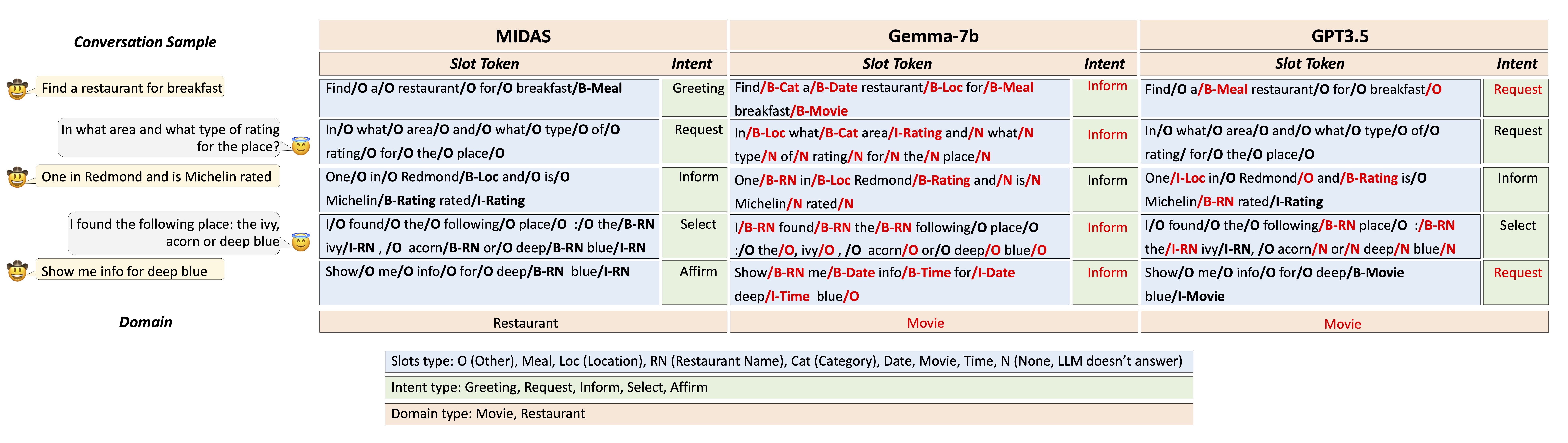}\\
(a)\\
\includegraphics[width=0.95\linewidth]{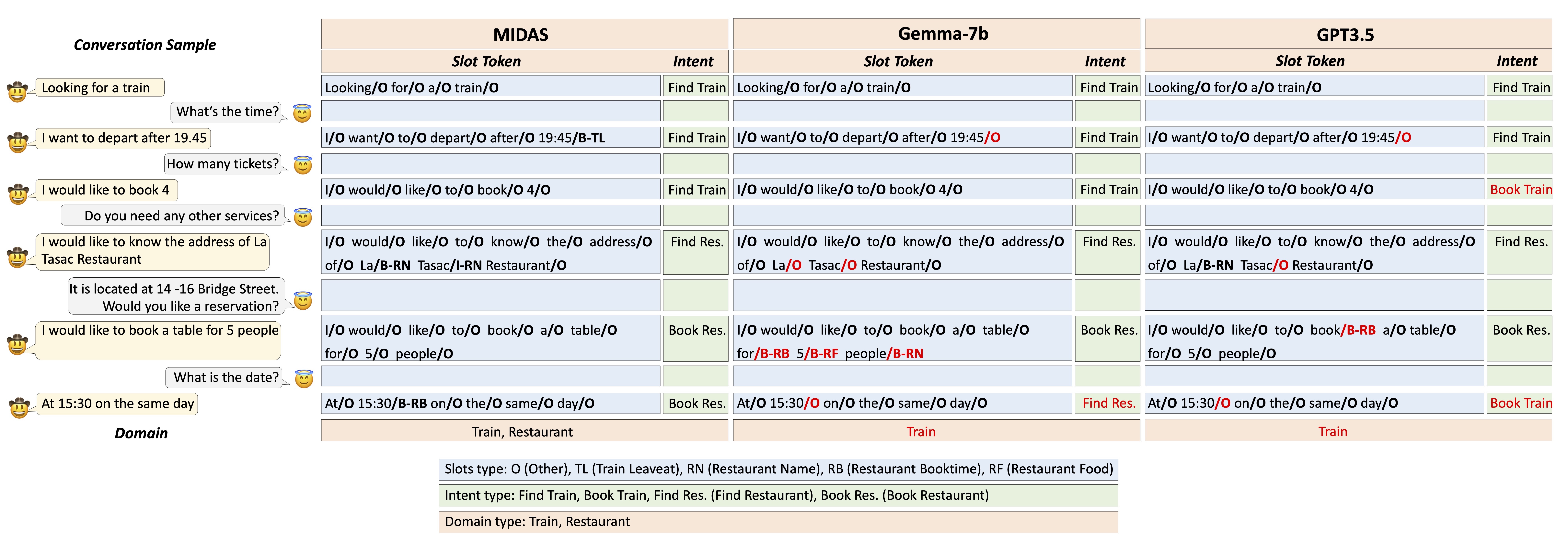}\\
(b)
% \vspace*{-7mm}
\caption{Two examples for qualitative analysis: (a) shows the results on the M2M dataset, and (b) shows the results on the MWOZ dataset. Each example shows the results when MIDAS matches the ground truth. The three cells below each example display the type lists for slot, intent, and domain, and red text indicates errors in each column of the results table.}
\label{fig:qual_analy}
% \vspace{-1mm}
\end{figure*}

\section{In-depth PLM/LLM analysis}
\label{ap:case-study}

In this section, we provide an in-depth quantitative and qualitative analysis, incorporating a detailed comparison between PLMs and LLMs.

\subsection{Compared with PLMs}
We evaluate MIDAS with a qualitative assessment of the three NLU tasks on MWOZ and M2M, compared with two representative PLMs, BERT and RoBERTa. In Table \ref{tab:ap-case}, we test all three NLU tasks, including intent classification, slot filling, and domain classification.
The first two utterances are from M2M, while the rest are from MultiWOZ 2.2 (MWOZ). 
The first eight results come from \textbf{MIDAS (BERT)}, trained with three teachers $BERT_{ID}$, $BERT_{SF}$, and $BERT_{DC}$, and \textbf{BERT-only} refers a single fine-tuned BERT (\textit{BERT-Base}), whereas the remaining five results pertain to \textbf{MIDAS (RoBERTa)}, trained with three teachers $RoBERTa_{ID}$, $RoBERTa_{SF}$, and $RoBERTa_{DC}$, and \textbf{RoBERTa-only} refers a single fine-tuned RoBERTa (\textit{RoBERTa-Base}).
Although the single fine-tuned BERT \textit{(BERT-only)} or RoBERTa \textit{(RoBERTa-only)} can sometimes predict the slots correctly, it does not communicate/integrate with the word level and domain level classification. Instances such as these validate our hypothesis that leveraging diverse knowledge levels from multi-turn conversation datasets can improve the understanding of individual natural language understanding tasks, outperforming the advantages of learning with single-level dialogue knowledge.

\subsection{Compared with LLMs}
\label{ap:prompt}

\subsubsection{Quantitative analysis}
\label{ap:prompt-quant}

\begin{table}[htp]
\scriptsize
\setlength\extrarowheight{1pt}
\setlength\tabcolsep{2.5 pt}
\centering
    \begin{tabular}[b]{l|c|c|c|c|c|c}
        \hline
        \multirow{3}{*}{} & \multicolumn{2}{c|}{\cellcolor{YellowGreen!25}{\small\textbf{ID}}} & \multicolumn{2}{c|}{\cellcolor{YellowGreen!25}{\small\textbf{SF}}} & \multicolumn{2}{c}{\small\textbf{DC}} \\

        % \multirow{3}{*}{} & \multicolumn{2}{c|}{\small\textbf{ID}} & \multicolumn{2}{c|}{\small\textbf{SF}} & \multicolumn{2}{c}{\small\textbf{DC}} \\
        \cline{2-7}
        & \textbf{MWOZ} & \textbf{M2M} & \textbf{MWOZ} & \textbf{M2M} & \textbf{MWOZ} & \textbf{M2M}  \\
        & \textbf{(ACC)} & \textbf{(ACC)} & \textbf{(F1)} & \textbf{(F1)} & \textbf{(ACC)} & \textbf{(ACC)} \\
        \hline 
        \hline 
% BERT base                 & 0.2284              & 0.2311          &          -          &         -       & 0.4113               & 0.7347           \\
% BERT large                & 0.2811              & 0.1680           &          -          &         -       & 0.4390                & 0.7638           \\
% RoBerta base              & 0.0627              & 0.0430           &          -          &         -       & 0.0990                & 0.3432           \\
% RoBerta large             & 0.1421              & 0.0390           &         -           &        -        & 0.0926               & 0.4970             \\
LLaMa2-7b-chat         & 0.4751              & 0.3363          & 0.0217               & 0.0751           & 0.6528             & 0.5231         \\
LLaMa2-13b-chat        & 0.1679              & 0.2013          & 0.0891               & 0.1092           & 0.5602             & 0.4468         \\
LLaMa2-70b-chat  & 0.3896              & 0.3275          & 0.0619               & 0.0883           & 0.6987             & 0.6012         \\
Gemma-7b                       & 0.6515              & 0.4588          & 0.6653               & 0.4357           & 0.7227             & 0.5426         \\
GPT3.5                   & 0.6971              & 0.5100            & 0.8175               & 0.5516           & 0.7739             & 0.7740          \\
GPT4o                   & 0.6789              & 0.6410            & 0.8418               & 0.6616           & 0.7877             & 0.8503          \\
GPT4o$\dagger$                   & 0.7614              & 0.7510            & 0.8525               & 0.7132           & 0.7941             & 0.7051          \\

QWen2-7B-Ins.$\dagger$                   & 0.5459              & 0.278            & 0.1532               & 0.1192           & 0.6416             & 0.6541          \\

LLaMa3.1-8b-Ins.$\dagger$                 & 0.6422              & 0.2715            & 0.6276               & 0.5412           & 0.5973             & 0.5076          \\
\hline
\textbf{Our best model}           & \textbf{0.8502}              & \textbf{0.9427}          & \textbf{0.9928}               & \textbf{0.9856}           & \textbf{0.8816}             & \textbf{0.8952}         \\
     \hline
    \end{tabular}
    \vspace*{0mm}
    \caption{The comparison of the proposed models with prompt tuning methods using Large Language Models. $ID$, $SF$ and $DC$ indicate intent detection, slot filling and domain classification, respectively, as mentioned in Section \ref{subsec:metrics}. ACC and F1 stand for accuracy and micro F1, respectively, and scores in bold indicate leadership among the metrics. $\dagger$ refers to the application of 3-shot learning in the prompt with LLMs.}
    \label{tab:ap-prompt}
    % \vspace*{-5mm}
\end{table}

We measured the performance using the zero-shot prompt method to compare performance with Large Language Model (LLM).
The LLM LLaMa2\cite{touvron2023llama2openfoundation}, LLaMa3.1\cite{dubey2024llama3herdmodels}, Gemma\cite{gemmateam2024gemmaopenmodelsbased}, QWen2\cite{bai2023qwentechnicalreport}, GPT3.5 and GPT4o\cite{brown2020languagemodelsfewshotlearners}, were utilized. Notably, we also tested 3-shot learning on GPT4o, QWen2 and LLaMa3.1. The prompt for each task are shown in Table \ref{tab:app_all_prompts_temp}.

Table \ref{tab:ap-prompt} presents the experimental results of each baseline, compared with the performance of our best model. Notably, GPT4o 3-shot learning achieves the best results in all tasks, except for the DC task on the M2M dataset, though it still falls significantly short of our model’s performance. In the ID and SF tasks, LLaMa’s performance is markedly lower than that of Gemma and GPT, highlighting that factors such as architecture, training data, and training methods, beyond just the number of parameters, also influence LLM performance. 

% Additional details of the experiment can be found in Appendix \ref{app:prompt_cases}.

% RoBERTa consistently shows weaker performance than BERT across all classification tasks.
% Unlike MultiWOZ 2.2, which encompasses 8 domains, M2M comprises only 2 domains.
% It's noteworthy that even though the number of parameters for PLM is considerably smaller than that for LLM,
% BERT demonstrates superiority over LLM in binary classification.
% In simpler classification tasks like M2M DC with only two classes, BERT is comparable to GPT. 
Even within the LLaMa series, the number of model parameters doesn't always determine performance; the 7b model sometimes outperforms the 13b and 70b models. Note that only the 70b model was used with 4-bit quantisation.

Across all three tasks, LLMs occasionally generate out-of-scope class names, despite having all class names provided.
Additionally, in the SF task, LLMs don't always output answers corresponding to the length of the original text. 
Despite our prompt stating that no explanation is needed for efficiency, LLMs sometimes still generate explanations. 
These observations indicate that LLMs don't fully grasp the input.

\begin{table*}[tp]
\renewcommand{\arraystretch}{1.5}
\setlength{\tabcolsep}{2pt}
% \scriptsize
\centering
\begin{tabular}{p{0.8cm}|p{12cm}}
% \noalign{\hrule height 0.8pt}
\hline
\textbf{Task} & \multicolumn{1}{c}{\textbf{Prompt}} \\
\hline
\hline
ID & Definition: In this task, you are given a dialogue. Your job is to classify the following dialogue into one of the fifteen different intents. The intents and examples are: 
\newline
Intent: greeting 
Dialogues: 
1. hi , i want to make a restaurant reservation .
2. hi , i want to make a reservation for 6 pm .
3. hi ! can you book me a restaurant reservation ?
Intent: request 
Dialogues: 
1. okay , where do you want to go , and how many people will there be?
...
\newline
Input: [\{input\}]. Output(only output the intent):\\

\hline
SF  & In the task of slot filling, the B-, I-, and O- prefixes are commonly used to annotate slot types, indicating the boundaries and types of slots. These labels typically represent:
\newline
B- (Begin): Signifies the beginning of a slot, marking the start of a new slot.
I- (Inside): Represents the interior of a slot, indicating a continuation of the slot.
O (Outside): Denotes parts of the input that are not part of any slot.
\newline
For instance, in a sentence where we want to label a "date" slot, words containing date information might be tagged as "B-date" (indicating the beginning of a date slot), followed by consecutive words carrying date information tagged as "I-date" (indicating the continuation of the date slot), while words not containing date information would be tagged as "O" (indicating they are outside any slot). Here are some examples:
\newline
Dialogue: "the sushi boat for 6 .", slot types: ['O', 'B-restaurant\_name', 'I-restaurant\_name', 'O', 'B-num\_people', 'O']
...
\newline
Definition: In this task, you are given a dialogue. Your job is to classify the words in the following dialogue into one of the twenty-one different slots. The slots are: "B-category", "B-date", "B-location", ..., "O". Input: [\{input\}]. Output(Only output slot types. And the slot types should be output as a list without any explanation):\\

\hline
DC & In this task, you are given a dialogue. Your job is to classify the following dialogue into one of the two different intents. The domains and examples are: 
\newline
Domain: restaurant 
Dialogues: 
1. hi , i want to make a restaurant reservation.
2. a reservation for cheese cake factory for 3 people on next monday.
3. ok , please choose between amarin and sakoon restaurants.
\newline
Domain: movie 
Dialogues: 
1. i would like to buy movie tickets for 6:00 pm
2. which movie , and how many tickets do you need ?
3. i need 3 tickets for the movie called a man called love
\newline
Input: [\{input\}]. Output(only output the domain):\\
\hline
% \noalign{\hrule height 0.8pt}
\end{tabular}
\caption{The prompt we used for each dataset in our experiments.}
\label{tab:app_all_prompts_temp}
\end{table*}

\subsubsection{Qualitative analysis}
\label{ap:prompt-qual}

In the qualitative analysis, we first focus on two representative LLMs, Gemma-7b and GPT3.5, as shown in Figure \ref{fig:qual_analy}. 
From the M2M conversation shown in Figure \ref{fig:qual_analy}-(a), we found that both LLMs can not predict slot types based on context. 
For example, GPT3.5 predicts ``Michelin/B-RN rated/I-Rating'' instead of ``Michelin/B-Rating rated/I-Rating''. 
Except for the wrong understanding of the conversation, we found that both LLMs can not follow the prompt all the time.
For example, both LLMs do not predict the slot type for each token, where the missing predictions are represented by ``N''.
From the Multi-Domain Wizard-of-Oz 2.2 (MWOZ) conversation as shown in Figure \ref{fig:qual_analy}-(b), we can see that both LLMs can not make predictions in terms of the whole conversation, resulting the conflicts of the predictions of the domains and intents. 
For example, GPT3.5 predicts ``Book Train'' after ``Book Restaurant'' and Gemma-7b predicts ``Find Restaurant'' after ``Book Restaurant''. 
Another example is that both LLMs failed to predict the domain ``Restaurant'' of the last turn dialogue, even the Gemma-7b already predicted the intent as ``Find Restaurant''.

% \columnbreak

We further analyse the outputs from cutting-edge large language models, including QWen2, LLaMa3.1, and GPT-4o. As outlined in Section \ref{main:qual-analysis}, these models also faced challenges in making use of multi-level knowledge consistently and strictly adhering to prompt instructions. Those examples from both M2M dataset and MultiWOZ dataset are presented from Table \ref{tab:ap-case3} to Table \ref{tab:ap-case8}.

\subsection{Complete procedure cases}
% \label{app:prompt_cases}

\begin{figure}[hpt]
    \centering
    \includegraphics[width=1.0\linewidth]{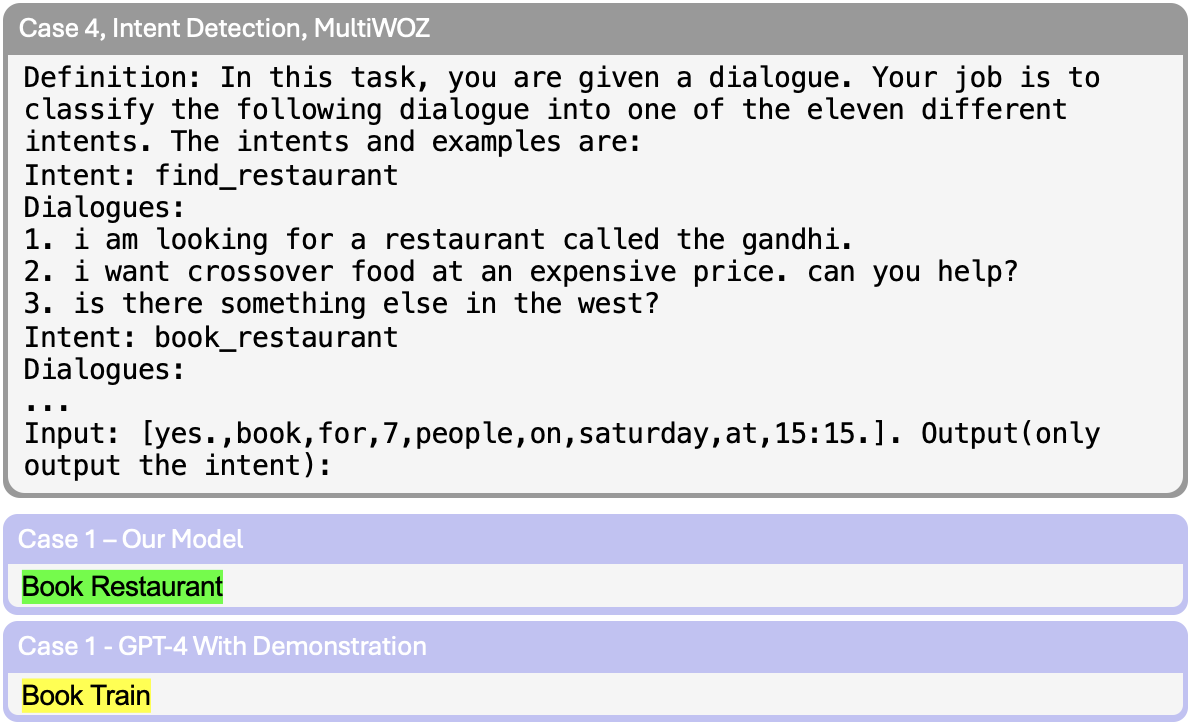}
    \caption{Prompt and output for a sample dialogue in \textbf{MultiWOZ} dataset, where the correct prediction is highlighted in green and wrong predictions are highlighted in red. Demonstration means few-shot (3-shot) learning. Compared to GPT4o, our model can \textbf{correctly} classify the intent of the given dialogue as \textbf{Book Restaurant}.}
    \label{fig:app_case4}
\end{figure}

\begin{figure}[hpt]
    \centering
    \includegraphics[width=1.0\linewidth]{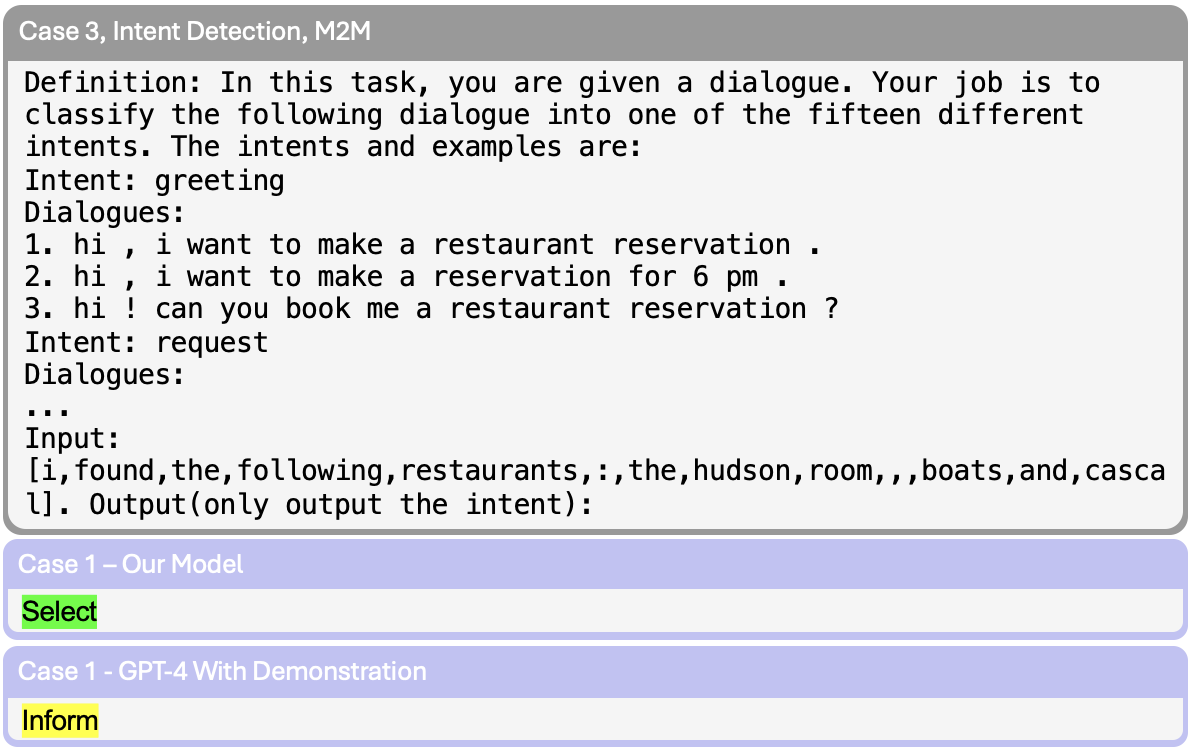}
    \caption{Prompt and output for a sample dialogue in \textbf{M2M} dataset, where the correct prediction is highlighted in green and wrong predictions are highlighted in red. Demonstration means few-shot (3-shot) learning. Compared to GPT4o, our model can \textbf{correctly} classify the intent of the given dialogue as \textbf{Select}.}
    \label{fig:app_case3}
\end{figure}

\begin{figure}[hpt]
    \centering
    \includegraphics[width=1.0\linewidth]{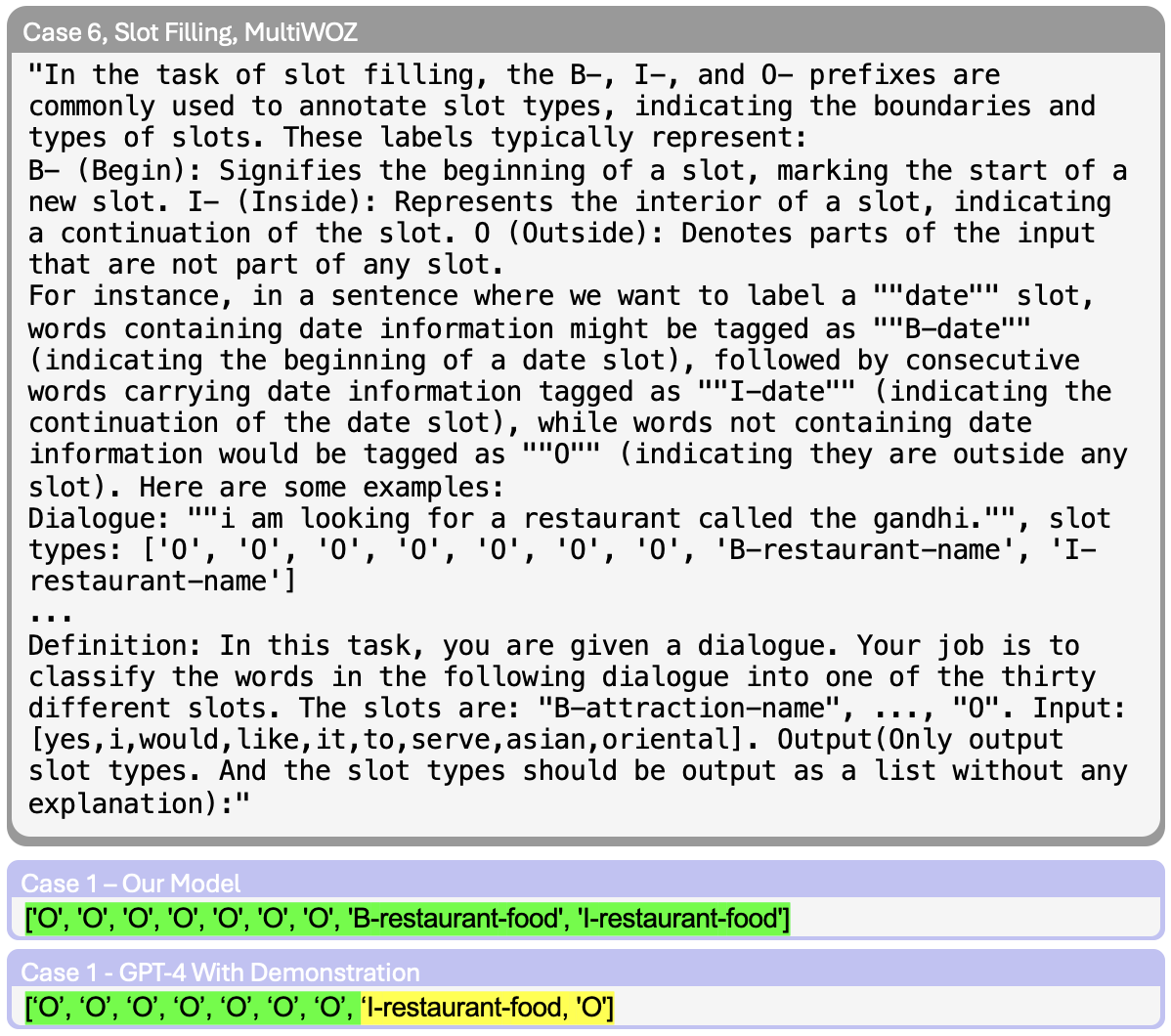}
    \caption{Prompt and output for a sample dialogue in \textbf{MultiWOZ} dataset, where the correct prediction is highlighted in green and wrong predictions are highlighted in red. Demonstration means few-shot (3-shot) learning. Compared to GPT4o, our model can \textbf{correctly} classify the slot types of the given dialogue.}
    \label{fig:app_case6}
\end{figure}

\begin{figure}[hpt]
    \centering
    \includegraphics[width=1.0\linewidth]{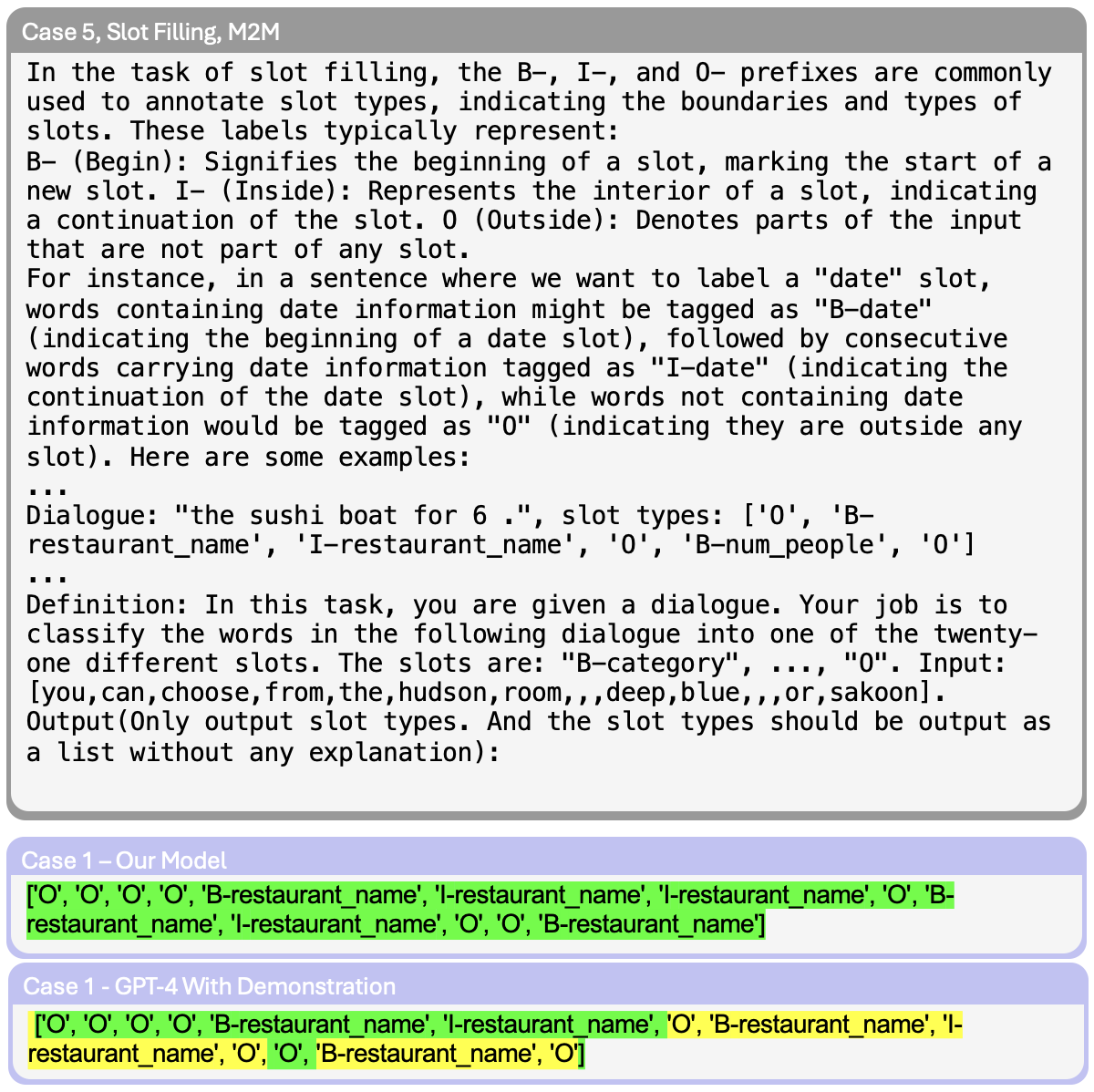}
    \caption{Prompt and output for a sample dialogue in \textbf{M2M} dataset, where the correct prediction is highlighted in green and wrong predictions are highlighted in red. Demonstration means few-shot (3-shot) learning. Compared to GPT4o, our model can \textbf{correctly} classify the slot types of the given dialogue.}
    \label{fig:app_case5}
\end{figure}

\begin{figure}[hpt]
    \centering
    \includegraphics[width=1.0\linewidth]{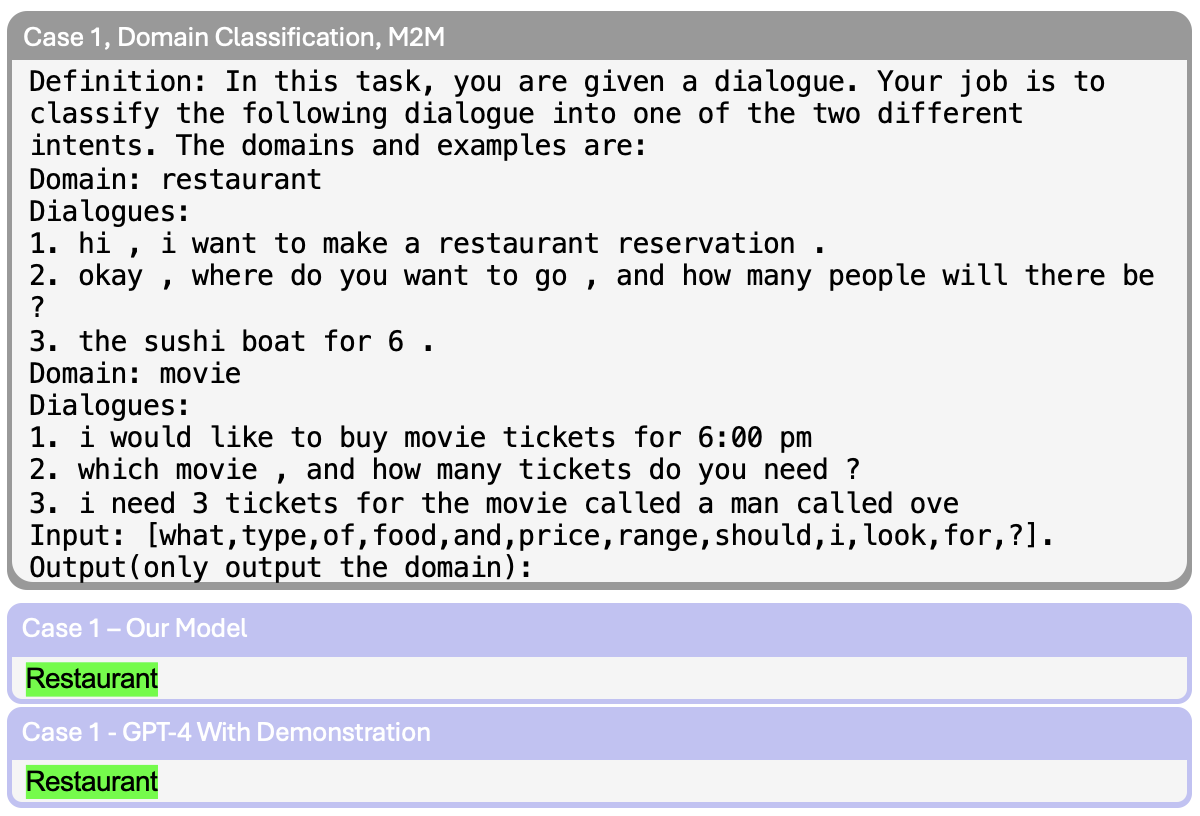}
    \caption{Prompt and output for a sample dialogue in \textbf{M2M} dataset, where the correct prediction is highlighted in green and wrong predictions are highlighted in red. Demonstration means few-shot (3-shot) learning. Both models can \textbf{correctly} classify the domain of the given dialogue as \textbf{Restrurant}.}
    \label{fig:app_case1}
\end{figure}

\begin{figure}[hpt]
    \centering
    \includegraphics[width=1.0\linewidth]{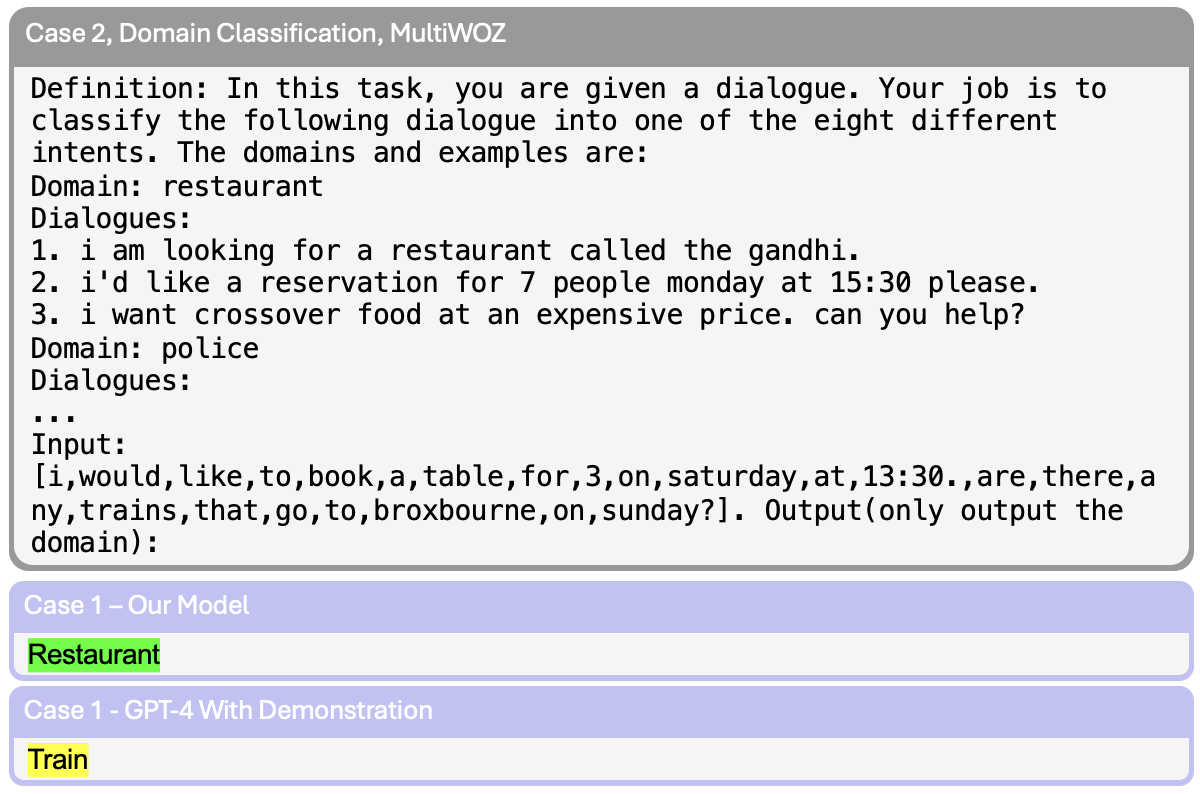}
    \caption{Prompt and output for a sample dialogue in \textbf{MultiWOZ} dataset, where the correct prediction is highlighted in green and wrong predictions are highlighted in red. Demonstration means few-shot (3-shot) learning. Compared to GPT4o, our model can \textbf{correctly} classify the domain of the given dialogue as \textbf{Restaurant}.}
    \label{fig:app_case2}
\end{figure}

In this section, we provide a complete question-answer demonstration for each task and dataset, offering a clearer explanation of how we utilize LLMs for testing, as shown from Figure \ref{fig:app_case4} to Figure \ref{fig:app_case2}.

\newpage

\begin{table*}[htp]
\tiny
\setlength\extrarowheight{1pt}
\setlength\tabcolsep{1pt}
\centering
    \begin{tabular}[b]{c|c|L{9.5cm}|c|c}
        \hline
        \textbf{Model} & ~~~\textbf{Turn}~~~ & \multicolumn{1}{c|}{~~~~~~~~~~~~~~~~\textbf{Tokens (Slot)}~~~~~~~~~~~~~~~~} & ~~~ \textbf{Intent}~~~ & ~~~\textbf{Domain}~~~ \\
        \hline 
        \hline 
        
%ut
\multirow{4}{*}{\textbf{Utterance}} & \textbf{1} &\hspace{1mm}what, are, you, in, the, mood, for, ? & - & -
\\ \cline{2-5} & \textbf{2} &\hspace{1mm}it, doesn, , t, matter, . & - & -
\\ \cline{2-5} & \textbf{3} &\hspace{1mm}how, about, the, hudson, room, or, los, altos, grill, ? & - & -
\\ \cline{2-5} & \textbf{4} &\hspace{1mm}sounds, good, . & - & -
%gt
\\ \cline{1-5}\\[-2.8mm]\cline{1-5} \\ [-2.7mm]\multirow{4}{*}{\textbf{GT}} & \textbf{1} &\hspace{1mm}O, O, O, O, O, O, O, O & request & restaurant
\\ \cline{2-5} & {\textbf{2}} &\hspace{1mm}O, O, O, O, O, O & inform & restaurant
\\ \cline{2-5} & {\textbf{3}} &\hspace{1mm}O, O, B-restaurant\_name, I-restaurant\_name, I-restaurant\_name, O, B-restaurant\_name, I-restaurant\_name, I-restaurant\_name, O & select & restaurant
\\ \cline{2-5} & {\textbf{4}} &\hspace{1mm}O, O, O & affirm & restaurant
%MIDAS (Ours)
\\ \cline{1-5}\\[-2.8mm]\cline{1-5} \\ [-2.7mm]\multirow{4}{*}{\textbf{MIDAS (Ours)}} & \textbf{1}& \cellcolor{YellowGreen!25}\hspace{1mm}O, O, O, O, O, O, O, O & \cellcolor{YellowGreen!25}request & \cellcolor{YellowGreen!25}restaurant
\\ \cline{2-5}\\[-2.8mm]\cline{2-5} \\ [-2.7mm] & \textbf{2}& \cellcolor{YellowGreen!25}\hspace{1mm}O, O, O, O, O, O & \cellcolor{YellowGreen!25}inform & \cellcolor{YellowGreen!25}restaurant
\\ \cline{2-5}\\[-2.8mm]\cline{2-5} \\ [-2.7mm] & \textbf{3}& \cellcolor{YellowGreen!25}\hspace{1mm}O, O, B-restaurant\_name, I-restaurant\_name, I-restaurant\_name, O, B-restaurant\_name, I-restaurant\_name, I-restaurant\_name, O & \cellcolor{YellowGreen!25}select & \cellcolor{YellowGreen!25}restaurant
\\ \cline{2-5}\\[-2.8mm]\cline{2-5} \\ [-2.7mm] & \textbf{4}& \cellcolor{YellowGreen!25}\hspace{1mm}O, O, O & \cellcolor{YellowGreen!25}affirm & \cellcolor{YellowGreen!25}restaurant
%GPT4o
\\ \cline{1-5}\\[-2.8mm]\cline{1-5} \\ [-2.7mm]\multirow{4}{*}{\textbf{GPT4o}} & \textbf{1}& \cellcolor{GoldenRod!25}\hspace{1mm}O, O, O, O, O, O, NaN, NaN & \cellcolor{YellowGreen!25}request & \cellcolor{YellowGreen!25}restaurant
\\ \cline{2-5}\\[-2.8mm]\cline{2-5} \\ [-2.7mm] & \textbf{2}& \cellcolor{GoldenRod!25}\hspace{1mm}O, O, O, O, O, NaN & \cellcolor{WildStrawberry!25}negate & \cellcolor{WildStrawberry!25}(O)
\\ \cline{2-5}\\[-2.8mm]\cline{2-5} \\ [-2.7mm] & \textbf{3}& \cellcolor{GoldenRod!25}\hspace{1mm}O, O, O, B-restaurant\_name, I-restaurant\_name, O, B-restaurant\_name, I-restaurant\_name, O, NaN & \cellcolor{YellowGreen!25}select & \cellcolor{YellowGreen!25}restaurant
\\ \cline{2-5}\\[-2.8mm]\cline{2-5} \\ [-2.7mm] & \textbf{4}& \cellcolor{YellowGreen!25}\hspace{1mm}O, O, O & \cellcolor{YellowGreen!25}affirm & \cellcolor{WildStrawberry!25}movie
%QWen2
\\ \cline{1-5}\\[-2.8mm]\cline{1-5} \\ [-2.7mm]\multirow{4}{*}{\textbf{QWen2}} & \textbf{1}& \cellcolor{GoldenRod!25}\hspace{1mm}B-category, NaN, NaN, NaN, NaN, NaN, NaN, NaN & \cellcolor{WildStrawberry!25}request\_alts & \cellcolor{YellowGreen!25}restaurant
\\ \cline{2-5}\\[-2.8mm]\cline{2-5} \\ [-2.7mm] & \textbf{2}& \cellcolor{GoldenRod!25}\hspace{1mm}O, O, NaN, NaN, NaN, NaN & \cellcolor{WildStrawberry!25}negate & \cellcolor{YellowGreen!25}restaurant
\\ \cline{2-5}\\[-2.8mm]\cline{2-5} \\ [-2.7mm] & \textbf{3}& \cellcolor{GoldenRod!25}\hspace{1mm}B-restaurant\_name, I-restaurant\_name, NaN, NaN, NaN, NaN, NaN, NaN, NaN, NaN & \cellcolor{WildStrawberry!25}request\_alts & \cellcolor{YellowGreen!25}restaurant
\\ \cline{2-5}\\[-2.8mm]\cline{2-5} \\ [-2.7mm] & \textbf{4}& \cellcolor{GoldenRod!25}\hspace{1mm}O, O, NaN & \cellcolor{WildStrawberry!25}inform & \cellcolor{WildStrawberry!25}movie
%LLaMa3.1
\\ \cline{1-5}\\[-2.8mm]\cline{1-5} \\ [-2.7mm]\multirow{4}{*}{\textbf{LLaMa3.1}} & \textbf{1}& \cellcolor{GoldenRod!25}\hspace{1mm}O, O, O, O, O, NaN, NaN, NaN & \cellcolor{YellowGreen!25}request & \cellcolor{WildStrawberry!25}movie
\\ \cline{2-5}\\[-2.8mm]\cline{2-5} \\ [-2.7mm] & \textbf{2}& \cellcolor{YellowGreen!25}\hspace{1mm}O, O, O, O, O, O & \cellcolor{WildStrawberry!25}other & \cellcolor{WildStrawberry!25}movie
\\ \cline{2-5}\\[-2.8mm]\cline{2-5} \\ [-2.7mm] & \textbf{3}& \cellcolor{GoldenRod!25}\hspace{1mm}O, O, O, O, O, O, O, O, O, O & \cellcolor{WildStrawberry!25}affirm & \cellcolor{WildStrawberry!25}movie
\\ \cline{2-5}\\[-2.8mm]\cline{2-5} \\ [-2.7mm] & \textbf{4}& \cellcolor{YellowGreen!25}\hspace{1mm}O, O, O & \cellcolor{WildStrawberry!25}other & \cellcolor{YellowGreen!25}restaurant
\\

    \hline
    \end{tabular}
    \caption{A four turns conversation from M2M dataset. GPT4o exhibited issues with generating out-of-domain outputs `(O)' in Turn 2, while all LLMs showed problems in the SF task by failing to follow instructions, resulting in NaN outputs.}
    \label{tab:ap-case3}
\end{table*}

\begin{table*}[htp]
\tiny
\setlength\extrarowheight{1pt}
\setlength\tabcolsep{1pt}
\centering
    \begin{tabular}[b]{c|c|L{9.5cm}|c|c}
        \hline
        \textbf{Model} & ~~~\textbf{Turn}~~~ & \multicolumn{1}{c|}{~~~~~~~~~~~~~~~~\textbf{Tokens (Slot)}~~~~~~~~~~~~~~~~} & ~~~ \textbf{Intent}~~~ & ~~~\textbf{Domain}~~~ \\
        \hline 
        \hline 
        
%ut
\multirow{4}{*}{\textbf{Utterance}} & \textbf{1} &\hspace{1mm}what, area, is, your, preference, ? & - & -
\\ \cline{2-5} & \textbf{2} &\hspace{1mm}orlando & - & -
\\ \cline{2-5} & \textbf{3} &\hspace{1mm}i, have, the, hudson, room, ,, boats, or, the, nest, ,, which, sounds, better, ? & - & -
\\ \cline{2-5} & \textbf{4} &\hspace{1mm}agree, on, the, hudson, room & - & -
%gt
\\ \cline{1-5}\\[-2.8mm]\cline{1-5} \\ [-2.7mm]\multirow{4}{*}{\textbf{GT}} & \textbf{1} &\hspace{1mm}O, O, O, O, O, O & request & restaurant
\\ \cline{2-5} & {\textbf{2}} &\hspace{1mm}B-location & inform & restaurant
\\ \cline{2-5} & {\textbf{3}} &\hspace{1mm}O, O, B-restaurant\_name, I-restaurant\_name, I-restaurant\_name, O, B-restaurant\_name, O, B-restaurant\_name, I-restaurant\_name, O, O, O, O, O & select & restaurant
\\ \cline{2-5} & {\textbf{4}} &\hspace{1mm}O, O, B-restaurant\_name, I-restaurant\_name, I-restaurant\_name & affirm & restaurant
%MIDAS (Ours)
\\ \cline{1-5}\\[-2.8mm]\cline{1-5} \\ [-2.7mm]\multirow{4}{*}{\textbf{MIDAS (Ours)}} & \textbf{1}& \cellcolor{YellowGreen!25}\hspace{1mm}O, O, O, O, O, O & \cellcolor{YellowGreen!25}request & \cellcolor{YellowGreen!25}restaurant
\\ \cline{2-5}\\[-2.8mm]\cline{2-5} \\ [-2.7mm] & \textbf{2}& \cellcolor{YellowGreen!25}\hspace{1mm}B-location & \cellcolor{YellowGreen!25}inform & \cellcolor{YellowGreen!25}restaurant
\\ \cline{2-5}\\[-2.8mm]\cline{2-5} \\ [-2.7mm] & \textbf{3}& \cellcolor{YellowGreen!25}\hspace{1mm}O, O, B-restaurant\_name, I-restaurant\_name, I-restaurant\_name, O, B-restaurant\_name, O, B-restaurant\_name, I-restaurant\_name, O, O, O, O, O & \cellcolor{YellowGreen!25}select & \cellcolor{YellowGreen!25}restaurant
\\ \cline{2-5}\\[-2.8mm]\cline{2-5} \\ [-2.7mm] & \textbf{4}& \cellcolor{YellowGreen!25}\hspace{1mm}O, O, B-restaurant\_name, I-restaurant\_name, I-restaurant\_name & \cellcolor{YellowGreen!25}affirm & \cellcolor{YellowGreen!25}restaurant
%GPT4o
\\ \cline{1-5}\\[-2.8mm]\cline{1-5} \\ [-2.7mm]\multirow{4}{*}{\textbf{GPT4o}} & \textbf{1}& \cellcolor{GoldenRod!25}\hspace{1mm}O, O, O, O, O, NaN & \cellcolor{YellowGreen!25}request & \cellcolor{YellowGreen!25}restaurant
\\ \cline{2-5}\\[-2.8mm]\cline{2-5} \\ [-2.7mm] & \textbf{2}& \cellcolor{YellowGreen!25}\hspace{1mm}B-location & \cellcolor{WildStrawberry!25}(O) & \cellcolor{WildStrawberry!25}(O)
\\ \cline{2-5}\\[-2.8mm]\cline{2-5} \\ [-2.7mm] & \textbf{3}& \cellcolor{GoldenRod!25}\hspace{1mm}O, O, O, B-restaurant\_name, I-restaurant\_name, O, O, O, O, O, B-restaurant\_name, O, O, O, O & \cellcolor{YellowGreen!25}select & \cellcolor{YellowGreen!25}restaurant
\\ \cline{2-5}\\[-2.8mm]\cline{2-5} \\ [-2.7mm] & \textbf{4}& \cellcolor{GoldenRod!25}\hspace{1mm}O, O, B-restaurant\_name, I-restaurant\_name, NaN & \cellcolor{WildStrawberry!25}inform & \cellcolor{YellowGreen!25}restaurant
%QWen2
\\ \cline{1-5}\\[-2.8mm]\cline{1-5} \\ [-2.7mm]\multirow{4}{*}{\textbf{QWen2}} & \textbf{1}& \cellcolor{GoldenRod!25}\hspace{1mm}B-location, NaN, NaN, NaN, NaN, NaN & \cellcolor{WildStrawberry!25}request\_alts & \cellcolor{WildStrawberry!25}movie
\\ \cline{2-5}\\[-2.8mm]\cline{2-5} \\ [-2.7mm] & \textbf{2}& \cellcolor{YellowGreen!25}\hspace{1mm}B-location & \cellcolor{YellowGreen!25}inform & \cellcolor{WildStrawberry!25}movie
\\ \cline{2-5}\\[-2.8mm]\cline{2-5} \\ [-2.7mm] & \textbf{3}& \cellcolor{GoldenRod!25}\hspace{1mm}B-location, B-location, NaN, NaN, NaN, NaN, NaN, NaN, NaN, NaN, NaN, NaN, NaN, NaN, NaN & \cellcolor{WildStrawberry!25}request\_alts & \cellcolor{WildStrawberry!25}movie
\\ \cline{2-5}\\[-2.8mm]\cline{2-5} \\ [-2.7mm] & \textbf{4}& \cellcolor{GoldenRod!25}\hspace{1mm}B-restaurant\_name, I-location, NaN, NaN, NaN & \cellcolor{WildStrawberry!25}select & \cellcolor{YellowGreen!25}restaurant
%LLaMa3.1
\\ \cline{1-5}\\[-2.8mm]\cline{1-5} \\ [-2.7mm]\multirow{4}{*}{\textbf{LLaMa3.1}} & \textbf{1}& \cellcolor{GoldenRod!25}\hspace{1mm}O, O, O, O, O, NaN & \cellcolor{WildStrawberry!25}greeting & \cellcolor{WildStrawberry!25}movie
\\ \cline{2-5}\\[-2.8mm]\cline{2-5} \\ [-2.7mm] & \textbf{2}& \cellcolor{GoldenRod!25}\hspace{1mm}O & \cellcolor{WildStrawberry!25}confirm & \cellcolor{WildStrawberry!25}movie
\\ \cline{2-5}\\[-2.8mm]\cline{2-5} \\ [-2.7mm] & \textbf{3}& \cellcolor{GoldenRod!25}\hspace{1mm}O, O, O, B-location, O, B-movie, O, NaN, NaN, NaN, NaN, NaN, NaN, NaN, NaN & \cellcolor{WildStrawberry!25}request\_alts & \cellcolor{WildStrawberry!25}movie
\\ \cline{2-5}\\[-2.8mm]\cline{2-5} \\ [-2.7mm] & \textbf{4}& \cellcolor{GoldenRod!25}\hspace{1mm}O, O, O, B-location, I-location & \cellcolor{WildStrawberry!25}thank\_you & \cellcolor{WildStrawberry!25}movie

\\

    \hline
    \end{tabular}
    \caption{A four turns conversation from M2M dataset. GPT4o exhibited issues with generating out-of-domain outputs `(O)' in Turn 2, while all LLMs showed problems in the SF task by failing to follow instructions, resulting in NaN outputs.}
    \label{tab:ap-case4}
\end{table*}

\begin{table*}[htp]
\tiny
\setlength\extrarowheight{1pt}
\setlength\tabcolsep{1pt}
\centering
    \begin{tabular}[b]{c|c|L{9.5cm}|c|c}
        \hline
        \textbf{Model} & ~~~\textbf{Turn}~~~ & \multicolumn{1}{c|}{~~~~~~~~~~~~~~~~\textbf{Tokens (Slot)}~~~~~~~~~~~~~~~~} & ~~~ \textbf{Intent}~~~ & ~~~\textbf{Domain}~~~ \\
        \hline 
        \hline 
        
%ut
\multirow{3}{*}{\textbf{Utterance}} & \textbf{1} &\hspace{1mm}how, many, tickets, would, you, like, to, buy, ? & - & -
\\ \cline{2-5} & \textbf{2} &\hspace{1mm}1 & - & -
\\ \cline{2-5} & \textbf{3} &\hspace{1mm}what, date, do, you, want, to, go, ? & - & -
%gt
\\ \cline{1-5}\\[-2.8mm]\cline{1-5} \\ [-2.7mm]\multirow{3}{*}{\textbf{GT}} & \textbf{1} &\hspace{1mm}O, O, O, O, O, O, O, O, O & request & movie
\\ \cline{2-5} & {\textbf{2}} &\hspace{1mm}B-num\_tickets & inform & movie
\\ \cline{2-5} & {\textbf{3}} &\hspace{1mm}O, O, O, O, O, O, O, O & request & movie
%MIDAS (Ours)
\\ \cline{1-5}\\[-2.8mm]\cline{1-5} \\ [-2.7mm]\multirow{3}{*}{\textbf{MIDAS (Ours)}} & \textbf{1}& \cellcolor{YellowGreen!25}\hspace{1mm}O, O, O, O, O, O, O, O, O & \cellcolor{YellowGreen!25}request & \cellcolor{YellowGreen!25}movie
\\ \cline{2-5}\\[-2.8mm]\cline{2-5} \\ [-2.7mm] & \textbf{2}& \cellcolor{YellowGreen!25}\hspace{1mm}B-num\_tickets & \cellcolor{YellowGreen!25}inform & \cellcolor{YellowGreen!25}movie
\\ \cline{2-5}\\[-2.8mm]\cline{2-5} \\ [-2.7mm] & \textbf{3}& \cellcolor{YellowGreen!25}\hspace{1mm}O, O, O, O, O, O, O, O & \cellcolor{YellowGreen!25}request & \cellcolor{YellowGreen!25}movie
%GPT4o
\\ \cline{1-5}\\[-2.8mm]\cline{1-5} \\ [-2.7mm]\multirow{3}{*}{\textbf{GPT4o}} & \textbf{1}& \cellcolor{GoldenRod!25}\hspace{1mm}O, O, O, O, O, O, O, NaN, NaN & \cellcolor{YellowGreen!25}request & \cellcolor{YellowGreen!25}movie
\\ \cline{2-5}\\[-2.8mm]\cline{2-5} \\ [-2.7mm] & \textbf{2}& \cellcolor{GoldenRod!25}\hspace{1mm}(O) & \cellcolor{WildStrawberry!25}(O) & \cellcolor{WildStrawberry!25}restaurant
\\ \cline{2-5}\\[-2.8mm]\cline{2-5} \\ [-2.7mm] & \textbf{3}& \cellcolor{GoldenRod!25}\hspace{1mm}O, B-date, O, O, O, O, NaN, NaN & \cellcolor{YellowGreen!25}request & \cellcolor{WildStrawberry!25}restaurant
%QWen2
\\ \cline{1-5}\\[-2.8mm]\cline{1-5} \\ [-2.7mm]\multirow{3}{*}{\textbf{QWen2}} & \textbf{1}& \cellcolor{GoldenRod!25}\hspace{1mm}B-num\_tickets, NaN, NaN, NaN, NaN, NaN, NaN, NaN, NaN & \cellcolor{WildStrawberry!25}request\_alts & \cellcolor{YellowGreen!25}movie
\\ \cline{2-5}\\[-2.8mm]\cline{2-5} \\ [-2.7mm] & \textbf{2}& \cellcolor{GoldenRod!25}\hspace{1mm}B-category & \cellcolor{YellowGreen!25}inform & \cellcolor{YellowGreen!25}movie
\\ \cline{2-5}\\[-2.8mm]\cline{2-5} \\ [-2.7mm] & \textbf{3}& \cellcolor{GoldenRod!25}\hspace{1mm}B-date, NaN, NaN, NaN, NaN, NaN, NaN, NaN & \cellcolor{WildStrawberry!25}request\_alts & \cellcolor{WildStrawberry!25}restaurant
%LLaMa3.1
\\ \cline{1-5}\\[-2.8mm]\cline{1-5} \\ [-2.7mm]\multirow{3}{*}{\textbf{LLaMa3.1}} & \textbf{1}& \cellcolor{GoldenRod!25}\hspace{1mm}O, O, O, O, O, O, B-num\_tickets, O, NaN & \cellcolor{YellowGreen!25}request & \cellcolor{WildStrawberry!25}restaurant
\\ \cline{2-5}\\[-2.8mm]\cline{2-5} \\ [-2.7mm] & \textbf{2}& \cellcolor{GoldenRod!25}\hspace{1mm}B-category & \cellcolor{WildStrawberry!25}request & \cellcolor{YellowGreen!25}movie
\\ \cline{2-5}\\[-2.8mm]\cline{2-5} \\ [-2.7mm] & \textbf{3}& \cellcolor{YellowGreen!25}\hspace{1mm}O, O, O, O, O, O, O, O & \cellcolor{YellowGreen!25}request & \cellcolor{YellowGreen!25}movie

\\

    \hline
    \end{tabular}
    \caption{A three turns conversation from M2M dataset. GPT4o exhibited issues with generating out-of-domain outputs `(O)' in Turn 2, while all LLMs showed problems in the SF task by failing to follow instructions, resulting in NaN outputs. LLaMa3.1 failed to unify multi-level knowledge, resulting in predicting the domain as `restaurant' even after identifying `B-num\_tickets'}
    \label{tab:ap-case5}
\end{table*}

\begin{table*}[htp]
\tiny
\setlength\extrarowheight{1pt}
\setlength\tabcolsep{1pt}
\centering
    \begin{tabular}[b]{c|c|L{9.5cm}|c|c}
        \hline
        \textbf{Model} & ~~~\textbf{Turn}~~~ & \multicolumn{1}{c|}{~~~~~~~~~~~~~~~~\textbf{Tokens (Slot)}~~~~~~~~~~~~~~~~} & ~~~ \textbf{Intent}~~~ & ~~~\textbf{Domain}~~~ \\
        \hline 
        \hline 
        
%ut
\multirow{5}{*}{\textbf{Utterance}} & \textbf{1} &\hspace{1mm}1, pm, is, unavailable, again, ,, please, choose, from, 6, pm, or, 6.30, pm, . & - & -
\\ \cline{2-5} & \textbf{2} &\hspace{1mm}no, ,, thank, you & - & -
\\ \cline{2-5} & \textbf{3} &\hspace{1mm}please, provide, the, theatre, name, ,, date, ,, and, number, of, tickets & - & -
\\ \cline{2-5} & \textbf{4} &\hspace{1mm}the, theatre, name, is, aquarius, and, i, need, 1, ticket, for, friday, . & - & -
\\ \cline{2-5} & \textbf{5} &\hspace{1mm}what, movie, would, you, like, to, see, ? & - & -
%gt
\\ \cline{1-5}\\[-2.8mm]\cline{1-5} \\ [-2.7mm]\multirow{5}{*}{\textbf{GT}} & \textbf{1} &\hspace{1mm}B-time, I-time, O, O, O, O, O, O, O, B-time, I-time, O, B-time, I-time, O & negate & movie
\\ \cline{2-5} & {\textbf{2}} &\hspace{1mm}O, O, O, O & negate & movie
\\ \cline{2-5} & {\textbf{3}} &\hspace{1mm}O, O, O, O, O, O, O, O, O, O, O, O & request & movie
\\ \cline{2-5} & {\textbf{4}} &\hspace{1mm}O, O, O, O, B-theatre\_name, O, O, O, B-num\_tickets, O, O, B-date, O & inform & movie
\\ \cline{2-5} & {\textbf{5}} &\hspace{1mm}O, O, O, O, O, O, O, O & request & movie
%MIDAS (Ours)
\\ \cline{1-5}\\[-2.8mm]\cline{1-5} \\ [-2.7mm]\multirow{5}{*}{\textbf{MIDAS (Ours)}} & \textbf{1}& \cellcolor{YellowGreen!25}\hspace{1mm}B-time, I-time, O, O, O, O, O, O, O, B-time, I-time, O, B-time, I-time, O & \cellcolor{YellowGreen!25}negate & \cellcolor{YellowGreen!25}movie
\\ \cline{2-5}\\[-2.8mm]\cline{2-5} \\ [-2.7mm] & \textbf{2}& \cellcolor{YellowGreen!25}\hspace{1mm}O, O, O, O & \cellcolor{YellowGreen!25}negate & \cellcolor{YellowGreen!25}movie
\\ \cline{2-5}\\[-2.8mm]\cline{2-5} \\ [-2.7mm] & \textbf{3}& \cellcolor{YellowGreen!25}\hspace{1mm}O, O, O, O, O, O, O, O, O, O, O, O & \cellcolor{YellowGreen!25}request & \cellcolor{YellowGreen!25}movie
\\ \cline{2-5}\\[-2.8mm]\cline{2-5} \\ [-2.7mm] & \textbf{4}& \cellcolor{YellowGreen!25}\hspace{1mm}O, O, O, O, B-theatre\_name, O, O, O, B-num\_tickets, O, O, B-date, O & \cellcolor{YellowGreen!25}inform & \cellcolor{YellowGreen!25}movie
\\ \cline{2-5}\\[-2.8mm]\cline{2-5} \\ [-2.7mm] & \textbf{5}& \cellcolor{YellowGreen!25}\hspace{1mm}O, O, O, O, O, O, O, O & \cellcolor{YellowGreen!25}request & \cellcolor{YellowGreen!25}movie
%GPT4o
\\ \cline{1-5}\\[-2.8mm]\cline{1-5} \\ [-2.7mm]\multirow{5}{*}{\textbf{GPT4o}} & \textbf{1}& \cellcolor{GoldenRod!25}\hspace{1mm}B-time, I-time, O, O, O, O, O, O, O, B-time, I-time, O, O, B-time, I-time & \cellcolor{YellowGreen!25}negate & \cellcolor{YellowGreen!25}movie
\\ \cline{2-5}\\[-2.8mm]\cline{2-5} \\ [-2.7mm] & \textbf{2}& \cellcolor{GoldenRod!25}\hspace{1mm}O, O, O, NaN & \cellcolor{YellowGreen!25}negate & \cellcolor{WildStrawberry!25}(O)
\\ \cline{2-5}\\[-2.8mm]\cline{2-5} \\ [-2.7mm] & \textbf{3}& \cellcolor{GoldenRod!25}\hspace{1mm}O, O, O, B-theatre\_name, I-theatre\_name, O, B-date, O, O, B-num\_tickets, O, NaN & \cellcolor{YellowGreen!25}request & \cellcolor{YellowGreen!25}movie
\\ \cline{2-5}\\[-2.8mm]\cline{2-5} \\ [-2.7mm] & \textbf{4}& \cellcolor{GoldenRod!25}\hspace{1mm}O, B-theatre\_name, I-theatre\_name, O, O, O, O, B-num\_tickets, O, O, B-date, O, NaN & \cellcolor{YellowGreen!25}inform & \cellcolor{YellowGreen!25}movie
\\ \cline{2-5}\\[-2.8mm]\cline{2-5} \\ [-2.7mm] & \textbf{5}& \cellcolor{GoldenRod!25}\hspace{1mm}O, B-movie, O, O, O, O, O, NaN & \cellcolor{YellowGreen!25}request & \cellcolor{YellowGreen!25}movie
%QWen2
\\ \cline{1-5}\\[-2.8mm]\cline{1-5} \\ [-2.7mm]\multirow{5}{*}{\textbf{QWen2}} & \textbf{1}& \cellcolor{GoldenRod!25}\hspace{1mm}B-time, I-time, I-time, NaN, NaN, NaN, NaN, NaN, NaN, NaN, NaN, NaN, NaN, NaN, NaN & \cellcolor{WildStrawberry!25}request\_alts & \cellcolor{WildStrawberry!25}(O)
\\ \cline{2-5}\\[-2.8mm]\cline{2-5} \\ [-2.7mm] & \textbf{2}& \cellcolor{GoldenRod!25}\hspace{1mm}O, O, NaN, NaN & \cellcolor{YellowGreen!25}negate & \cellcolor{WildStrawberry!25}restaurant
\\ \cline{2-5}\\[-2.8mm]\cline{2-5} \\ [-2.7mm] & \textbf{3}& \cellcolor{GoldenRod!25}\hspace{1mm}B-theatre\_name, B-date, B-num\_tickets, NaN, NaN, NaN, NaN, NaN, NaN, NaN, NaN, NaN & \cellcolor{WildStrawberry!25}request\_alts & \cellcolor{YellowGreen!25}movie
\\ \cline{2-5}\\[-2.8mm]\cline{2-5} \\ [-2.7mm] & \textbf{4}& \cellcolor{GoldenRod!25}\hspace{1mm}B-theatre\_name, B-date, B-num\_tickets, NaN, NaN, NaN, NaN, NaN, NaN, NaN, NaN, NaN, NaN & \cellcolor{WildStrawberry!25}request & \cellcolor{YellowGreen!25}movie
\\ \cline{2-5}\\[-2.8mm]\cline{2-5} \\ [-2.7mm] & \textbf{5}& \cellcolor{GoldenRod!25}\hspace{1mm}B-movie, NaN, NaN, NaN, NaN, NaN, NaN, NaN & \cellcolor{WildStrawberry!25}request\_alts & \cellcolor{YellowGreen!25}movie
%LLaMa3.1
\\ \cline{1-5}\\[-2.8mm]\cline{1-5} \\ [-2.7mm]\multirow{5}{*}{\textbf{LLaMa3.1}} & \textbf{1}& \cellcolor{GoldenRod!25}\hspace{1mm}O, O, O, B-time, I-time, O, O, O, O, O, O, O, O, O, O & \cellcolor{WildStrawberry!25}request & \cellcolor{WildStrawberry!25}restaurant
\\ \cline{2-5}\\[-2.8mm]\cline{2-5} \\ [-2.7mm] & \textbf{2}& \cellcolor{GoldenRod!25}\hspace{1mm}O, O, NaN, NaN & \cellcolor{WildStrawberry!25}thank\_you & \cellcolor{WildStrawberry!25}restaurant
\\ \cline{2-5}\\[-2.8mm]\cline{2-5} \\ [-2.7mm] & \textbf{3}& \cellcolor{GoldenRod!25}\hspace{1mm}B-theatre\_name, I-date, B-num\_tickets, NaN, NaN, NaN, NaN, NaN, NaN, NaN, NaN, NaN & \cellcolor{WildStrawberry!25}other & \cellcolor{YellowGreen!25}movie
\\ \cline{2-5}\\[-2.8mm]\cline{2-5} \\ [-2.7mm] & \textbf{4}& \cellcolor{GoldenRod!25}\hspace{1mm}B-theatre\_name, I-theatre\_name, O, O, O, B-num\_tickets, (O)I-num\_tickets, O, B-date, I-date, O, NaN, NaN & \cellcolor{WildStrawberry!25}request & \cellcolor{YellowGreen!25}movie
\\ \cline{2-5}\\[-2.8mm]\cline{2-5} \\ [-2.7mm] & \textbf{5}& \cellcolor{YellowGreen!25}\hspace{1mm}O, O, O, O, O, O, O, O & \cellcolor{YellowGreen!25}request & \cellcolor{YellowGreen!25}movie

\\

    \hline
    \end{tabular}
    \caption{A five turns conversation from M2M dataset. GPT4o, QWen2 and LLaMa3.1 exhibited issues with generating out-of-domain outputs `(O)' in Turn 2, Turn 1 and Turn 4 respectively, while all LLMs showed problems in the SF task by failing to follow instructions, resulting in NaN outputs.}
    \label{tab:ap-case6}
\end{table*}

% Multiwoz cases

\begin{table*}[htp]
\tiny
\setlength\extrarowheight{1pt}
\setlength\tabcolsep{1pt}
\centering
    \begin{tabular}[b]{c|c|L{9.5cm}|c|c}
        \hline
        \textbf{Model} & ~~~\textbf{Turn}~~~ & \multicolumn{1}{c|}{~~~~~~~~~~~~~~~~\textbf{Tokens (Slot)}~~~~~~~~~~~~~~~~} & ~~~ \textbf{Intent}~~~ & ~~~\textbf{Domain}~~~ \\
        \hline 
        \hline 
        
%ut
\multirow{9}{*}{\textbf{Utterance}} & \textbf{1} &\hspace{1mm}i, choose, the, ashley, hotel., what, is, their, address,, please? & - & -
\\ \cline{2-5} & \textbf{2} &\hspace{1mm}no,, i, just, need, the, address. & - & -
\\ \cline{2-5} & \textbf{3} &\hspace{1mm}hey, i, am, looking, for, a, train, from, cambridge, to, bishops, stortford., mind, helping, out? & - & -
\\ \cline{2-5} & \textbf{4} &\hspace{1mm}i, want, to, leave, on, monday, and, arrive, by, 18:45. & - & -
\\ \cline{2-5} & \textbf{5} &\hspace{1mm}could, you, give, me, the, travel, time, and, price, of, that, train, please? & - & -
\\ \cline{2-5} & \textbf{6} &\hspace{1mm}i, also, want, a, cheap, chinese, restaurant. & - & -
\\ \cline{2-5} & \textbf{7} &\hspace{1mm}id, like, to, be, in, the, centre, please. & - & -
\\ \cline{2-5} & \textbf{8} &\hspace{1mm}yes., monday,, 8, people,, 10:30. & - & -
\\ \cline{2-5} & \textbf{9} &\hspace{1mm}i, am, planning, a, trip, in, cambridge & - & -
%gt
\\ \cline{1-5}\\[-2.8mm]\cline{1-5} \\ [-2.7mm]\multirow{9}{*}{\textbf{GT}} & \textbf{1} &\hspace{1mm}O, O, O, B-hotel-name, I-hotel-name, O, O, O, O, O & find\_hotel & hotel
\\ \cline{2-5} & {\textbf{2}} &\hspace{1mm}O, O, O, O, O, O & find\_hotel & hotel
\\ \cline{2-5} & {\textbf{3}} &\hspace{1mm}O, O, O, O, O, O, O, O, O, O, O, O, O, O, O & find\_train & train
\\ \cline{2-5} & {\textbf{4}} &\hspace{1mm}O, O, O, O, O, O, O, O, O, B-train-arriveby & find\_train & train
\\ \cline{2-5} & {\textbf{5}} &\hspace{1mm}O, O, O, O, O, O, O, O, O, O, O, O, O & find\_train & train
\\ \cline{2-5} & {\textbf{6}} &\hspace{1mm}O, O, O, O, O, B-restaurant-food, O & find\_restaurant & restaurant
\\ \cline{2-5} & {\textbf{7}} &\hspace{1mm}O, O, O, O, O, O, O, O & find\_restaurant & restaurant
\\ \cline{2-5} & {\textbf{8}} &\hspace{1mm}O, O, O, O, B-restaurant-booktime & book\_restaurant & restaurant
\\ \cline{2-5} & {\textbf{9}} &\hspace{1mm}O, O, O, O, O, O, O & find\_train & train
%MIDAS (Ours)
\\ \cline{1-5}\\[-2.8mm]\cline{1-5} \\ [-2.7mm]\multirow{9}{*}{\textbf{MIDAS (Ours)}} & \textbf{1}& \cellcolor{YellowGreen!25}\hspace{1mm}O, O, O, B-hotel-name, I-hotel-name, O, O, O, O, O & \cellcolor{YellowGreen!25}find\_hotel & \cellcolor{YellowGreen!25}hotel
\\ \cline{2-5}\\[-2.8mm]\cline{2-5} \\ [-2.7mm] & \textbf{2}& \cellcolor{YellowGreen!25}\hspace{1mm}O, O, O, O, O, O & \cellcolor{YellowGreen!25}find\_hotel & \cellcolor{YellowGreen!25}hotel
\\ \cline{2-5}\\[-2.8mm]\cline{2-5} \\ [-2.7mm] & \textbf{3}& \cellcolor{YellowGreen!25}\hspace{1mm}O, O, O, O, O, O, O, O, O, O, O, O, O, O, O & \cellcolor{YellowGreen!25}find\_train & \cellcolor{YellowGreen!25}train
\\ \cline{2-5}\\[-2.8mm]\cline{2-5} \\ [-2.7mm] & \textbf{4}& \cellcolor{YellowGreen!25}\hspace{1mm}O, O, O, O, O, O, O, O, O, B-train-arriveby & \cellcolor{YellowGreen!25}find\_train & \cellcolor{YellowGreen!25}train
\\ \cline{2-5}\\[-2.8mm]\cline{2-5} \\ [-2.7mm] & \textbf{5}& \cellcolor{YellowGreen!25}\hspace{1mm}O, O, O, O, O, O, O, O, O, O, O, O, O & \cellcolor{YellowGreen!25}find\_train & \cellcolor{YellowGreen!25}train
\\ \cline{2-5}\\[-2.8mm]\cline{2-5} \\ [-2.7mm] & \textbf{6}& \cellcolor{YellowGreen!25}\hspace{1mm}O, O, O, O, O, B-restaurant-food, O & \cellcolor{YellowGreen!25}find\_restaurant & \cellcolor{YellowGreen!25}restaurant
\\ \cline{2-5}\\[-2.8mm]\cline{2-5} \\ [-2.7mm] & \textbf{7}& \cellcolor{YellowGreen!25}\hspace{1mm}O, O, O, O, O, O, O, O & \cellcolor{YellowGreen!25}find\_restaurant & \cellcolor{YellowGreen!25}restaurant
\\ \cline{2-5}\\[-2.8mm]\cline{2-5} \\ [-2.7mm] & \textbf{8}& \cellcolor{YellowGreen!25}\hspace{1mm}O, O, O, O, B-restaurant-booktime & \cellcolor{YellowGreen!25}book\_restaurant & \cellcolor{YellowGreen!25}restaurant
\\ \cline{2-5}\\[-2.8mm]\cline{2-5} \\ [-2.7mm] & \textbf{9}& \cellcolor{YellowGreen!25}\hspace{1mm}O, O, O, O, O, O, O & \cellcolor{YellowGreen!25}find\_train & \cellcolor{YellowGreen!25}train
%GPT4o
\\ \cline{1-5}\\[-2.8mm]\cline{1-5} \\ [-2.7mm]\multirow{9}{*}{\textbf{GPT4o}} & \textbf{1}& \cellcolor{YellowGreen!25}\hspace{1mm}O, O, O, B-hotel-name, I-hotel-name, O, O, O, O, O & \cellcolor{YellowGreen!25}find\_hotel & \cellcolor{YellowGreen!25}hotel
\\ \cline{2-5}\\[-2.8mm]\cline{2-5} \\ [-2.7mm] & \textbf{2}& \cellcolor{GoldenRod!25}\hspace{1mm}O, O, O, O, O, NaN & \cellcolor{WildStrawberry!25}find\_police & \cellcolor{WildStrawberry!25}police
\\ \cline{2-5}\\[-2.8mm]\cline{2-5} \\ [-2.7mm] & \textbf{3}& \cellcolor{GoldenRod!25}\hspace{1mm}O, O, O, O, O, O, O, (O)B-train-departure, O, (O)B-train-destination, (O)I-train-destination, O, O, O, O & \cellcolor{YellowGreen!25}find\_train & \cellcolor{YellowGreen!25}train
\\ \cline{2-5}\\[-2.8mm]\cline{2-5} \\ [-2.7mm] & \textbf{4}& \cellcolor{GoldenRod!25}\hspace{1mm}O, O, O, O, O, O, B-train-arriveby, I-train-arriveby, NaN, NaN & \cellcolor{YellowGreen!25}find\_train & \cellcolor{YellowGreen!25}train
\\ \cline{2-5}\\[-2.8mm]\cline{2-5} \\ [-2.7mm] & \textbf{5}& \cellcolor{YellowGreen!25}\hspace{1mm}O, O, O, O, O, O, O, O, O, O, O, O, O & \cellcolor{YellowGreen!25}find\_train & \cellcolor{YellowGreen!25}train
\\ \cline{2-5}\\[-2.8mm]\cline{2-5} \\ [-2.7mm] & \textbf{6}& \cellcolor{GoldenRod!25}\hspace{1mm}O, O, O, O, B-restaurant-food, I-restaurant-food, O & \cellcolor{YellowGreen!25}find\_restaurant & \cellcolor{YellowGreen!25}restaurant
\\ \cline{2-5}\\[-2.8mm]\cline{2-5} \\ [-2.7mm] & \textbf{7}& \cellcolor{GoldenRod!25}\hspace{1mm}O, O, O, O, O, O, O, NaN & \cellcolor{WildStrawberry!25}find\_hotel & \cellcolor{WildStrawberry!25}hotel
\\ \cline{2-5}\\[-2.8mm]\cline{2-5} \\ [-2.7mm] & \textbf{8}& \cellcolor{GoldenRod!25}\hspace{1mm}O, O, O, B-restaurant-booktime, O & \cellcolor{YellowGreen!25}book\_restaurant & \cellcolor{WildStrawberry!25}train
\\ \cline{2-5}\\[-2.8mm]\cline{2-5} \\ [-2.7mm] & \textbf{9}& \cellcolor{GoldenRod!25}\hspace{1mm}O, O, O, O, O, B-bus-destination, NaN & \cellcolor{WildStrawberry!25}find\_attraction & \cellcolor{WildStrawberry!25}attraction
%QWen2
\\ \cline{1-5}\\[-2.8mm]\cline{1-5} \\ [-2.7mm]\multirow{9}{*}{\textbf{QWen2}} & \textbf{1}& \cellcolor{GoldenRod!25}\hspace{1mm}B-hotel-name, I-hotel-name, NaN, NaN, NaN, NaN, NaN, NaN, NaN, NaN & \cellcolor{YellowGreen!25}find\_hotel & \cellcolor{YellowGreen!25}hotel
\\ \cline{2-5}\\[-2.8mm]\cline{2-5} \\ [-2.7mm] & \textbf{2}& \cellcolor{GoldenRod!25}\hspace{1mm}O, O, NaN, NaN, NaN, NaN & \cellcolor{YellowGreen!25}find\_hotel & \cellcolor{YellowGreen!25}hotel
\\ \cline{2-5}\\[-2.8mm]\cline{2-5} \\ [-2.7mm] & \textbf{3}& \cellcolor{GoldenRod!25}\hspace{1mm}(O)B-train-destination, (O)B-train-origin, NaN, NaN, NaN, NaN, NaN, NaN, NaN, NaN, NaN, NaN, NaN, NaN, NaN & \cellcolor{WildStrawberry!25}book\_train & \cellcolor{YellowGreen!25}train
\\ \cline{2-5}\\[-2.8mm]\cline{2-5} \\ [-2.7mm] & \textbf{4}& \cellcolor{GoldenRod!25}\hspace{1mm}B-train-leaveat, (O)I-train-arrivevi, NaN, NaN, NaN, NaN, NaN, NaN, NaN, NaN & \cellcolor{WildStrawberry!25}book\_train & \cellcolor{WildStrawberry!25}hotel
\\ \cline{2-5}\\[-2.8mm]\cline{2-5} \\ [-2.7mm] & \textbf{5}& \cellcolor{GoldenRod!25}\hspace{1mm}O, O, NaN, NaN, NaN, NaN, NaN, NaN, NaN, NaN, NaN, NaN, NaN & \cellcolor{YellowGreen!25}find\_train & \cellcolor{WildStrawberry!25}taxi
\\ \cline{2-5}\\[-2.8mm]\cline{2-5} \\ [-2.7mm] & \textbf{6}& \cellcolor{GoldenRod!25}\hspace{1mm}I-restaurant-food, NaN, NaN, NaN, NaN, NaN, NaN & \cellcolor{YellowGreen!25}find\_restaurant & \cellcolor{YellowGreen!25}restaurant
\\ \cline{2-5}\\[-2.8mm]\cline{2-5} \\ [-2.7mm] & \textbf{7}& \cellcolor{GoldenRod!25}\hspace{1mm}I-hotel-stars, NaN, NaN, NaN, NaN, NaN, NaN, NaN & \cellcolor{WildStrawberry!25}find\_hotel & \cellcolor{WildStrawberry!25}hotel
\\ \cline{2-5}\\[-2.8mm]\cline{2-5} \\ [-2.7mm] & \textbf{8}& \cellcolor{GoldenRod!25}\hspace{1mm}B-train-leaveat, I-restaurant-booktime, NaN, NaN, NaN & \cellcolor{WildStrawberry!25}book\_train & \cellcolor{WildStrawberry!25}hotel
\\ \cline{2-5}\\[-2.8mm]\cline{2-5} \\ [-2.7mm] & \textbf{9}& \cellcolor{GoldenRod!25}\hspace{1mm}(O)B-city, B-hotel-name, NaN, NaN, NaN, NaN, NaN & \cellcolor{WildStrawberry!25}find\_attraction & \cellcolor{WildStrawberry!25}hotel
%LLaMa3.1
\\ \cline{1-5}\\[-2.8mm]\cline{1-5} \\ [-2.7mm]\multirow{9}{*}{\textbf{LLaMa3.1}} & \textbf{1}& \cellcolor{GoldenRod!25}\hspace{1mm}O, O, O, O, B-hotel-name, O, O, O, O, O & \cellcolor{YellowGreen!25}find\_hotel & \cellcolor{YellowGreen!25}hotel
\\ \cline{2-5}\\[-2.8mm]\cline{2-5} \\ [-2.7mm] & \textbf{2}& \cellcolor{GoldenRod!25}\hspace{1mm}B-restaurant-name, I-restaurant-name, O, O, NaN, NaN & \cellcolor{YellowGreen!25}find\_hotel & \cellcolor{WildStrawberry!25}restaurant
\\ \cline{2-5}\\[-2.8mm]\cline{2-5} \\ [-2.7mm] & \textbf{3}& \cellcolor{GoldenRod!25}\hspace{1mm}B-train-leaveat, O, (O)B-train-destination, O, O, O, O, O, O, O, O, O, O, O, O & \cellcolor{YellowGreen!25}find\_train & \cellcolor{YellowGreen!25}train
\\ \cline{2-5}\\[-2.8mm]\cline{2-5} \\ [-2.7mm] & \textbf{4}& \cellcolor{GoldenRod!25}\hspace{1mm}B-train-leaveat, O, O, B-train-arriveby, O, O, O, O, O, O & \cellcolor{WildStrawberry!25}book\_train & \cellcolor{YellowGreen!25}train
\\ \cline{2-5}\\[-2.8mm]\cline{2-5} \\ [-2.7mm] & \textbf{5}& \cellcolor{GoldenRod!25}\hspace{1mm}B-train-arriveby, B-train-leaveat, O, O, O, O, O, O, O, O, O, O, O & \cellcolor{YellowGreen!25}find\_train & \cellcolor{YellowGreen!25}train
\\ \cline{2-5}\\[-2.8mm]\cline{2-5} \\ [-2.7mm] & \textbf{6}& \cellcolor{GoldenRod!25}\hspace{1mm}O, O, O, B-restaurant-food, O, O, O & \cellcolor{YellowGreen!25}find\_restaurant & \cellcolor{YellowGreen!25}restaurant
\\ \cline{2-5}\\[-2.8mm]\cline{2-5} \\ [-2.7mm] & \textbf{7}& \cellcolor{YellowGreen!25}\hspace{1mm}O, O, O, O, O, O, O, O & \cellcolor{WildStrawberry!25}find\_taxi & \cellcolor{WildStrawberry!25}attraction
\\ \cline{2-5}\\[-2.8mm]\cline{2-5} \\ [-2.7mm] & \textbf{8}& \cellcolor{GoldenRod!25}\hspace{1mm}B-attraction-name, I-attraction-name, O, O, O & \cellcolor{WildStrawberry!25}find\_restaurant & \cellcolor{WildStrawberry!25}bus
\\ \cline{2-5}\\[-2.8mm]\cline{2-5} \\ [-2.7mm] & \textbf{9}& \cellcolor{YellowGreen!25}\hspace{1mm}O, O, O, O, O, O, O & \cellcolor{WildStrawberry!25}find\_attraction & \cellcolor{WildStrawberry!25}restaurant

\\

    \hline
    \end{tabular}
    \caption{A nine turns conversation from MultiWOZ dataset. All LLMs showed problems generating out-of-domain outputs ‘(O)’ and problems in the SF task by failing to follow instructions, resulting in NaN outputs.}
    \label{tab:ap-case7}
\end{table*}

\begin{table*}[htp]
\tiny
\setlength\extrarowheight{1pt}
\setlength\tabcolsep{1pt}
\centering
    \begin{tabular}[b]{c|c|L{9.5cm}|c|c}
        \hline
        \textbf{Model} & ~~~\textbf{Turn}~~~ & \multicolumn{1}{c|}{~~~~~~~~~~~~~~~~\textbf{Tokens (Slot)}~~~~~~~~~~~~~~~~} & ~~~ \textbf{Intent}~~~ & ~~~\textbf{Domain}~~~ \\
        \hline 
        \hline 
        
%ut
\multirow{6}{*}{\textbf{Utterance}} & \textbf{1} &\hspace{1mm}yes, i, am, looking, for, a, place, to, stay, in, cambridge, that, is, 3, stars, and, expensive., can, you, help, me? & - & -
\\ \cline{2-5} & \textbf{2} &\hspace{1mm}do, those, both, have, 3, star, ratings, and, are, expensive? & - & -
\\ \cline{2-5} & \textbf{3} &\hspace{1mm}do, they, have, free, parking? & - & -
\\ \cline{2-5} & \textbf{4} &\hspace{1mm}not, today, thanks., im, also, want, to, find, a, cinema, in, the, west, part, of, town. & - & -
\\ \cline{2-5} & \textbf{5} &\hspace{1mm}are, there, any, colleges, in, the, west, that, i, could, visit, instead? & - & -
\\ \cline{2-5} & \textbf{6} &\hspace{1mm}what, is, the, one, that, is, free?, can, i, get, the, phone, number, and, postcode? & - & -
%gt
\\ \cline{1-5}\\[-2.8mm]\cline{1-5} \\ [-2.7mm]\multirow{6}{*}{\textbf{GT}} & \textbf{1} &\hspace{1mm}O, O, O, O, O, O, O, O, O, O, O, O, O, O, O, O, O, O, O, O, O & find\_hotel & hotel
\\ \cline{2-5} & {\textbf{2}} &\hspace{1mm}O, O, O, O, O, O, O, O, O, O & find\_hotel & hotel
\\ \cline{2-5} & {\textbf{3}} &\hspace{1mm}O, O, O, O, O & find\_hotel & hotel
\\ \cline{2-5} & {\textbf{4}} &\hspace{1mm}O, O, O, O, O, O, O, O, O, O, O, O, O, O, O, O & find\_attraction & attraction
\\ \cline{2-5} & {\textbf{5}} &\hspace{1mm}O, O, O, O, O, O, O, O, O, O, O, O & find\_attraction & attraction
\\ \cline{2-5} & {\textbf{6}} &\hspace{1mm}O, O, O, O, O, O, O, O, O, O, O, O, O, O, O & find\_attraction & attraction
%MIDAS (Ours)
\\ \cline{1-5}\\[-2.8mm]\cline{1-5} \\ [-2.7mm]\multirow{6}{*}{\textbf{MIDAS (Ours)}} & \textbf{1}& \cellcolor{YellowGreen!25}\hspace{1mm}O, O, O, O, O, O, O, O, O, O, O, O, O, O, O, O, O, O, O, O, O & \cellcolor{YellowGreen!25}find\_hotel & \cellcolor{YellowGreen!25}hotel
\\ \cline{2-5}\\[-2.8mm]\cline{2-5} \\ [-2.7mm] & \textbf{2}& \cellcolor{YellowGreen!25}\hspace{1mm}O, O, O, O, O, O, O, O, O, O & \cellcolor{YellowGreen!25}find\_hotel & \cellcolor{YellowGreen!25}hotel
\\ \cline{2-5}\\[-2.8mm]\cline{2-5} \\ [-2.7mm] & \textbf{3}& \cellcolor{YellowGreen!25}\hspace{1mm}O, O, O, O, O & \cellcolor{YellowGreen!25}find\_hotel & \cellcolor{YellowGreen!25}hotel
\\ \cline{2-5}\\[-2.8mm]\cline{2-5} \\ [-2.7mm] & \textbf{4}& \cellcolor{YellowGreen!25}\hspace{1mm}O, O, O, O, O, O, O, O, O, O, O, O, O, O, O, O & \cellcolor{YellowGreen!25}find\_attraction & \cellcolor{YellowGreen!25}attraction
\\ \cline{2-5}\\[-2.8mm]\cline{2-5} \\ [-2.7mm] & \textbf{5}& \cellcolor{YellowGreen!25}\hspace{1mm}O, O, O, O, O, O, O, O, O, O, O, O & \cellcolor{YellowGreen!25}find\_attraction & \cellcolor{YellowGreen!25}attraction
\\ \cline{2-5}\\[-2.8mm]\cline{2-5} \\ [-2.7mm] & \textbf{6}& \cellcolor{YellowGreen!25}\hspace{1mm}O, O, O, O, O, O, O, O, O, O, O, O, O, O, O & \cellcolor{YellowGreen!25}find\_attraction & \cellcolor{YellowGreen!25}attraction
%GPT4o
\\ \cline{1-5}\\[-2.8mm]\cline{1-5} \\ [-2.7mm]\multirow{6}{*}{\textbf{GPT4o}} & \textbf{1}& \cellcolor{GoldenRod!25}\hspace{1mm}O, O, O, O, O, O, O, O, O, O, O, O, O, B-hotel-stars, I-hotel-stars, O, O, O, O, O, NaN & \cellcolor{YellowGreen!25}find\_hotel & \cellcolor{YellowGreen!25}hotel
\\ \cline{2-5}\\[-2.8mm]\cline{2-5} \\ [-2.7mm] & \textbf{2}& \cellcolor{GoldenRod!25}\hspace{1mm}O, O, O, O, B-hotel-stars, I-hotel-stars, O, O, O, NaN & \cellcolor{WildStrawberry!25}find\_restaurant & \cellcolor{WildStrawberry!25}restaurant
\\ \cline{2-5}\\[-2.8mm]\cline{2-5} \\ [-2.7mm] & \textbf{3}& \cellcolor{YellowGreen!25}\hspace{1mm}O, O, O, O, O & \cellcolor{YellowGreen!25}find\_hotel & \cellcolor{YellowGreen!25}hotel
\\ \cline{2-5}\\[-2.8mm]\cline{2-5} \\ [-2.7mm] & \textbf{4}& \cellcolor{YellowGreen!25}\hspace{1mm}O, O, O, O, O, O, O, O, O, O, O, O, O, O, O, O & \cellcolor{YellowGreen!25}find\_attraction & \cellcolor{YellowGreen!25}attraction
\\ \cline{2-5}\\[-2.8mm]\cline{2-5} \\ [-2.7mm] & \textbf{5}& \cellcolor{GoldenRod!25}\hspace{1mm}O, O, O, O, O, O, O, O, O, O, NaN, NaN & \cellcolor{YellowGreen!25}find\_attraction & \cellcolor{YellowGreen!25}attraction
\\ \cline{2-5}\\[-2.8mm]\cline{2-5} \\ [-2.7mm] & \textbf{6}& \cellcolor{YellowGreen!25}\hspace{1mm}O, O, O, O, O, O, O, O, O, O, O, O, O, O, O & \cellcolor{YellowGreen!25}find\_attraction & \cellcolor{YellowGreen!25}attraction
%QWen2
\\ \cline{1-5}\\[-2.8mm]\cline{1-5} \\ [-2.7mm]\multirow{6}{*}{\textbf{QWen2}} & \textbf{1}& \cellcolor{GoldenRod!25}\hspace{1mm}B-hotel-name, B-hotel-stars, (O)B-price-range, NaN, NaN, NaN, NaN, NaN, NaN, NaN, NaN, NaN, NaN, NaN, NaN, NaN, NaN, NaN, NaN, NaN, NaN & \cellcolor{YellowGreen!25}find\_hotel & \cellcolor{YellowGreen!25}hotel
\\ \cline{2-5}\\[-2.8mm]\cline{2-5} \\ [-2.7mm] & \textbf{2}& \cellcolor{GoldenRod!25}\hspace{1mm}B-hotel-stars, B-hotel-type, NaN, NaN, NaN, NaN, NaN, NaN, NaN, NaN & \cellcolor{YellowGreen!25}find\_hotel & \cellcolor{WildStrawberry!25}restaurant
\\ \cline{2-5}\\[-2.8mm]\cline{2-5} \\ [-2.7mm] & \textbf{3}& \cellcolor{GoldenRod!25}\hspace{1mm}O, O, NaN, NaN, NaN & \cellcolor{YellowGreen!25}find\_hotel & \cellcolor{WildStrawberry!25}restaurant
\\ \cline{2-5}\\[-2.8mm]\cline{2-5} \\ [-2.7mm] & \textbf{4}& \cellcolor{GoldenRod!25}\hspace{1mm}B-attraction-name, NaN, NaN, NaN, NaN, NaN, NaN, NaN, NaN, NaN, NaN, NaN, NaN, NaN, NaN, NaN & \cellcolor{YellowGreen!25}find\_attraction & \cellcolor{YellowGreen!25}attraction
\\ \cline{2-5}\\[-2.8mm]\cline{2-5} \\ [-2.7mm] & \textbf{5}& \cellcolor{YellowGreen!25}\hspace{1mm}O, O, O, O, O, O, O, O, O, O, O, O & \cellcolor{YellowGreen!25}find\_attraction & \cellcolor{YellowGreen!25}attraction
\\ \cline{2-5}\\[-2.8mm]\cline{2-5} \\ [-2.7mm] & \textbf{6}& \cellcolor{GoldenRod!25}\hspace{1mm}O, O, O, NaN, NaN, NaN, NaN, NaN, NaN, NaN, NaN, NaN, NaN, NaN, NaN & \cellcolor{YellowGreen!25}find\_attraction & \cellcolor{WildStrawberry!25}restaurant
%LLaMa3.1
\\ \cline{1-5}\\[-2.8mm]\cline{1-5} \\ [-2.7mm]\multirow{6}{*}{\textbf{LLaMa3.1}} & \textbf{1}& \cellcolor{GoldenRod!25}\hspace{1mm}B-hotel-stars, I-hotel-stars, O, B-hotel-type, O, B-hotel-stars, I-hotel-stars, O, B-hotel-type, O, B-hotel-stars, I-hotel-stars, O, B-hotel-type, O, NaN, NaN, NaN, NaN, NaN, NaN & \cellcolor{WildStrawberry!25}book\_hotel & \cellcolor{YellowGreen!25}hotel
\\ \cline{2-5}\\[-2.8mm]\cline{2-5} \\ [-2.7mm] & \textbf{2}& \cellcolor{YellowGreen!25}\hspace{1mm}O, O, O, O, O, O, O, O, O, O & \cellcolor{YellowGreen!25}find\_hotel & \cellcolor{WildStrawberry!25}restaurant
\\ \cline{2-5}\\[-2.8mm]\cline{2-5} \\ [-2.7mm] & \textbf{3}& \cellcolor{GoldenRod!25}\hspace{1mm}O, O, NaN, NaN, NaN & \cellcolor{YellowGreen!25}find\_hotel & \cellcolor{YellowGreen!25}hotel
\\ \cline{2-5}\\[-2.8mm]\cline{2-5} \\ [-2.7mm] & \textbf{4}& \cellcolor{YellowGreen!25}\hspace{1mm}O, O, O, O, O, O, O, O, O, O, O, O, O, O, O, O & \cellcolor{YellowGreen!25}find\_attraction & \cellcolor{WildStrawberry!25}restaurant
\\ \cline{2-5}\\[-2.8mm]\cline{2-5} \\ [-2.7mm] & \textbf{5}& \cellcolor{YellowGreen!25}\hspace{1mm}O, O, O, O, O, O, O, O, O, O, O, O & \cellcolor{YellowGreen!25}find\_attraction & \cellcolor{YellowGreen!25}attraction
\\ \cline{2-5}\\[-2.8mm]\cline{2-5} \\ [-2.7mm] & \textbf{6}& \cellcolor{YellowGreen!25}\hspace{1mm}O, O, O, O, O, O, O, O, O, O, O, O, O, O, O & \cellcolor{WildStrawberry!25}find\_hotel & \cellcolor{WildStrawberry!25}hotel \\             
    \hline
    \end{tabular}
    \caption{A six turns conversation from MultiWOZ dataset. QWen2 exhibited issues with generating out-of-domain outputs `(O)' in Turn 1, while all LLMs showed problems in the SF task by failing to follow instructions, resulting in NaN outputs.}
    \label{tab:ap-case8}
\end{table*}

\end{document}